\documentclass[10pt,twocolumn,letterpaper]{article}

\usepackage[pagenumbers]{cvpr} 

\usepackage{graphicx}
\usepackage{amsmath}
\usepackage{amssymb}
\usepackage{booktabs}
\usepackage{enumitem}
\usepackage{wrapfig}

\usepackage[linesnumbered,ruled]{algorithm2e}

\usepackage{graphicx}
\usepackage{amsmath, amssymb}
\usepackage{enumitem, array}
\usepackage[flushleft]{threeparttable}
\usepackage{algpseudocode}

\newcommand{\Req}{\textbf{Require:}\hspace*{0.5em}}

\newcommand{\X}{\hspace*{3mm}}
\newcommand{\XX}{\X\X}
\newcommand{\XXX}{\X\X\X}
\newcommand{\XXXX}{\X\X\X\X}

\newcommand{\vy}{\mathbf{y}}

\newcommand{\vx}{\mathbf{x}}

\usepackage{xspace}

\makeatletter
\DeclareRobustCommand\onedot{\futurelet\@let@token\@onedot}
\def\@onedot{\ifx\@let@token.\else.\null\fi\xspace}

\def\eg{\emph{e.g}\onedot}

\def\etc{\emph{etc}\onedot}

\makeatother

\usepackage{hyperref}
\hypersetup{hidelinks,
backref=true,
pagebackref=true,
hyperindex=true,
breaklinks=true,
colorlinks=true,
urlcolor=blue,
bookmarks=true,
bookmarksopen=false,
pdftitle={Title},
pdfauthor={Author}}

\usepackage{amssymb}
\usepackage{pifont}
\newcommand{\cmark}{\ding{51}}%
\newcommand{\xmark}{\ding{55}}%
\usepackage{multirow}
\usepackage{wrapfig}

\usepackage{amsmath,amsfonts,bm}

\def\eqref#1{equation~\ref{#1}}

\def\1{\bm{1}}

\def\vx{{\bm{x}}}
\def\vy{{\bm{y}}}

\DeclareMathAlphabet{\mathsfit}{\encodingdefault}{\sfdefault}{m}{sl}
\SetMathAlphabet{\mathsfit}{bold}{\encodingdefault}{\sfdefault}{bx}{n}

\DeclareMathOperator*{\argmax}{arg\,max}

\usepackage{hyperref}
\usepackage{url}
\usepackage{caption}
\usepackage{subcaption}
 \usepackage{multirow}
 \usepackage{enumitem}
 \usepackage{wrapfig}
\usepackage[T1]{fontenc}    
\usepackage{amsfonts}       
\usepackage{nicefrac}       
\usepackage{microtype}      
\usepackage{xcolor}         
\usepackage{float}

\usepackage[capitalize]{cleveref}
\crefname{section}{Sec.}{Secs.}
\Crefname{section}{Section}{Sections}
\Crefname{table}{Table}{Tables}
\crefname{table}{Tab.}{Tabs.}

\def\cvprPaperID{10878} 
\def\confName{CVPR}
\def\confYear{2022}

\begin{document}

\title{ 
Adaptive Early-Learning Correction for Segmentation from Noisy Annotations
}
\newcommand*\samethanks[1][\value{footnote}]{\footnotemark[#1]}
\author{First Author\\
Institution1\\
Institution1 address\\
{\tt\small firstauthor@i1.org}
\and
Second Author\\
Institution2\\
First line of institution2 address\\
{\tt\small secondauthor@i2.org}
}
\author{
 Sheng Liu{\thanks{The first two authors contribute equally, order decided by coin flipping.}} $^{\,1}$~~Kangning Liu{\samethanks}  $^{\,1}$~~Weicheng Zhu$^{\,1}$~~Yiqiu Shen$^{\,1}$~~Carlos Fernandez-Granda$^{1,2}$\\
 $^{1}$ NYU Center for Data Science\\
 $^{2}$ NYU Courant Institute of Mathematical Sciences
}

\maketitle
\begin{abstract}
Deep learning in the presence of noisy annotations has been studied extensively in classification, but much less in segmentation tasks. In this work, we study the learning dynamics of deep segmentation networks trained on inaccurately annotated data. We observe a phenomenon that has been previously reported in the context of classification: the networks tend to first fit the clean pixel-level labels during an ``early-learning'' phase, before eventually memorizing the false annotations. However, in contrast to classification, memorization in segmentation does not arise simultaneously for all semantic categories. Inspired by these findings, we propose a new method for segmentation from noisy annotations with two key elements.  First, we detect the beginning of the memorization phase separately for each category during training. This allows us to adaptively correct the noisy annotations in order to exploit early learning. Second, we incorporate a regularization term that enforces consistency across scales to boost robustness against annotation noise. Our method outperforms standard approaches on a medical-imaging segmentation task where noises are synthesized to mimic human annotation errors. It also provides robustness to realistic noisy annotations present in weakly-supervised semantic segmentation, achieving state-of-the-art results on PASCAL VOC 2012.  \footnote{Code is available at \url{https://github.com/Kangningthu/ADELE}}

\end{abstract}

\section{Introduction}
\label{sec:Introduction}

\begin{figure}[ ]
\resizebox{1\linewidth}{!}{
     \resizebox{1\linewidth}{!}{
    \setlength{\fboxrule}{4pt} 
     \setlength{\fboxsep}{0cm}
        \begin{tabular}{c c c c c }
    \centering
        \Large Input &  \Large Ground Truth &  \Large  Baseline &  \Large Baseline+ADELE \\
        \centering
        \includegraphics[ width=0.35\linewidth]{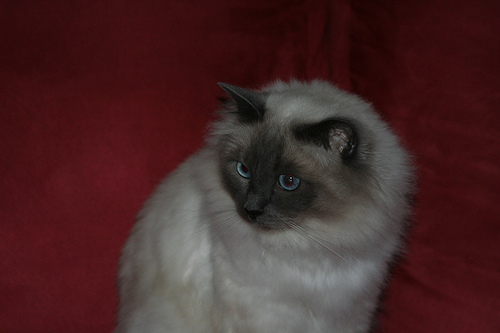}
        &   \includegraphics[ width=0.35\linewidth]{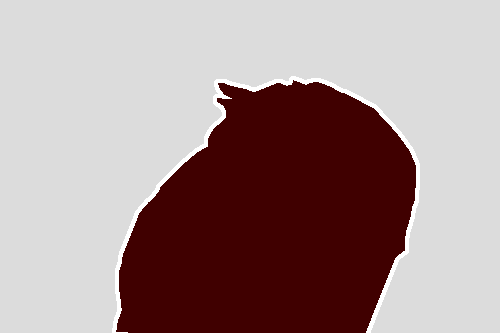}
         &   
         \includegraphics[ width=0.35\linewidth]{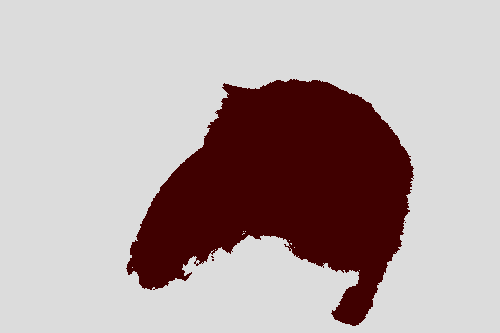} &                  \includegraphics[ width=0.35\linewidth]{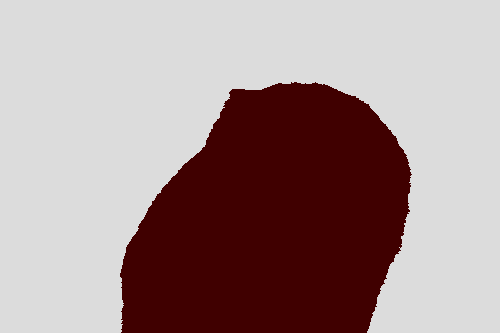}
\\
        \includegraphics[ width=0.35\linewidth]{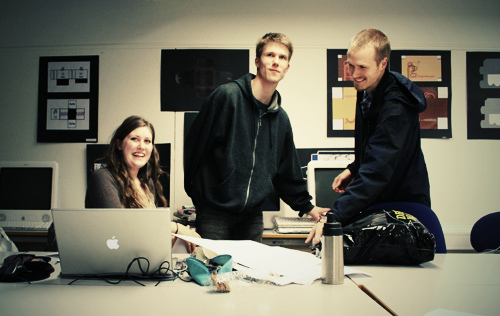}
        &   \includegraphics[ width=0.35\linewidth]{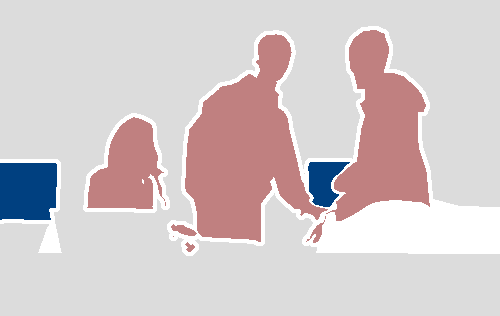}
         &   
         \includegraphics[ width=0.35\linewidth]{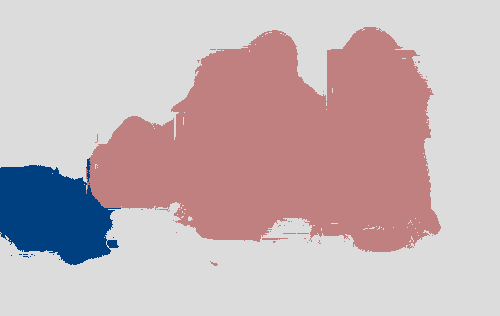} &                  \includegraphics[ width=0.35\linewidth]{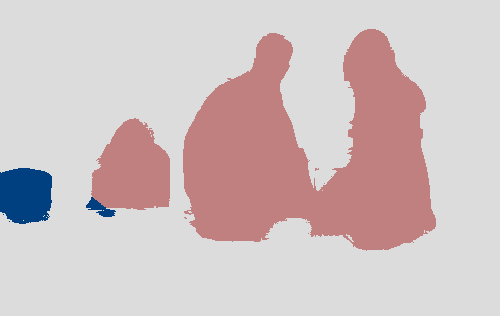}\\
        \includegraphics[ width=0.35\linewidth,trim={0 0 0px 0},clip]{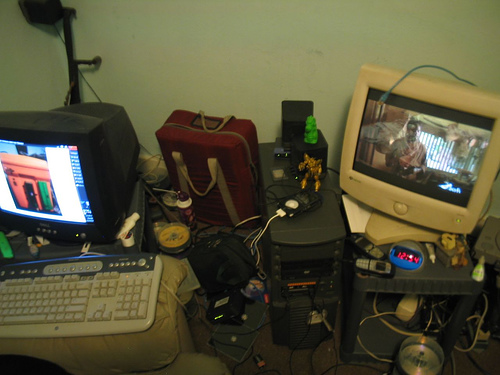}
         &   
         \includegraphics[ width=0.35\linewidth,trim={0 0 0 0},clip]{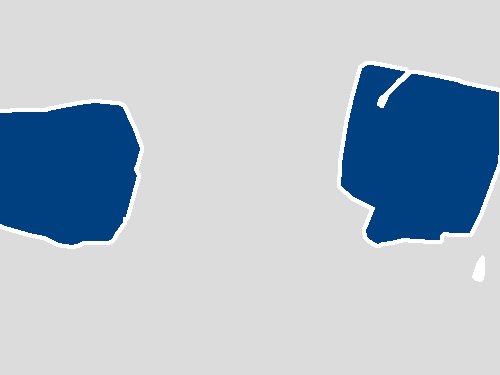} &                  \includegraphics[ width=0.35\linewidth,trim={0 0 0 0},clip]{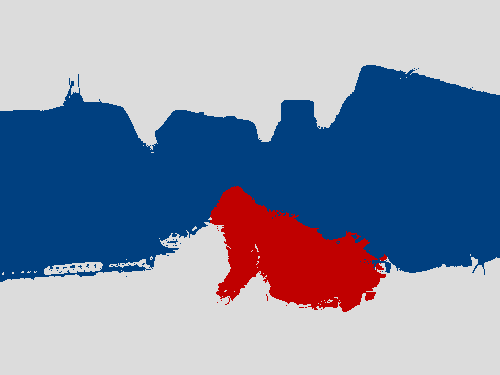}
        &   \includegraphics[ width=0.35\linewidth,trim={0px 0 0px 0},clip]{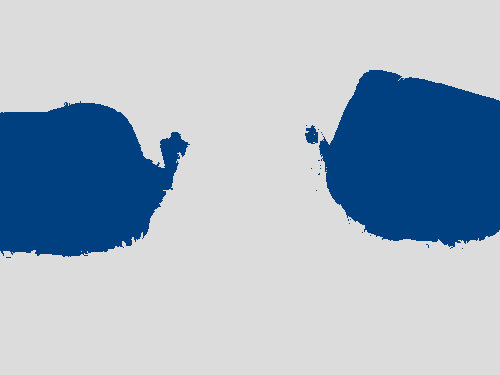}\\
        \includegraphics[ width=0.35\linewidth,trim={60px 0 40px 0},clip]{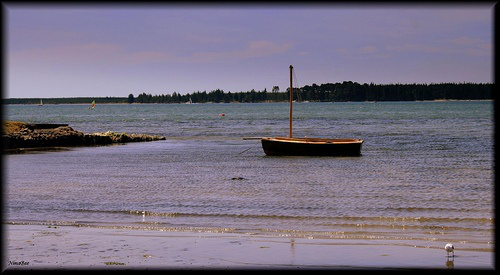}
         &   
         \includegraphics[ width=0.35\linewidth,trim={60px 0 40px 0},clip]{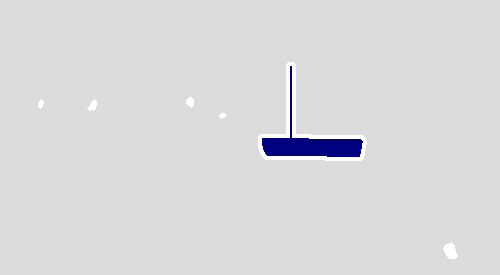} &                  \includegraphics[ width=0.35\linewidth,trim={60px 0 40px 0},clip]{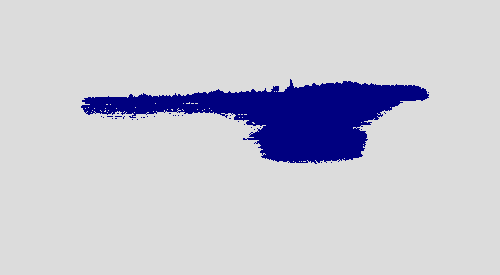}
        &   \includegraphics[ width=0.35\linewidth,trim={60px 0 40px 0},clip]{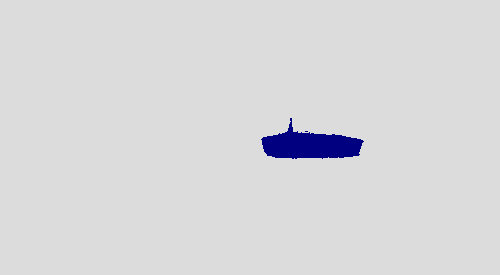}\\
     \end{tabular}  
     }
}
\centering
\caption{ Visualization of the segmentation results of the baseline method SEAM~\cite{wang2020self} and the baseline combined with the proposed ADaptive Early-Learning corrEction (ADELE). Our proposed ADELE improves segmentation quality.
More examples can be found in  Appendix~\ref{append:addfigures}.}
\label{fig:compare_fig_val}
\end{figure}

Semantic segmentation is a fundamental problem in computer vision. The goal is to assign a label to each pixel in an image, indicating its semantic category. Deep learning models based on convolutional neural networks (CNNs) achieve state-of-the-art performance~\cite{chen2017deeplab,zhao2017pyramid,sandler2018mobilenetv2,wang2018non}.
These models are typically trained in a supervised fashion, which requires pixel-level annotations. 
Unfortunately, gathering pixel-level annotations is very costly, and may require significant domain expertise in some applications~\cite{treml2016speeding,havaei2017brain,schlemper2018attention,liu2021weakly}. Furthermore, annotation noise is inevitable in some applications. For example, in medical imaging, segmentation annotation may suffer from inter-reader annotation variations~\cite{kats2019soft,zhang2020disentangling}.
Learning to perform semantic segmentation from noisy annotations is thus an important topic in practice.

Prior works on learning from noisy labels focus on classification tasks~\cite{liu2020early,Tanaka2018JointOF,xia2021robust}. There are comparatively fewer works on segmentation, where existing works focus on designing noise-robust network architecture~\cite{wang2020noise} or incorporating domain specific prior knowledge~\cite{shu2019lvc}. We instead focus on improving the performance in a more general perspective by studying the learning dynamics. We observe that the networks tend to first fit the clean annotations during an ``early-learning'' phase, before eventually memorizing the false annotations, thus jeopardizing generalization performance. This phenomenon has been reported in the context of classification~\cite{liu2020early}. However, this phenomenon in semantic segmentation differs significantly from its counterpart in classification in the following ways: 

\begin{itemize}[leftmargin=*]
    \item {The noise in segmentation labels is often spatially dependent. Therefore, it is beneficial to leverage spatial information during training.} 
    \item {In semantic segmentation, early learning and memorization do not occur simultaneously for all semantic categories due to pixel-wise imbalanced labels. Previous methods~\cite{li2020dividemix,liu2020early} in noisy label classification often assume class balanced data and thus either detecting or handling wrong labels for different classes at the same time. }
    \item {The annotation noise in semantic segmentation can be ubiquitous (all examples have some errors) while the state-of-the-art methods in classification~\cite{li2020dividemix,zhou2020robust,liu2020early} assume that some samples are completely clean.}  

\end{itemize}

\begin{figure}
    \centering
    \includegraphics[width=0.47\textwidth]{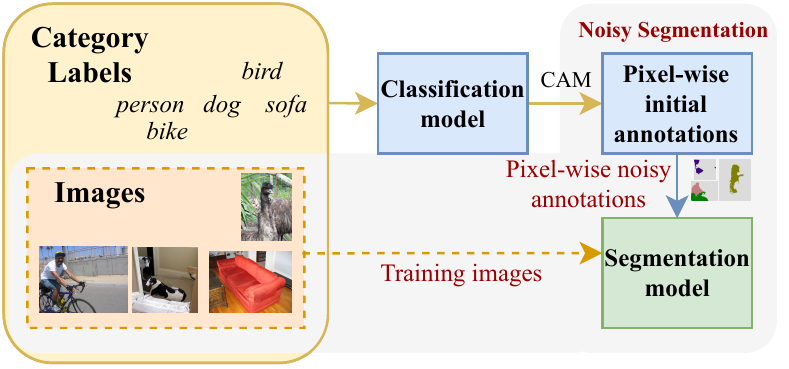}
    \caption{A prevailing pipeline for training WSSS. We aim to improve the segmentation model from noisy annotations.}
    \label{fig:WSSS}
\end{figure}

Inspired by these observations, we propose a new method, ADELE (ADaptive Early-Learning corrEction), that is designed for segmentation from noisy annotations. Our method detects the beginning of the memorization phase by monitoring the Intersection over Union (IoU) curve for each category during training. This allows it to adaptively correct the noisy annotations in order to exploit early-learning for individual classes. We also incorporate a regularization term to promote spatial consistency, which further improves the robustness of segmentation networks to annotation noise.

To verify the effectiveness of our method, we consider a setting where noisy annotations are synthesized and controllable. We also consider a practical setting -- Weakly-Supervised Semantic Segmentation (WSSS), which aims to perform segmentation based on weak supervision signals, such as image-level labels~\cite{kolesnikov2016seed,wei2016stc}, bounding box~\cite{dai2015boxsup,song2019box}, or scribbles~\cite{lin2016scribblesup}. We focus on a popular pipeline in WSSS. This pipeline consists of two steps (See Figure~\ref{fig:WSSS}). First, a classification model is used to generate pixel-level annotations. This is often achieved by applying variations of Class Activation Maps (CAM)~\cite{zhou2016learning} combined with post-processing techniques~\cite{krahenbuhl2011efficient,ahn2018learning}. Second, these pixel-level annotations are used to train a segmentation model (such as deeplabv1~\cite{chen2014semantic}). Generated by a classification model, the pixel-wise annotations supplied to the segmentation model are inevitably noisy, thus the second step is indeed a noisy segmentation problem. We therefore apply ADELE to the second step. In summary, our main contributions are:
\begin{itemize}[leftmargin=*]
\item We analyze the behavior of segmentation networks when trained with noisy pixel-level annotations. We show that the training dynamics can be separated into an early-learning and a memorization stage in segmentation with annotation noise. Crucially, we discover that these dynamics differ across each semantic category.

\item We propose a novel approach (ADELE) to perform semantic segmentation with noisy pixel-level annotations, which exploits early learning by adaptively correcting the annotations using the model output.
\item We evaluate ADELE on the thoracic organ segmentation task where annotations are corrupted to resemble human errors. ADELE is able to avoid memorization, outperforming standard baselines. We also perform extensive experiments to study ADELE on various types and levels of noises. 

\item ADELE achieves the state of the art on PASCAL VOC 2012 for WSSS. We show that ADELE can be combined with several different existing methods for extracting pixel-level annotations~\cite{ahn2018learning,wang2020self,fan2020learning} in WSSS, consistently improving the segmentation performance by a substantial margin.  
\end{itemize}

\begin{figure*}[htbp]
\centering
\includegraphics[width=1\textwidth]{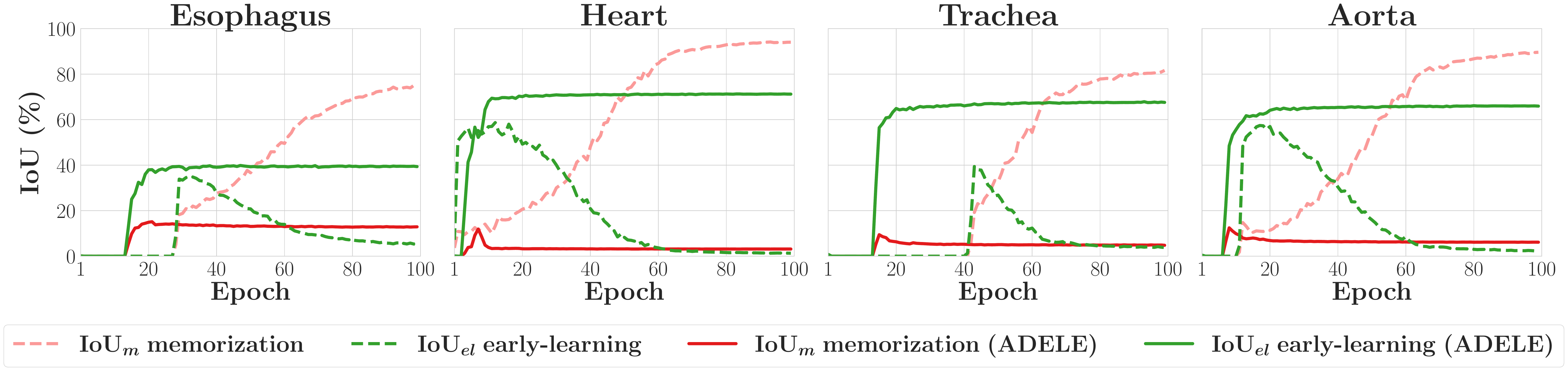}
  \smallskip
  \centering
\caption{We visualize the effect of early learning ($\text{IoU}_{el}$, green curves) and memorization ($\text{IoU}_{m}$, red curves) on \textit{incorrectly annotated pixels} with (solid lines) and without (dashed lines) ADELE for each foreground category of a medical dataset SegThor~\cite{lambert2020segthor}. The model is a UNet trained with noisy annotations that mimic human errors.
$\text{IoU}_{el}$ is the IOU between the model output and the ground truth computed over the incorrectly-labeled pixels. $\text{IoU}_{m}$ is the IOU between the model output and the incorrect annotations. For all classes, $\text{IoU}_m$ increases substantially as training proceeds  because the model gradually memorizes the incorrect annotations. This occurs at \textit{different speeds} for different categories. In contrast, $\text{IoU}_{el}$ first increases during an early-learning stage where the model learns to correctly segment the incorrectly-labeled pixels, but eventually decreases as memorization occurs. Like memorization, early-learning also happens at varying speeds for the different semantic categories. See Figure~\ref{fig:early_learning_voc} in Appendix for the plot on PASCAL VOC.
}
\label{fig:EL}
\end{figure*}

\begin{figure*}[t]
\centering
     \resizebox{0.9\linewidth}{!}{
    \setlength{\fboxrule}{4pt} 
     \setlength{\fboxsep}{0cm}
     \centering
\begin{tabular}{>{\centering\arraybackslash}m{0.16\linewidth} >{\centering\arraybackslash}m{0.16\linewidth} >{\centering\arraybackslash}m{0.16\linewidth} >{\centering\arraybackslash}m{0.16\linewidth} >{\centering\arraybackslash}m{0.16\linewidth} >{\centering\arraybackslash}m{0.16\linewidth}}
         Input  &   Ground truth &  Noisy annotations &  Model output after early learning &  Model output after memorization  &  Corrected annotations in ADELE \\
        \includegraphics[height=1.01\linewidth]{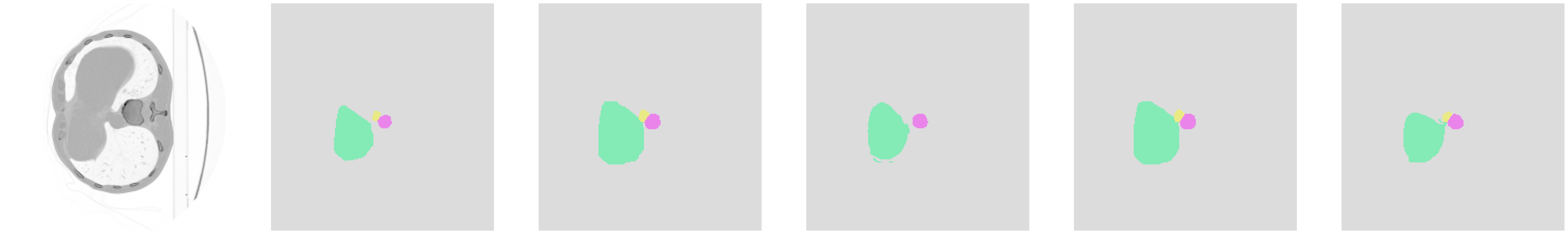} \\
        \includegraphics[height=1.01\linewidth]{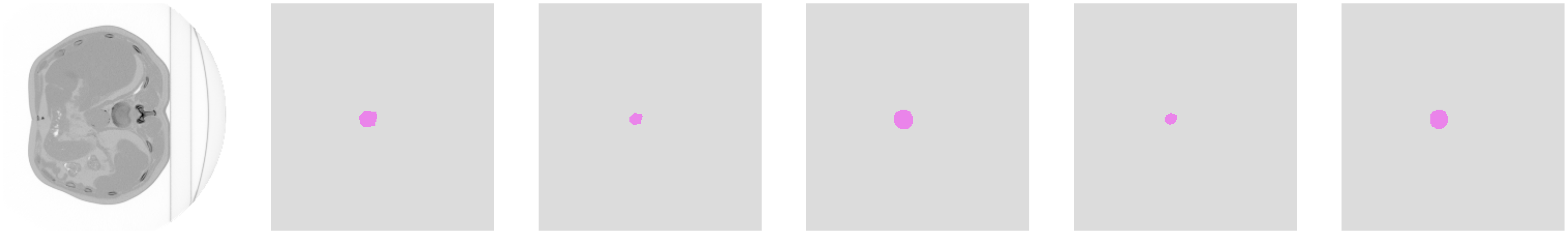} \\
        \midrule[0.7pt]
        \centering
         \includegraphics[ height=1\linewidth,trim={75px 0 50px 0},clip]{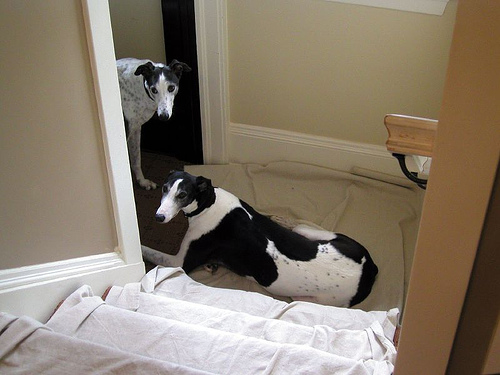}
         &         \includegraphics[ height=1\linewidth,trim={75px 0 50px 0},clip]{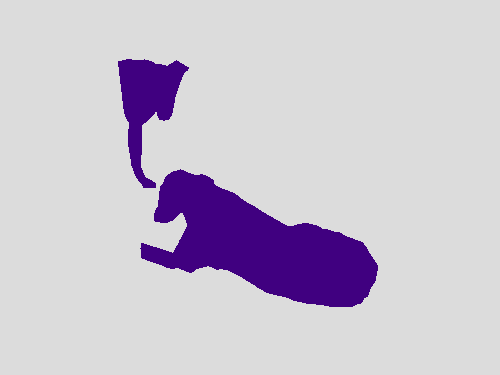} &    
         \includegraphics[ height=1\linewidth,trim={75px 0 50px 0},clip]{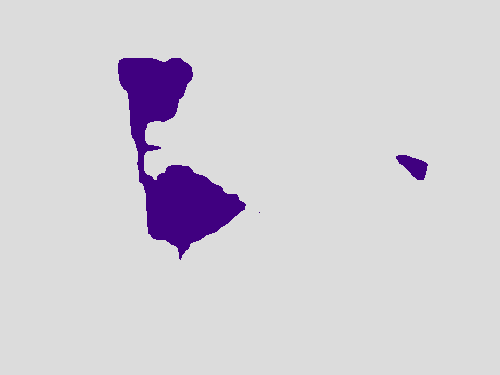} &         {{\includegraphics[ height=1\linewidth,trim={75px 0 50px 0},clip]{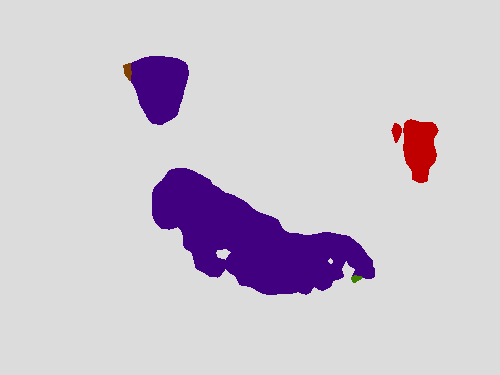}}}
        &   {{\includegraphics[ height=1\linewidth,trim={75px 0 50px 0},clip]{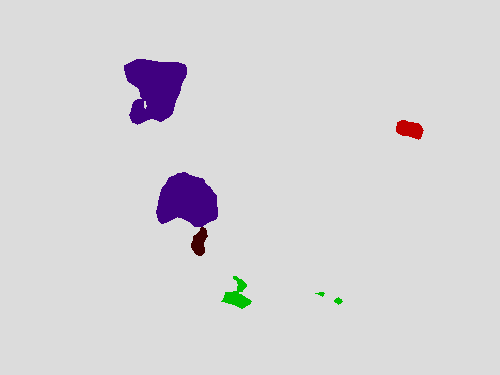}}
        } 
        &  
        \includegraphics[ height=1\linewidth,trim={75px 0 50px 0},clip]{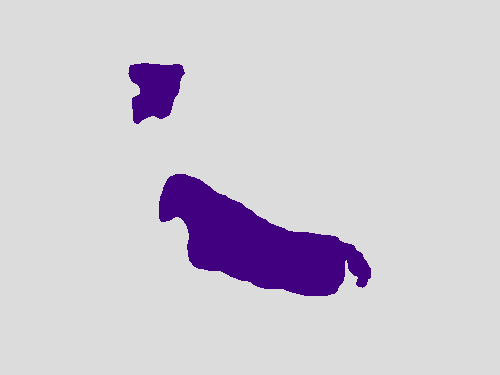}
         \\
        \includegraphics[ height=1\linewidth,trim={75px 0 50px 0},clip]{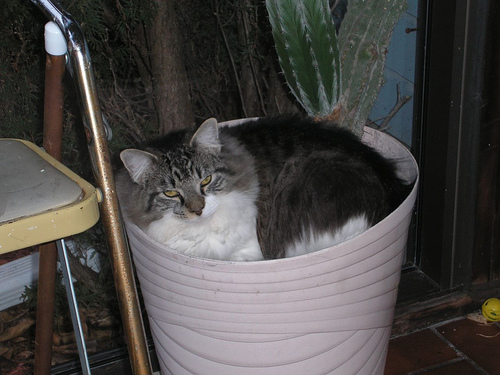}
         &   
              \includegraphics[ height=1\linewidth,trim={75px 0 50px 0},clip]{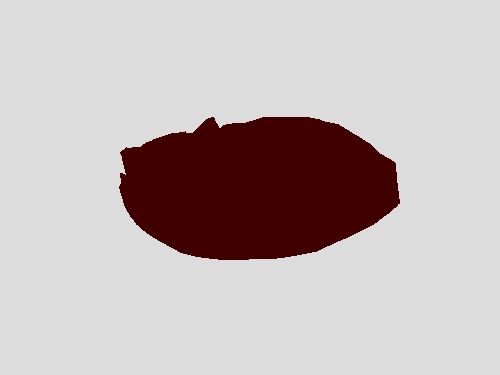} &
         \includegraphics[ height=1\linewidth,trim={75px 0 50px 0},clip]{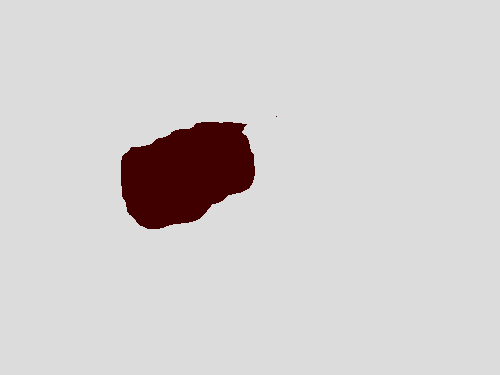} &            {{\includegraphics[ height=1\linewidth,trim={75px 0 50px 0},clip]{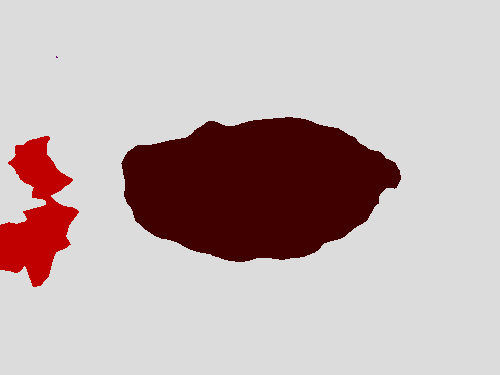}}}
        &    {{\includegraphics[ height=1\linewidth,trim={75px 0 50px 0},clip]{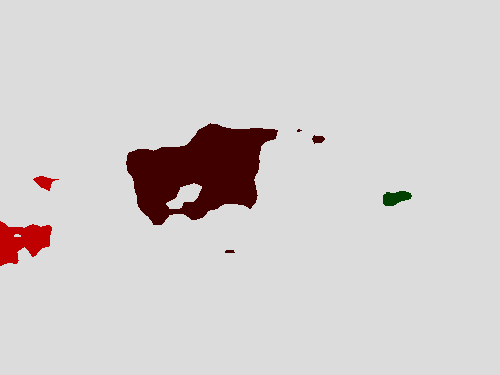}}
        } 
        &\includegraphics[ height=1\linewidth,trim={75px 0 50px 0},clip]{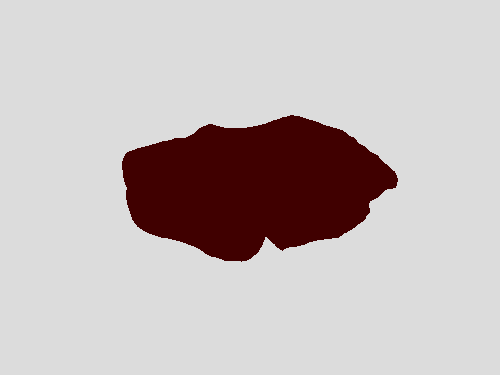}
        \\
        \includegraphics[ height=1\linewidth,trim={67px 0 100px 0},clip]{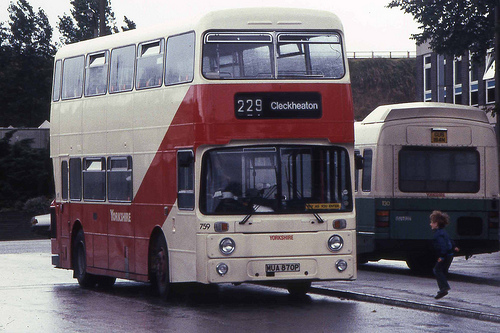}
         &          \includegraphics[ height=1\linewidth,trim={67px 0 100px 0},clip]{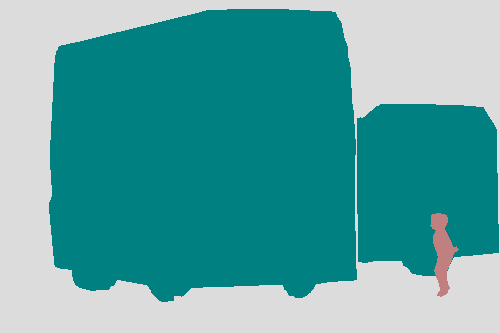} &  
         \includegraphics[ height=1\linewidth,trim={67px 0 100px 0},clip]{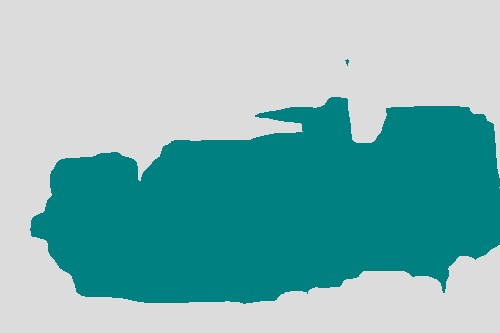} &    {{\includegraphics[ height=1\linewidth,trim={67px 0 100px 0},clip]{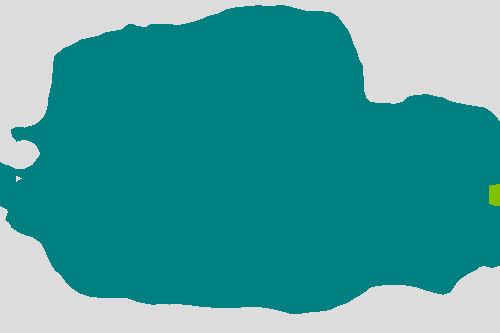}}}
        & {{\includegraphics[height=1\linewidth,trim={67px 0 100px 0},clip]{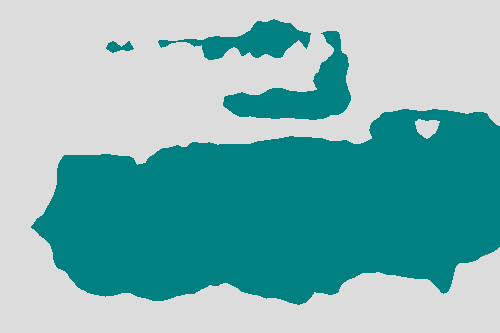}}}
        &\includegraphics[ height=1\linewidth,trim={67px 0 100px 0},clip]{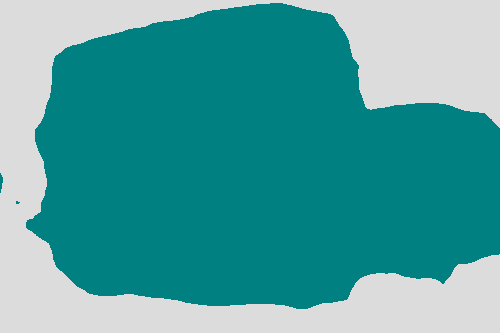}
         \\
         \includegraphics[
        height=1\linewidth,trim={75px 0 50px 0},clip]{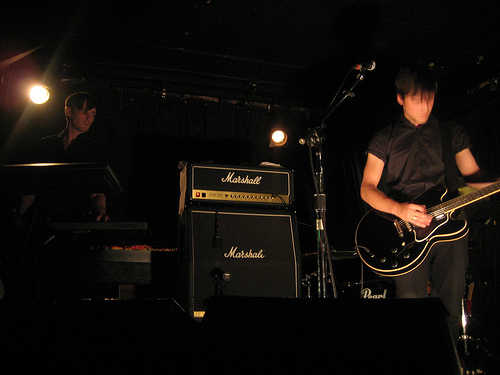}
      &          \includegraphics[ height=1\linewidth,trim={75px 0 50px 0},clip]{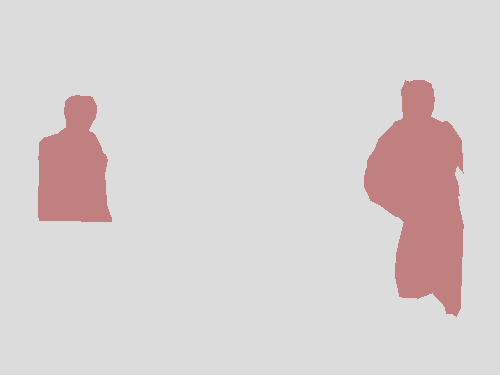}  &   
      \includegraphics[ height=1\linewidth,trim={75px 0 50px 0},clip]{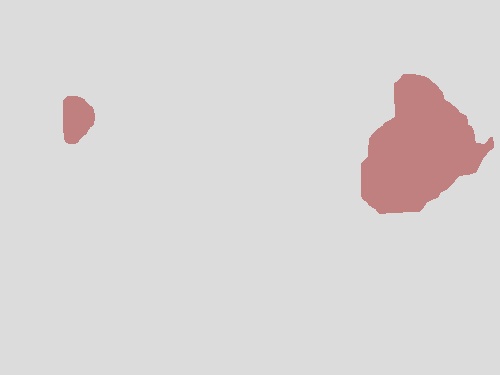}  &         {\includegraphics[ height=1\linewidth,trim={75px 0 50px 0},clip]{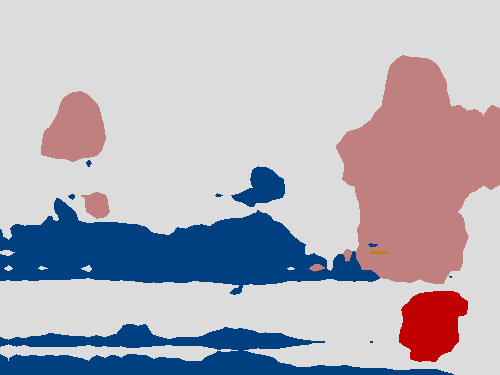}}
        &   {\includegraphics[ height=1\linewidth,trim={75px 0 50px 0},clip]{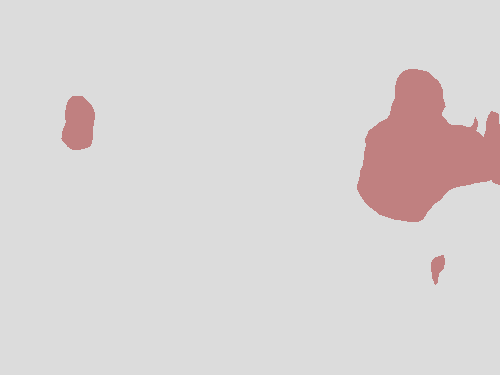}
        } 
        &\includegraphics[ height=1\linewidth,trim={75px 0 50px 0},clip]{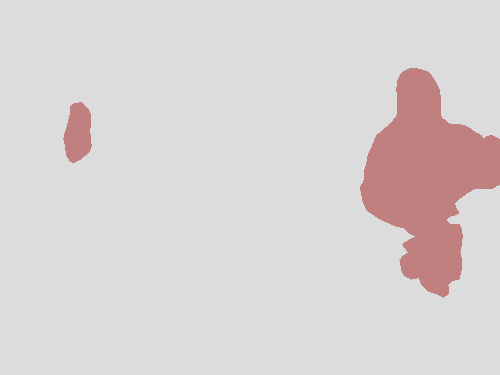}\\
     \end{tabular}
     }
\centering
\caption{Visual examples illustrating the early-learning and memorization phenomena. For several images in a medical dataset Segthor~\cite{lambert2020segthor} (top tow rows) and the WSSS dataset VOC 2012~\cite{everingham2015pascal} (bottom four rows), we show the ground-truth annotations (second column), noisy annotations (third column) obtained by a synthetic corruption process for the medical data and by the classification-based SEAM~\cite{wang2020self} model for WSSS, the output of a model segmentation model trained on the noisy annotations after early learning (fourth column), and the output of the same model after memorization (fifth column).  The model for the medical dataset is a UNet. The WSSS model is a standard DeepLab-v1 network trained with the SEAM annotations. As suggested by the graphs in Figure~\ref{fig:EL} after early learning the model corrects some of the annotation errors, but these appear again after memorization. ADELE is able to correct the labels leveraging the early learning output, thereby avoiding memorization (sixth column).
We set the background color to light gray for ease of visualization.
}
\label{fig:compare_fig2}
\end{figure*}

\section{Methodology }
\label{sec:method}
\subsection{Early learning and memorization in segmentation from noisy annotations}
In a typical classification setting with label noise, a subset of the images are incorrectly labeled. It has been observed in prior works that deep neural networks tend to first fit the training data with clean labels during an $\textit{early-learning}$ phase, before eventually $\textit{memorizing}$ the examples with incorrect labels~\cite{ arpit2017closer,liu2020early}. Here, we show that this phenomenon also occurs in segmentation when the available pixel-wise annotations are noisy (i.e. some of the pixels are incorrect). We consider two different problems. First, segmentation in medical imaging, where annotation noise is mainly due to human error. Second, the annotation noise in weakly-supervised semantic segmentation due to the bias of classification models, as they mostly focus on discriminative regions, and the post-processing errors may result in systematic over or under segmentation. 

Given noisy annotations for which \textbf{we know the ground truth}, we can quantify the early-learning and memorization phenomena by analyzing the model output \textbf{on the pixels that are incorrectly labeled}: 
\begin{itemize}[leftmargin=5.5mm]
\item \textbf{early learning $\text{IoU}_{el}$}: We quantify early learning using the overlap (measured in terms of the Intersection over Union (IoU) metric) between the outputs and the corresponding ground truth label on the pixels that are incorrectly labeled, denoted by $\text{IoU}_{el}$.
\item \textbf{memorization $\text{IoU}_m$}: We quantify memorization using the overlap (measured in IoU) between the CNN outputs and the incorrect labels, denoted by $\text{IoU}_m$.
\end{itemize}

Figure~\ref{fig:EL} demonstrates the phenomena of early-learning and memorization on a randomly corrupted CT-scan segmentation dataset (SegTHOR~\cite{lambert2020segthor}). We analyze the learning curve on the incorrectly-annotated pixels during the training process. The plots show the $\text{IoU}_m$ (dashed red line) and $\text{IoU}_{el}$ (dashed green line) at different training epochs. For all classes, the IoU between the output and the incorrect labels ($\text{IoU}_m$) increases substantially as training proceeds because the model gradually memorizes the incorrect annotations. This memorization process occurs at \textbf{varying speeds} for different semantic categories
(compare \emph{heart} and \emph{Aorts} with \emph{Traches} or \emph{Esophagus} in the SegThor dataset). The IoU between the output and the correct labels ($\text{IoU}_{el}$) follows a completely different trajectory: it first increases during an early-learning stage where the model learns to correctly segment the incorrectly-labeled pixels, but eventually decreases as memorization occurs (for the WSSS dataset, we observe a very similar phenomenon shown in Figure~\ref{fig:ELzoomin} in the Appendix). Like memorization, early-learning also happens at varying speeds for the different semantic categories.

Figure~\ref{fig:compare_fig2} illustrates the effect of early learning and memorization on the model output. In the medical-imaging application, the noisy annotations (third column) are synthesized to resemble human annotation errors which either miss or encompass the ground truth regions (compare to second column). Right after early learning, these regions are identified by the segmentation model (fourth column), but after memorization the model overfits to the incorrect annotations and \emph{forgets} how to segment these regions correctly (fifth column). Similar effects are observed in WSSS, in which the noisy annotations generated by the classification model are missing some object regions, perhaps because they are not particularly discriminative (e.g. the body of the dog, cat and people in the first, second, and fourth row respectively, or the upper half of the bus in the third row). The segmentation model first identify these regions but eventually overfits to the incorrect annotations. 
Our goal in this work is to modify the training of segmentation models on noisy annotations in order to prevent memorization. This is achieved by combining two strategies described in the next two sections. Figure~\ref{fig:EL} and Figure~\ref{fig:compare_fig2} shows that the resulting method substantially mitigates memorization (solid red lines) and promotes continued learning beyond the early-learning stage (solid green lines).

\begin{figure*}[t]
\centering
\resizebox{1\linewidth}{!}{
\begin{tabular}{>{\centering\arraybackslash}m{0.5\linewidth}  >{\centering\arraybackslash}m{0.5\linewidth} >{\centering\arraybackslash}m{0.5\linewidth}}
    & \Large    Esophagus  &  \Large Heart    \\
\includegraphics[height=0.57\linewidth]{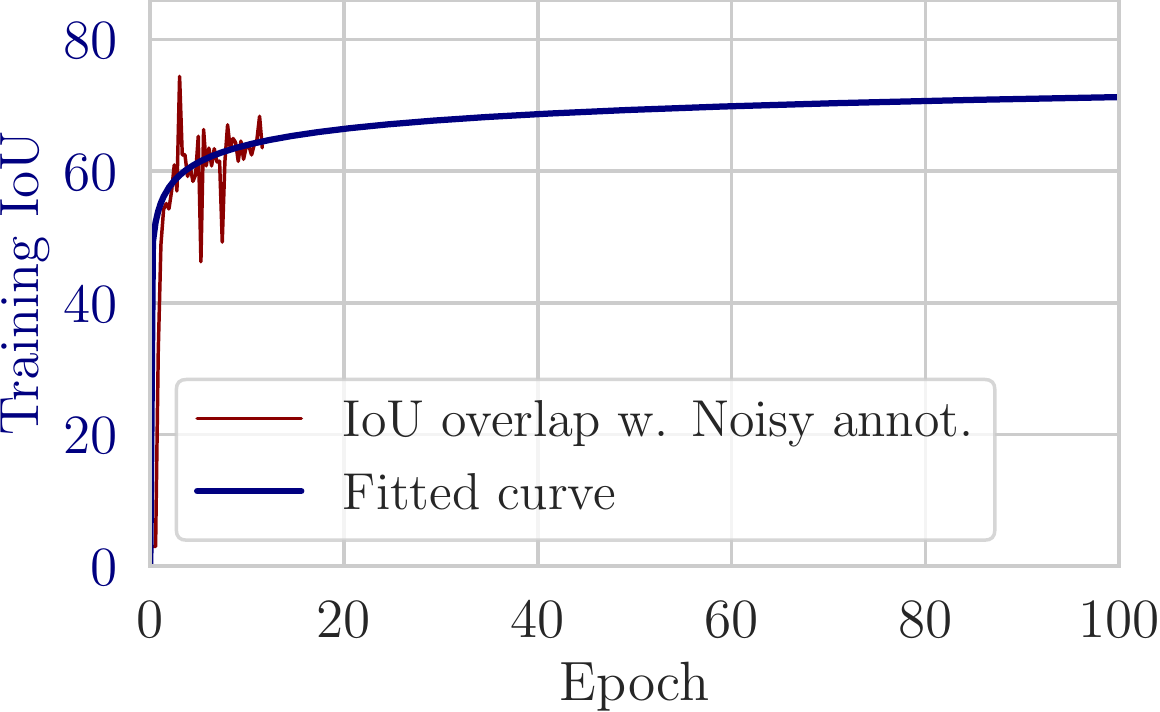} & \includegraphics[width=1\linewidth]{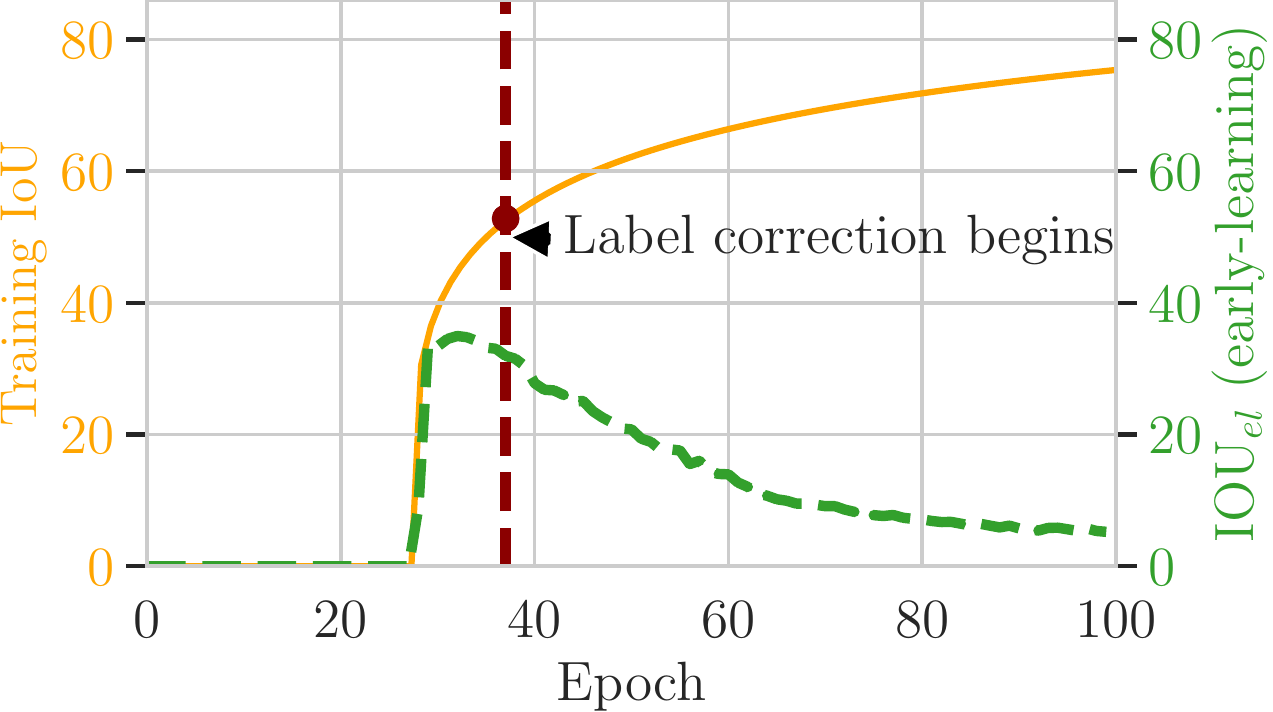} & \includegraphics[width=1\linewidth]{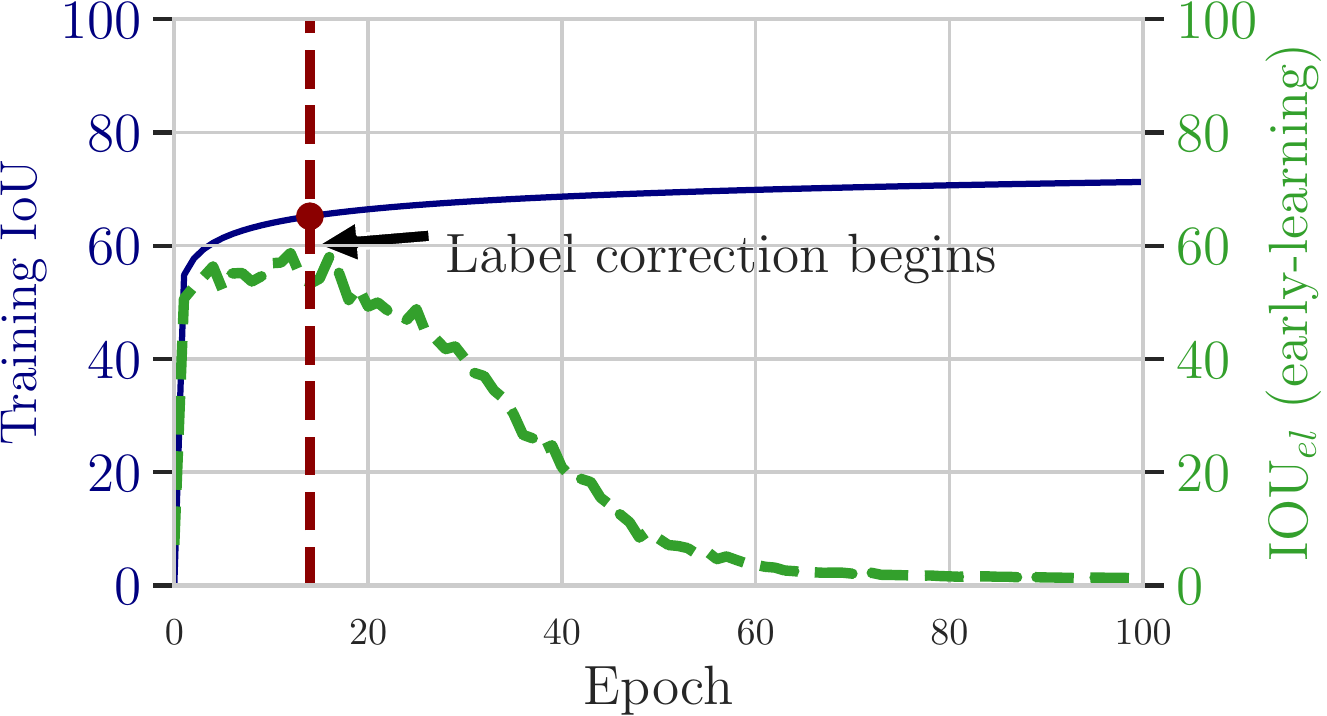} \\
      &  \Large Trachea & \Large Aorta   \\
\includegraphics[height=0.57\linewidth]{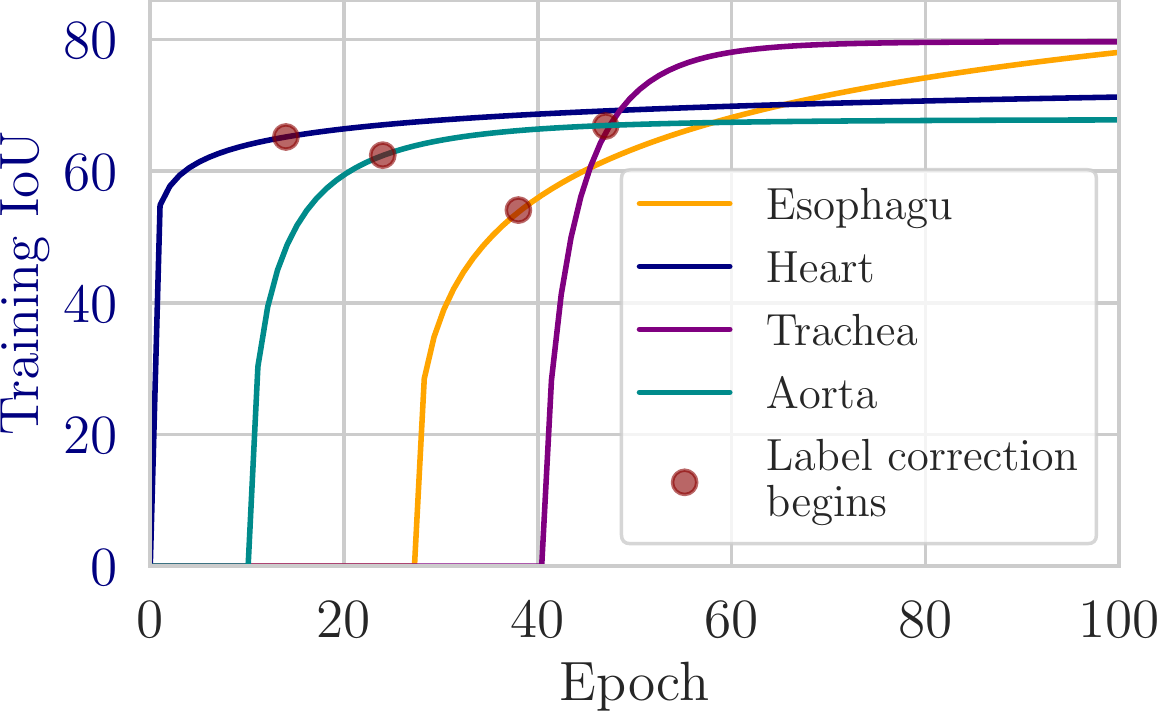}  &\includegraphics[width=1\linewidth]{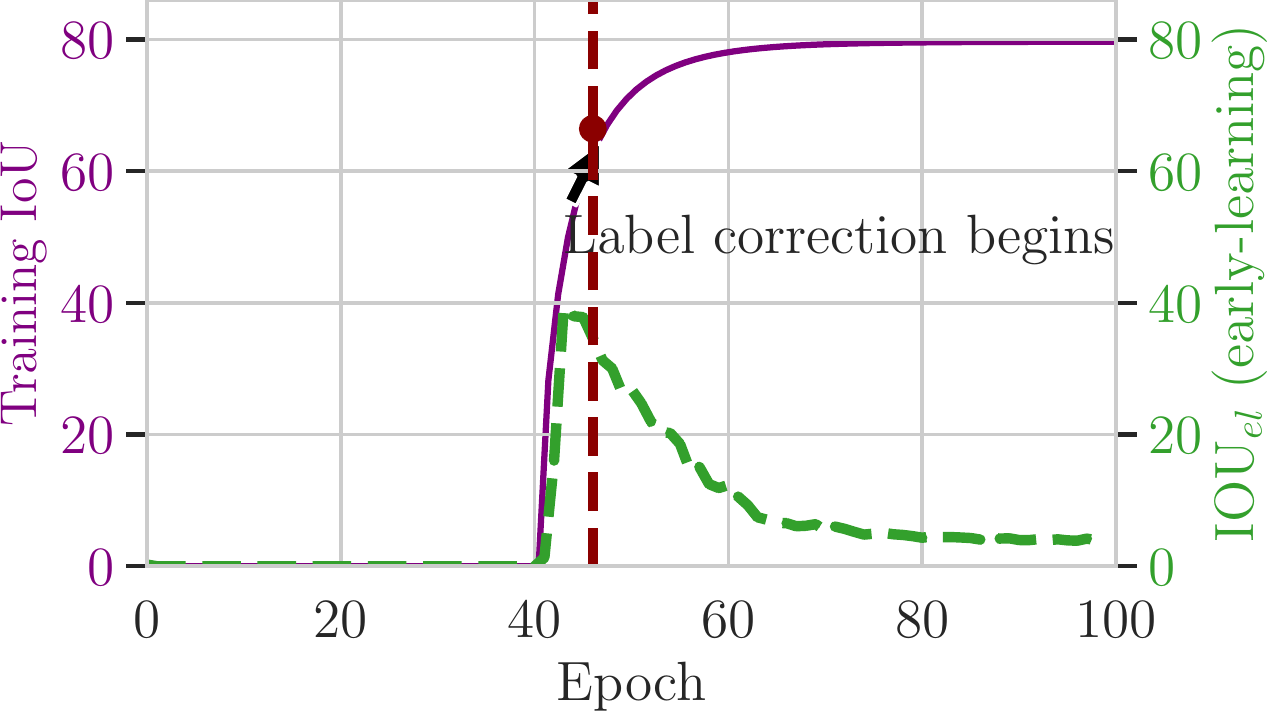}  &\includegraphics[width=1\linewidth]{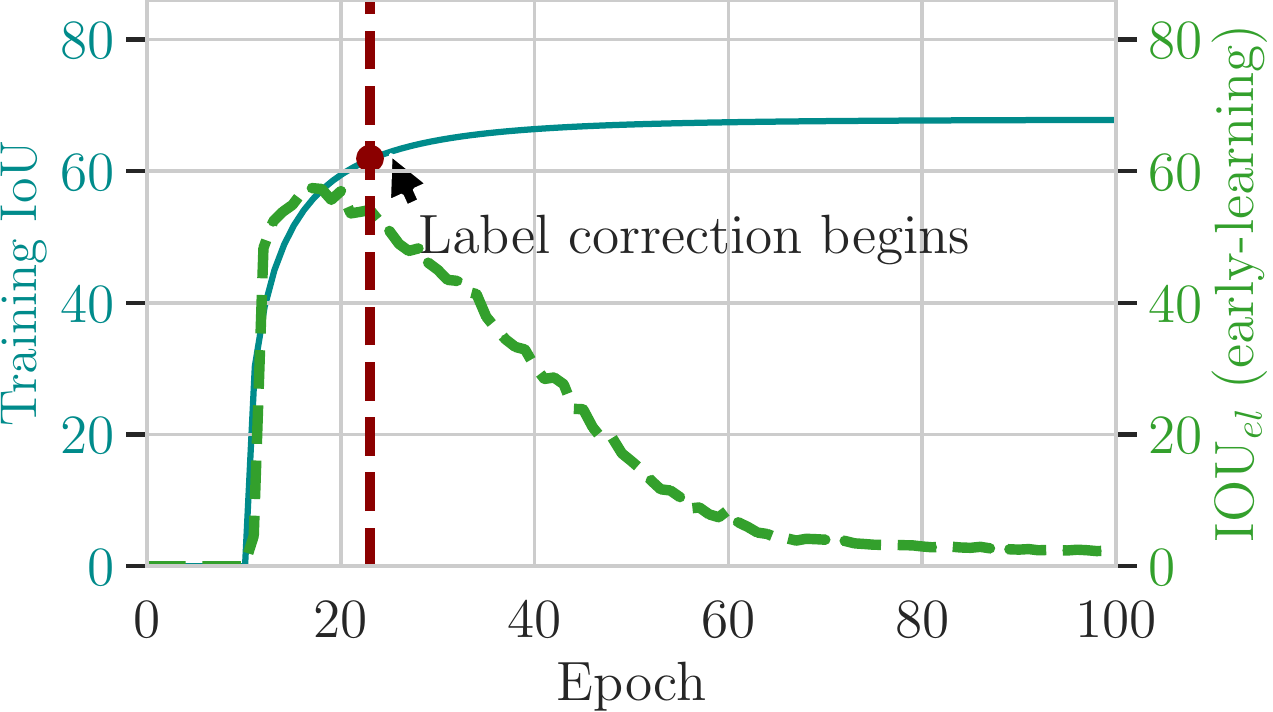} 
\end{tabular} 
}
\caption{Illustration of the proposed curve fitting method to decide when to begin label correction in ADELE (Results on SegThor). \textbf{First column:} On the top, we plot the IoU between the model predictions and the initial noisy annotations for the same model used in Figures~\ref{fig:EL} and~\ref{fig:compare_fig2} and the corresponding fit with the parametric model in Equation~\ref{eq: fitting_function}. The label correction beginning iteration is based on the relative slope change of the fitted curve. The bottom image shows the label correction times for different semantic categories, showing that they are quite different. 
\textbf{Second and third columns:} the green lines show the $\text{IoU}_{el}$ for different categories \emph{Esophagus}, \emph{Heart}, \emph{Trachea} and \emph{Aorta}. The $\text{IoU}_{el}$ equals the IoU between the model output and the ground truth computed over the incorrectly-labeled pixels, and therefore quantifies early-learning. The label correction begins close to the end of the early-learning phase, as desired. 
More result in section~\ref{append:addfigures} in Appendix shows that this also occurs for VOC 2012. 
}
\label{fig:curvefitting}
\end{figure*}

\subsection{Adaptive label correction based on early-learning}
\label{subsec:ADELEmethod}

The early-learning phenomenon described in the previous section suggests a strategy to enhance  segmentation models: correcting the annotations using the model output. Similar ideas have inspired works in classification with noisy labels~\cite{Tanaka2018JointOF,yi2019probabilistic,Reed2015TrainingDN,liu2020early}. However, different from the classification task where the noise is mainly sample-wise, \textbf{the annotation noise is ubiquitous across examples and distributed in a pixel-wise manner}.  There is a key consideration for this approach to succeed: the annotations cannot be corrected too soon, because this degrades their quality. Determining when to correct the pixel-level annotations using the model output is challenging for two reasons:
\begin{itemize}[leftmargin=5.5mm]

    \item {Correcting all classes at the same time can be sub-optimal.}
    \item {During training, we do not have access to the performance of the model on ground-truth annotations (otherwise we would just use them to train the model in the first place!)}.
\end{itemize}

To overcome these challenges we propose to update the annotations corresponding to different categories at different times by detecting when early learning has occurred and memorization is about to begin using the training performance of the model.

In our experiments, we observe that the segmentation performance on the training set (measured by the IoU between the model output and the noisy annotations) improves rapidly during early learning, and then much more slowly during memorization (see the rightmost graph in Figure~\ref{fig:curvefitting}). We propose to use this deceleration to decide when to update the noisy annotations. To estimate the deceleration we first fit the following exponential parametric model to the training IoU using least squares: 
\begin{align}
    f(t) = a \left(1- {e}^{-b \cdot t^c}\right),
    \label{eq: fitting_function}
\end{align}
where $t$ represents training time and $0 < a \leq 1$, $b \geq 0$, and $c \geq 0$ are fitting parameters.  
Then we compute the derivative $f'(t)$ of the parametric model with respect to $t$ at $t=1$ and at the current iteration.\footnote{The derivative is given by 
$ f'(t) = abc e^{-bt^c} t^{c-1}$.} For each semantic category, the annotations are corrected when the relative change in derivative is above a certain threshold $r$, i.e. when
\begin{align}
\label{eq:threshold_r}
    \frac{|f'(1) - f'(t)|}{|f'(1)|} > r,
\end{align}
which we set to 0.9, and at every subsequent epoch. We only correct annotations for which the model output has confidence above a certain threshold $\tau$, which we set to 0.8. A detailed description about the label correction  is attached in the Appendix \ref{append:implementationdetails}. 
As shown in Table~\ref{tab:abaltion_merged}, adaptive label correction based on early learning improves segmentation models in the medical-imaging applications and WSSS, both on its own and in combination with multiscale-consistency regularization. Figure~\ref{fig:compare_fig2} shows some examples of annotation corrections (rightmost column).

\subsection{Multiscale consistency}

\begin{figure*}
    \centering
    \begin{minipage}{0.4\linewidth}
     \includegraphics[width=1\linewidth]{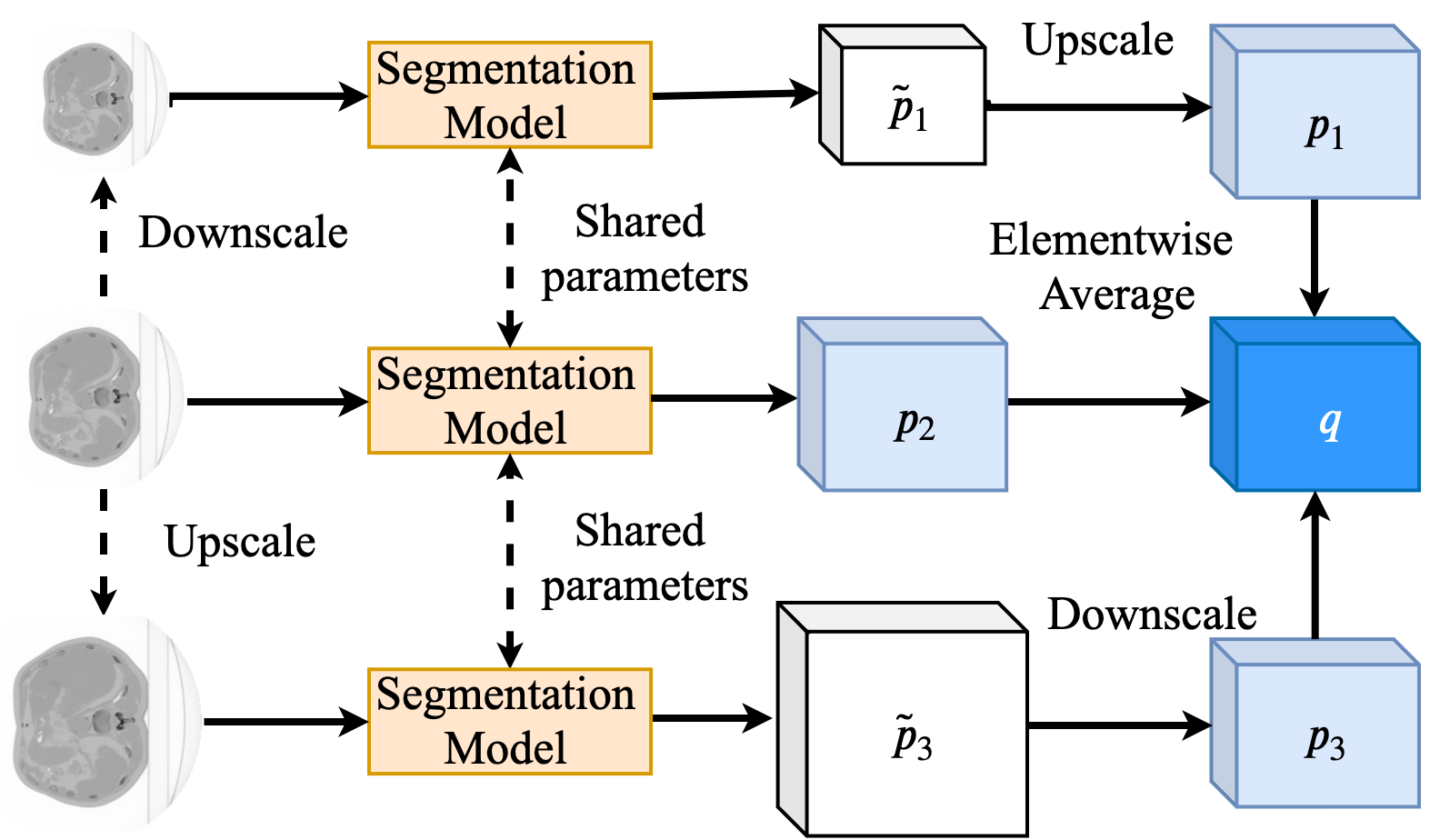}
    \end{minipage}
    \hspace{4mm}
\begin{minipage}{0.5\linewidth}
    \includegraphics[width=1\linewidth]{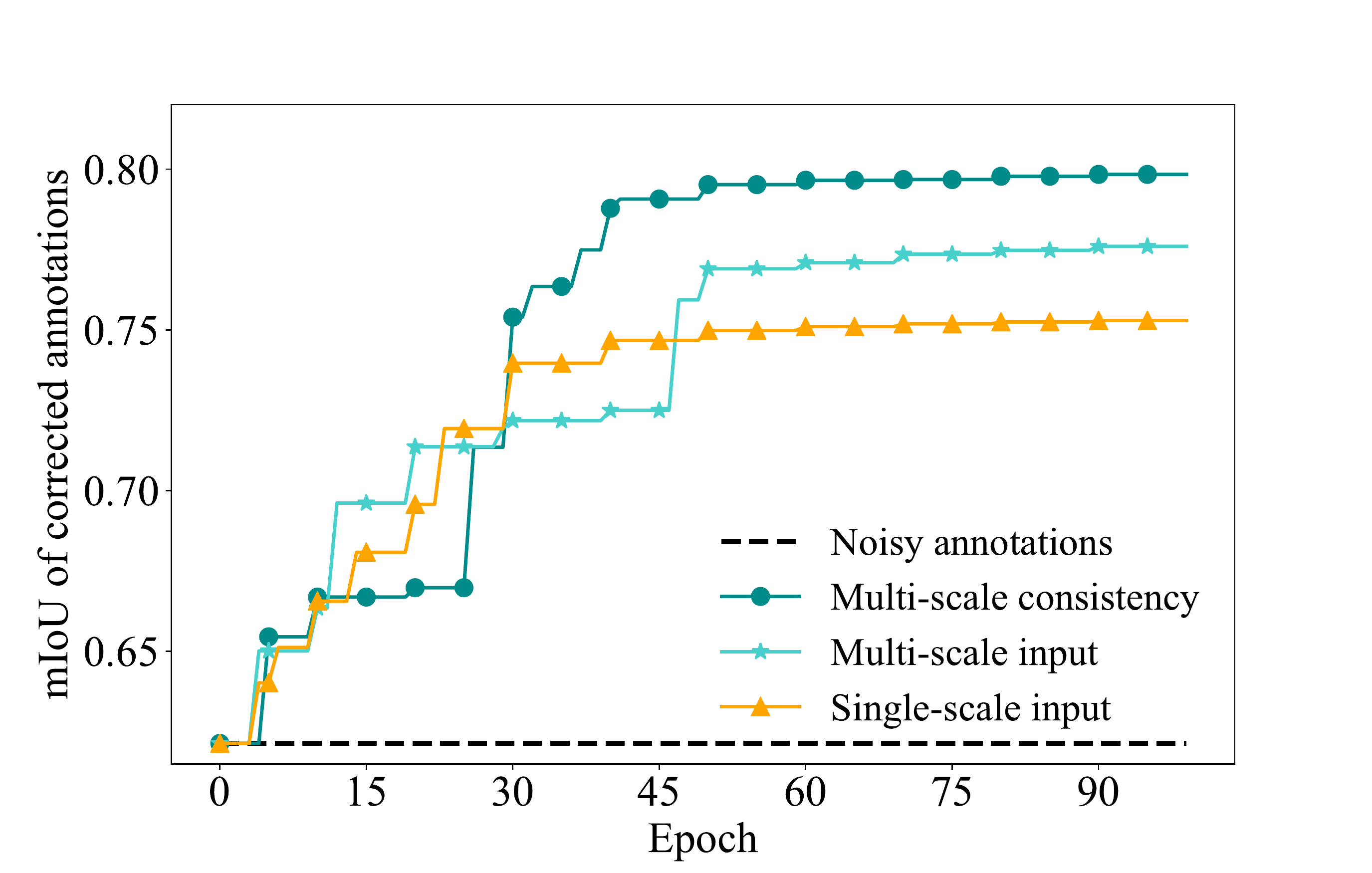}
    \end{minipage}
        \caption{\textbf{Left:} In the proposed multiscale-consistency regularization, rescaled copies of the same input (here upscaled $\times 1.5$ and downscaled $\times 0.7$) are fed into the segmentation model. The outputs ($\tilde{p}_1$, $p_2$ and $\tilde{p}_3$) are rescaled to have the same dimensionality ($p_1$, $p_2$ and $p_3$).  Regularization promotes consistency between these rescaled outputs and their elementwise average $q$. \textbf{Right:} Multi-scale consistency regularization 
       leads to more accurate corrected annotations (results on SegThor, results for VOC 2012 can be found in Figure \ref{fig:correction_quality_VOC}).}
    \label{fig:mutliscale}
\end{figure*}

As we previously mentioned, model outputs after early-learning are used to correct noisy annotations. Therefore, the quality of model outputs is crucial for the effectiveness of the proposed method. Following a common procedure that has shown to result in more accurate segmentation from the outputs~\cite{lin2018multi,yang2018denseaspp}, we average model outputs corresponding to multiple rescaled copies of inputs to form the final segmentation, and use them to correct labels. Furthermore, we incorporate a regularization that imposes consistency of the outputs across multi-scales and is able to make averaged outputs more accurate (See the right graph of Figure~\ref{fig:mutliscale}). This idea is inspired by consistency regularizations, a popular concept in the semi-supervised learning literature~\cite{laine2016temporal,tarvainen2017mean,miyato2018virtual,berthelot2019mixmatch,sohn2020fixmatch,french2019semi,kim2020structured} that encourages the model to produce predictions that are robust to arbitrary semantic-preserving spatial perturbations. In segmentation with noisy annotation, we introduce the consistency loss to provide an extra supervision signal to the network, preventing the network from only training on the noisy segmentation annotations, and overfitting to them. This regularization effect is also observed in the literature of classification with label noise~\cite{cheng2020learning,li2020dividemix}. Since our method uses the network predictions to correct labels, it is crucial to avoid overfitting to the noisy segmentation.

To be more specific, let $s$ be the number of scaling operations. In our experiments we set $s=3$ (downscaling $\times 0.7$, no scaling, and upscaling $\times 1.5$). We denote by $p_k(x)$, $1\leq k\leq s$, the model predictions  for an input $x$ rescaled according to these operations (see Figure~\ref{fig:mutliscale}). We propose to use a regularization term $\mathcal{L}_{\text{Multiscale}}$ to promote consistency between $p_k(x)$, $1\leq k \leq s$, and the average $q(x)=\frac{1}{s}\sum_{k=1}^{s}p_k(x)$:
\begin{align}
    \mathcal{L}_{\text{Multiscale}}(x) = 
    -\frac{1}{s}\sum_{k = 1}^s \text{KL}\left(p_{k} (x) \parallel  q(x) \right ),
\end{align}
where KL denotes the Kullback-Leibler divergence. The term is only applied to the input $x$ where the maximum entry of $q(x)$  is above a threshold $\rho$ (equal to 0.8 for all experiments). The regularization is weighted by a parameter $\lambda$ (set to one in all experiments) and then combined with a cross-entropy loss based on the available annotations. As shown in Tables~\ref{tab:abaltion_merged}, with multiscale consistency regularization, adaptive label correction further improves segmentation performance in both medical-imaging applications and the WSSS.

\section{Related work}

\noindent \textbf{Classification from noisy labels.} Early learning and memorization were first discovered in image classification from noisy labels~\cite{liu2020early}.  
Several methods exploit early learning to improve classification models by correcting the labels or adding regularization~\cite{Tanaka2018JointOF,yi2019probabilistic,Reed2015TrainingDN,liu2020early,xia2021robust}. 
Here we show that segmentation from noisy labels also exhibits early learning and memorization. However, these dynamics are different for different semantic categories. ADELE exploits this to perform correction in a class-adaptive fashion.

\noindent \textbf{Segmentation from noisy annotations.} 
Segmentation from noisy annotations is an important problem, especially in the medical domain~\cite{asman2012formulating}. 
Some recent works address this problem by explicitly taking into account systematic human labeling errors~\cite{ zhang2020disentangling}, and by modifying the segmentation loss to increase robustness~\cite{shu2019lvc,wang2020noise}. \cite{min2019two} propose to discover noisy gradient by collecting information from two networks connected with mutual attention. \cite{luo2021deep} shows that the network learns high-level spatial structures for fluorescence microscopy images. These structures are then leveraged as supervision signals to alleviate influence from wrong annotations. These methods mainly focus on improving the robustness by exploiting some setting-specific information (\emph{e.g.} network architecture, dataset, requiring some samples with completely clean annotation). In contrast, we propose to study the learning dynamics of noisy segmentation and propose ADELE, which performs label correction by exploiting early learning. 

\noindent \textbf{Weakly supervised semantic segmentation (WSSS).} 
Recent methods for WSSS~\cite{ahn2018learning,zhang2020reliability,fan2020learning} are mostly based on the approach introduced by Ref.~\cite{kolesnikov2016seed,wei2016stc}, where a classification model is first used to produce pixel-level annotations~\cite{zhou2016learning}, which are then used to train a segmentation model. 
These techniques mostly focus on improving the initial pixel-level annotations, by modifying the classification model itself~\cite{Wei_2017_CVPR,wei2018revisiting,li2018weakly,wang2020self}, or by post-processing these annotations~\cite{vernaza2017learning,ahn2018learning,ahn2019weakly}. 
However, the resulting annotations are still noisy
~\cite{zhang2020causal} (see Figure~\ref{fig:compare_fig2}). Our goal is to improve the segmentation model by adaptively accounting for this noise. Similar approach to our method has been observed in object detection where network outputs are dynamically used for training~\cite{jie2017deep}. In semantic segmentation,
the work that is most similar to our label-correction strategy is~\cite{huang2018weakly}, which is inspired by traditional \emph{seeded region-growing} techniques~\cite{adams1994seeded}. This method  estimates the foreground using an additional model~\cite{jiang2013salient}, and initializes the foreground segmentation estimate with classification-based annotations. This estimate is used to train a segmentation model, which is then used to iteratively update the estimate.  ADELE seeks to \emph{correct} the initial annotations, as opposed to \emph{growing them}, and does not need to identify the foreground estimate or an initial subset of highly-accurate annotations.

\section{Segmentation on Medical Images with Annotation Noise}
\label{sec:segthor}
Segmentation from noisy annotations is a fundamental challenge in the medical domain, where available annotations are often hampered by human error~\cite{zhang2020disentangling}. Here, we evaluate ADELE on a segmentation task where the goal is to identify organs from computed tomography images.

\noindent \textbf{Settings.} 
The dataset consists of 3D CT scans from the SegTHOR dataset~\cite{lambert2020segthor}. Each pixel is assigned to the esophagus, heart, trachea, aorta, or background. We treat each 2D slice of the 3D scan as an example, resizing to $256 \times 256$ pixels. We randomly split the slices into a training set of 3638 slices, a validation set of 570 slices, and a test set of 580 slices. Each patient only appears in one of these subsets.  
We generate annotation noise by applying random degrees of dilation and erosion to the ground-truth segmentation labels, mimicking common human errors~\cite{zhang2020disentangling} (see Figure~\ref{fig:compare_fig2}). In the main experiment, the noisy annotation is with a mIoU of 0.6 w.r.t the ground truth annotation. We further control the degree of dilation and erosion to simulate noisy annotation sets with different noise levels for testing the model robustness. 
We corrupt all annotations in the training set, but not in the validation and test sets. 
Our \textbf{evaluation metric} is Mean Intersection over Union (mIoU).

{
\centering
\centering
\resizebox{1.0\linewidth}{!}{
\begin{tabular}{ >{\centering\arraybackslash}m{0.2\linewidth}| >{\centering\arraybackslash}m{0.20\linewidth} >{\centering\arraybackslash}m{0.25\linewidth} >{\centering\arraybackslash}m{0.20\linewidth}}
\toprule
\centering
& Baseline & {\small ADELE w/o class adaptive} & ADELE \\
\hline
 Best val & 62.6$\pm$2.3  &	40.7$\pm$2.5&\textbf{71.1$\pm$0.7} \\
 Max test& {63.3$\pm$2.0} & 40.7$\pm$2.4 &\textbf{71.2$\pm$0.6}\\
Last Epoch& {59.1$\pm$1.3} &40.5$\pm$2.3 &	\textbf{70.8$\pm$0.7} \\
  \bottomrule
\end{tabular}
}
\captionof{table}{The mIoU ($\%$) comparison of the baseline and ADELE with or without class-adaptively correcting labels, on the test set of SegTHOR~\cite{lambert2020segthor}. We report the test mIoU of the model that performs best on the validation set (\textbf{Best Val}), the test mIoU at the last epoch (\textbf{Last Epoch}), and the highest test performance during training (\textbf{Max Test}). We report the mean and standard deviation after training the model with five realizations of the noisy annotations.}
\label{Table:segTHORtestwerrorbar1}
}

\begin{figure}
\centering
\includegraphics[width=0.45\textwidth]{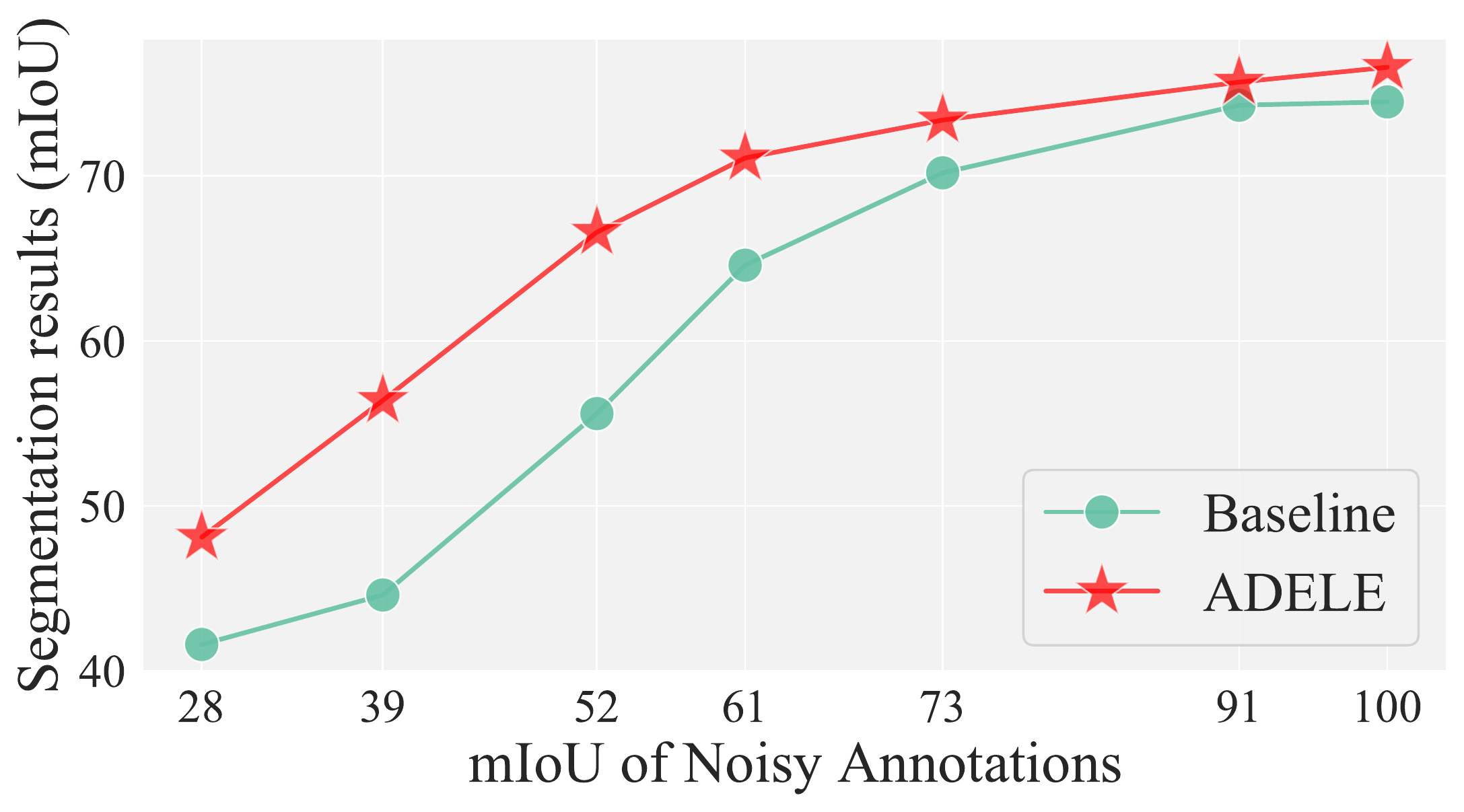}
\captionof{figure}{ The performance comparison of the baseline and ADELE on the test set of SegTHOR~\cite{lambert2020segthor}. The model is trained on noisy annotations with various levels of corruption (measured in mIoU with the clean ground truth annotations). ADELE is able to improve the model performance across a wide range of corruption levels.}
\label{fig:noise_level_baselines}
\end{figure}

\begin{table*}[h]
\resizebox{1.0\linewidth}{!}{
\begin{tabular}{>{\centering\arraybackslash}m{0.13\linewidth} | >{\centering\arraybackslash}m{0.12\linewidth} >{\centering\arraybackslash}m{0.18\linewidth} >{\centering\arraybackslash}m{0.24\linewidth}|
>{\centering\arraybackslash}m{0.12\linewidth} >{\centering\arraybackslash}m{0.18\linewidth} >{\centering\arraybackslash}m{0.24\linewidth}}
\toprule
&  \multicolumn{3}{c|}{SegTHOR} & \multicolumn{3}{c}{PASCAL VOC 2012} \\
\midrule[0.7pt]
 Label correction & {Single scale} & {Multiscale input augmentation} & {Multiscale consistency regularization } &{Single scale} & {Multiscale input augmentation} & {Multiscale consistency regularization }\\
\midrule[0.7pt]
 \xmark & 58.8 & 60.7 & 62.5 & 64.5 & 65.5 & 66.7 \\
 \cmark & {65.2} & 69.8 &\textbf{72.2} & {65.6} & 67.3 &\textbf{69.3}  \\
 \bottomrule
\end{tabular}
}
\caption{Ablation study for ADELE on  SegTHOR~\cite{lambert2020segthor} and PASCAL VOC 2012~\cite{everingham2015pascal}.
We report the mIoU achieved at the last epoch on the validation set for both dataset. Class-adaptive label correction mechanism achieves the best performance when combined with multi-scale consistency regularization. }
\label{tab:abaltion_merged}
\end{table*}

  \begin{table*}[ht]
 \centering
\resizebox{\linewidth}{!}{
\begin{tabular}{>{\centering\arraybackslash}m{0.16\linewidth}| >{\centering\arraybackslash}m{0.16\linewidth}| >{\centering\arraybackslash}m{0.16\linewidth}| >{\centering\arraybackslash}m{0.16\linewidth}| >{\centering\arraybackslash}m{0.16\linewidth}| >{\centering\arraybackslash}m{0.16\linewidth}| >{\centering\arraybackslash}m{0.16\linewidth}|
>{\centering\arraybackslash}m{0.16\linewidth}| >{\centering\arraybackslash}m{0.16\linewidth}|
>{\centering\arraybackslash}m{0.16\linewidth}| >{\centering\arraybackslash}m{0.16\linewidth}}
		\toprule	
		    &\multicolumn{7}{c!{\vrule width 1.6pt}} { \Large Previous methods} & \multicolumn{3}{c} {\Large \textbf{ADELE} + }\\
		   \cmidrule{2-7} \cmidrule{8-11}	
			  & DSRG~\cite{huang2018weakly} &
			  ICD~\cite{fan2020learning} &SCE~\cite{chang2020weakly} & AffinityNet~\cite{ahn2018learning}  & SSDD~\cite{shimoda2019self}&	
			 SEAM~\cite{wang2020self}    &
			  CONTA~\cite{zhang2020causal} & AffinityNet~\cite{ahn2018learning} &  SEAM~\cite{wang2020self}  & ICD~\cite{fan2020learning}  \\
			  \midrule			
		  & \multicolumn{3}{c|}{\Large ResNet-101} & \multicolumn{7}{c}{\Large ResNet-38}\\
		 \midrule	
		 {\Large Val}    & {\Large 61.4} & {\Large 64.1} & {\Large \color{blue} {66.1}} &{\Large 61.7}  & {\Large 64.9} & {\Large 64.5} & {\Large \color{blue} {66.1}} & 
        \Large 64.8
		 &{\Large \color{red} \textbf{69.3}} &{\Large 68.6}\\
		 \midrule	
		 {\Large Test}   & {\Large 63.2} & {\Large 64.3} & {\Large{ 65.9}} &{\Large 63.7} & {\Large 65.5} & {\Large 65.7} & {\Large \color{blue} {66.7}} &  
        {\Large 65.5}
		 &{\Large 68.8} &{\Large \color{red} \textbf{68.9}} \\
		\bottomrule
	\end{tabular}
	}
	     \caption{Comparison with state-of-the-art methods on the Pascal VOC 2012 dataset using mIoU ($\%$). The {\color{red} \textbf{best }} and {\color{blue} the best previous method} performance under each set are highlighted in red and blue respectively. The version of CONTA~\cite{zhang2020causal} reported here is deployed combined with SEAM~\cite{wang2020self}. The results clearly show that ADELE outperforms other approaches. }
    \label{Table:comparewSoTA}
\end{table*}

\noindent \textbf{Results.} For a fair comparison, we choose a UNet trained with multi-scale inputs as our \textbf{baseline}. We report the mIoU of the baseline and ADELE on the test set of SegTHOR dataset in Table~\ref{Table:segTHORtestwerrorbar1}. ADELE outperforms the baseline method at all three evaluation epochs. Moreover, correcting labels at the same time for all classes will have a detrimental effect on the performance. 

\noindent \textbf{Impacts of noise levels.} Figure~\ref{fig:noise_level_baselines} provides empirical evidence that ADELE is robust to a wide range of noises. The mIoU of noisy annotations (x-axis) indicates the correctness of the noisy annotations. Thus the smaller the mIoU shows the higher noise levels. The improvements achieved by ADELE are substantial when the noise levels are moderate. 

\noindent \textbf{Ablation study for each part of ADELE.} We perform an ablation study to understand how different parts of ADELE contribute to the final performance. From Table~\ref{tab:abaltion_merged}, we observe that the model trained with multiple rescaled versions of the input (illustrated in left graph of Figure~\ref{fig:mutliscale}) performs better than the model trained only with the original scale of the input. The proposed spatial consistency regularization further improves the performance. Most importantly, combining any of these methods with label correction would substantially improve the performance. ADELE, which combines label correction with the proposed regularization, achieves the best performance. We also include ablation studies for the hyperparameters $r$, $\tau$ and $\rho$ in Appendix~\ref{append:additionalablation}. 
Additional segmentation results are provided in Appendix~\ref{append:addfigures}.

\section{Noisy Annotations in Weakly-supervised Semantic Segmentation}
\label{Sec:ExpWSSS}
We adopt a prevailing pipeline for training WSSS (described in detail in Section~\ref{sec:Introduction}), in which some pixel-wise annotations are generated using image level labels to supervise a segmentation network. These pixel-wise annotations are noisy. Therefore, we apply ADELE to this WSSS pipeline.  

We evaluate ADELE on a standard WSSS dataset -- PASCAL VOC 2012 \cite{everingham2015pascal}, which has 21 annotation classes (including background), and contains 1464, 1449 and 1456 images in the \emph{training}, \emph{validation (val)} and \emph{test} sets respectively. Following~\cite{shimoda2019self,zhang2020reliability,wang2020self,zhang2020causal,sun2020mining,yao2021non}, we use an augmented training set with 10582 images with annotations from~\cite{hariharan2011semantic}.

\noindent \textbf{Baseline Models.} To demonstrate the broad applicability of our approach, we apply ADELE using pixel-level annotations generated by three popular WSSS models: AffinityNet~\cite{ahn2018learning}, SEAM~\cite{wang2020self} and ICD~\cite{fan2020learning}, which do not rely on external datasets or external saliency maps. The annotations are produced by a classification model combined with the post-processing specified in~\cite{ahn2018learning,wang2020self,fan2020learning}. We provide details on the training procedure in Section~\ref{append:implementationdetails} in the Appendix. We use the same inference pipeline as SEAM~\cite{wang2020self}, which includes multi-scale inference~\cite{ahn2018learning,wang2020self,fan2020learning, zhang2020splitting} and CRF~\cite{krahenbuhl2011efficient}.

\noindent \textbf{Comparison with the state-of-the-art.} 
Table \ref{Table:comparewSoTA} compares the performance of the proposed method ADELE to state-of-the-art WSSS methods on PASCAL VOC 2012. ADELE improves the performance of AffinityNet~\cite{ahn2018learning}, SEAM~\cite{wang2020self} and ICD~\cite{fan2020learning} substantially on the validation and test sets. Moreover, ADELE combined with SEAM~\cite{wang2020self} and ICD~\cite{fan2020learning} achieves state-of-the-art performance on both sets. Although it uses only image-level labels, ADELE outperforms  state-of-the-art methods ~\cite{jiang2019integral,zhang2020splitting,sun2020mining,yao2021non} that rely on external saliency models~\cite{jiang2013salient}. To show that our method is complementary with other more advanced WSSS methods,  we have conducted an experiment with a recent WSSS method NSROM~\cite{yao2021non}, which uses external saliency models. ADELE+NSROM achieves mIoU of 71.6 and 72.0 on the validation and test set respectively, which is the SoTA for WSSS with ResNet segmentation backbone (see Appendix~\ref{Append:addtable}).

Figure~\ref{fig:perclasscomparision} compares the performance of SEAM and the performance of ADELE combined with SEAM on the validation set separately for each semantic category.
ADELE improves performance for most categories, with the exception of a few categories where the baseline model does not perform well 
(\eg chair, bike). On Figure~\ref{fig:compare_fig_val} and~\ref{fig:compare_fig_val_fail}, we show some qualitative segmentation results from the validation set. Figure~\ref{fig:compare_fig_val} shows examples where ADELE successfully improves the SEAM segmentation. Figure~\ref{fig:compare_fig_val_fail} shows examples where it does not. In both the output of SEAM has highly structured segmentation errors: the prediction encompasses the bike but completely fails to capture its inner structure, and the chair is missclassified as a sofa. This supports the conclusion that ADELE provides less improvement when the baseline method performs poorly.

\begin{figure}
    \centering
\includegraphics[width=1\linewidth,trim={80 0 80 5}, clip]{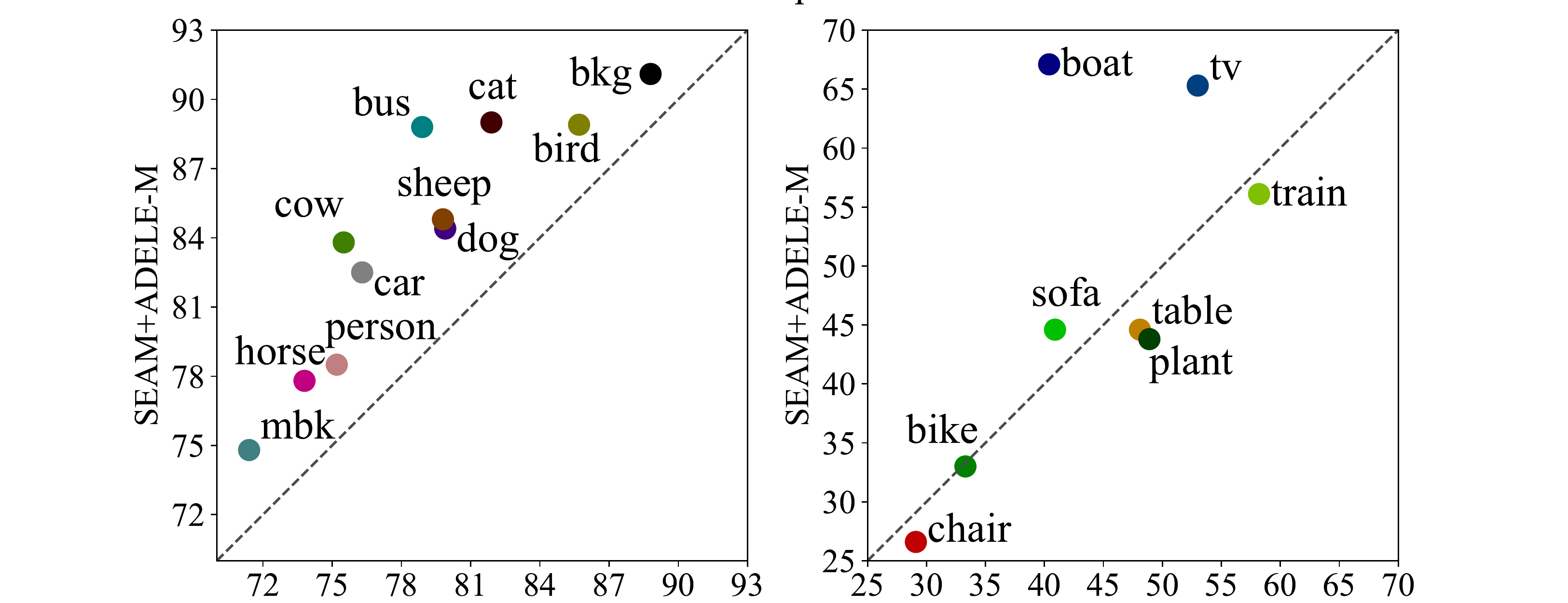}
    \caption{Category-wise comparison of the IoU ($\%$) of SEAM~\cite{wang2020self} and SEAM combined with the proposed method ADELE on the validation set of PASCAL VOC 2012. We separate the categories based on IoUs for better visualization. }
    \label{fig:perclasscomparision}
\end{figure}
\begin{figure}[]
\resizebox{1\linewidth}{!}{
     \resizebox{1\linewidth}{!}{
    \setlength{\fboxrule}{4pt} 
     \setlength{\fboxsep}{0cm}
        \begin{tabular}{c c c c c }
    \centering
        \Large Input &  \Large Ground Truth &  \Large  SEAM &  \Large SEAM+ADELE \\
        \centering
        \includegraphics[ width=0.35\linewidth]{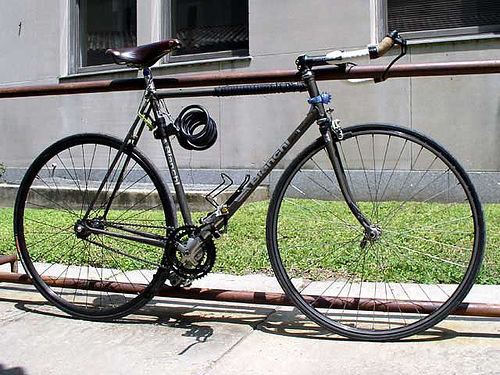}
        &   \includegraphics[ width=0.35\linewidth]{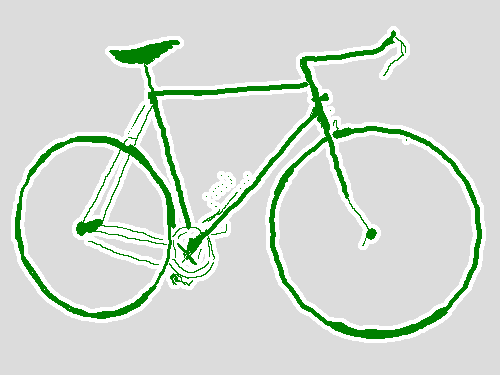}
         &   
         \includegraphics[ width=0.35\linewidth]{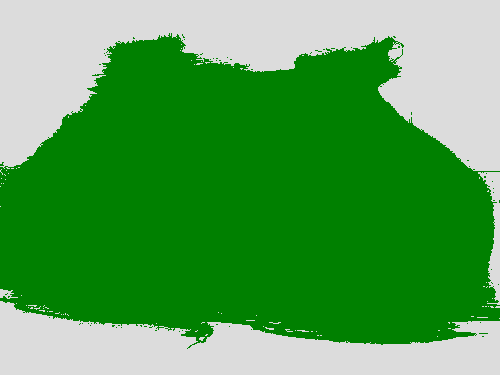} &                  \includegraphics[ width=0.35\linewidth]{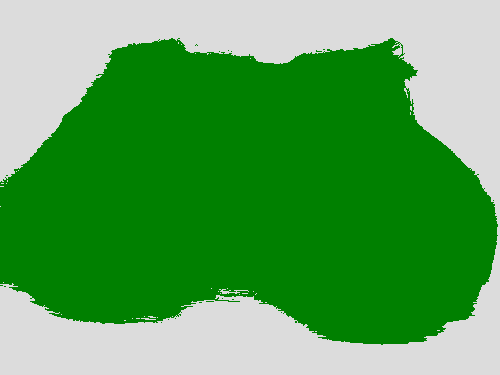}\\
        \includegraphics[ width=0.35\linewidth]{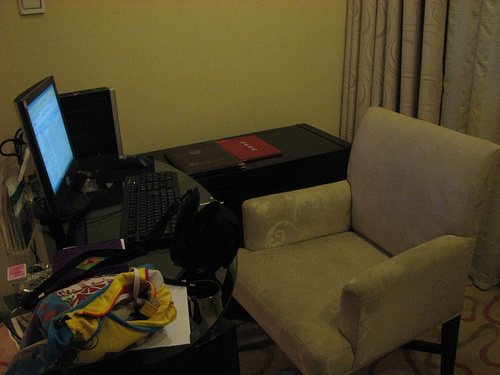}
        &   \includegraphics[ width=0.35\linewidth]{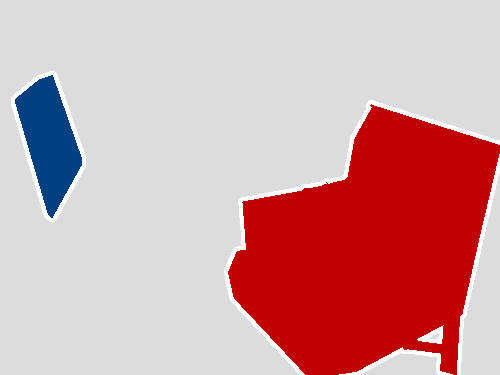}
         &   
         \includegraphics[ width=0.35\linewidth]{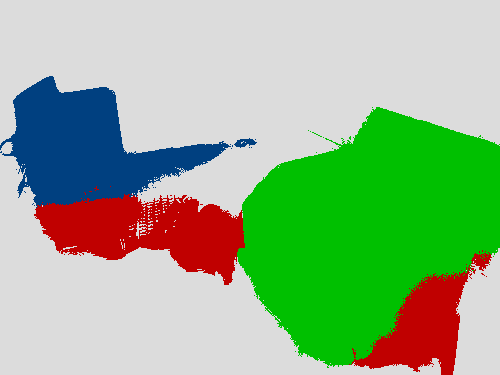} &                  \includegraphics[ width=0.35\linewidth]{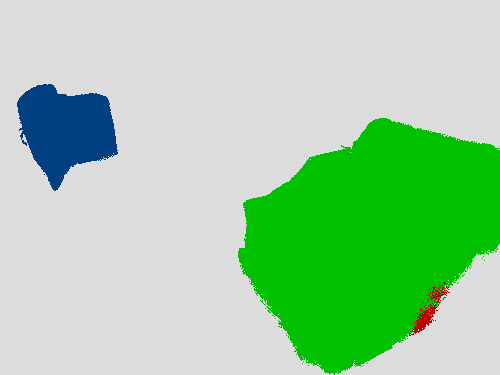}\\
     \end{tabular}  
     }
}
\centering
\caption{ Visualization of the segmentation results of both methods for several examples. ADELE fails to improves segmentation for the bicycle and chair due to  highly structured segmentation errors. 
We set the background color to gray for ease of visualization. }
\label{fig:compare_fig_val_fail}
\end{figure}

\section{Limitations}
\label{sec:limitations}
The success of ADELE seems to rely to some extent on the quality of the initial annotations. When these annotations are of poor quality,
ADELE may only produce a marginal improvement or even have negative impact (see Figure~\ref{fig:perclasscomparision} and ~\ref{fig:compare_fig_val}). 
An related limitation is that when the annotation noise is highly structured, early-learning may not occur, because there may not be sufficient information in the noisy annotations to correct the errors. In that case label correction based on early-learning will be unsuccessful. Illustrative examples are provided in the fifth and sixth rows of Figure~\ref{fig:compare_fig_val}), where the initial annotations completely encompass the bicycle, and completely missclassify the chair as a sofa. 

\section{Conclusion}
In this work, we introduce a novel method to improve the robustness of segmentation models trained on noisy annotations. Inspired from the early-learning phenomenon, we proposed ADELE to boost the performance on the segmentation of thoracic organ, where noise is incorporated to resemble human annotation errors. Moreover, standard segmentation networks, equipped with ADELE, achieve the state-of-the-art results for WSSS on PASCAL VOC 2012. 
We hope that this work will trigger interest in the design of new forms of segmentation methods that provide robustness to annotation noise, as this is a crucial challenge in applications such as medicine. We also hope that the work will motivate further study of the early-learning and memorization phenomena in settings beyond classification. 


\subsubsection*{Acknowledgments}
SL, KL, WZ, and YS were partially supported by NSF NRT grant HDR-1922658. SL was partially supported by NSF grant DMS 2009752 and Alzheimer’s Association grant AARG-NTF-21-848627. KL was partially supported by NIH grant (R01LM013316). YS was partially supported by NIH grant (P41EB017183, R21CA225175). CFG acknowledges support from NSF OAC 2103936.

{\small
\bibliographystyle{ieee_fullname}
\bibliography{cvpr}
}


\appendix

\begin{center}
    {\Large
\center
\onecolumn 
\textbf{ Appendix for ``Adaptive Early-Learning Correction for Segmentation from Noisy Annotations''}}
    \end{center}

\bigskip

The appendix is organized as follows.
\begin{itemize}[leftmargin=*]
    \item In Appendix~\ref{append:addresults}, we include additional figures illustrating the early-learning and memorization phenomena in PASCAL VOC 2012 and SegTHOR. We also include additional results for these two datasets.   
    \item In Appendix~\ref{append:implementationdetails}, we describe the implementation details of our proposed method ADELE, including a description of the hyperparameters, experiment settings, and other technical details. 
    \item In Appendix~\ref{append:additionalablation}, we report ablation studies on the influence of different components and hyperparameters of ADELE. 
    
\end{itemize}

\section*{License of the assets}
\label{license}

\subsection*{Licence for the codes}

We reproduce the code for AffinityNet~\cite{ahn2018learning}, SEAM~\cite{wang2020self}, ICD~\cite{fan2020learning}, all of which are under MIT License according to \url{https://opensource.org/licenses/MIT}.

\subsection*{Licence for the dataset}

As the PASCAL VOC~\cite{everingham2015pascal} includes images obtained from the "flickr" website, we respect the corresponding terms of use for "flickr" according to \url{https://www.flickr.com/help/terms}.

For the SegThor~\cite{lambert2020segthor} dataset, we follow the data use agreement according to \url{https://pagesperso.litislab.fr/cpetitjean/wp-content/uploads/sites/19/2021/03/DataUseAgreement_CHBDatabase.pdf}.

\begin{figure}[htbp]
    \centering
\includegraphics[width=\linewidth]{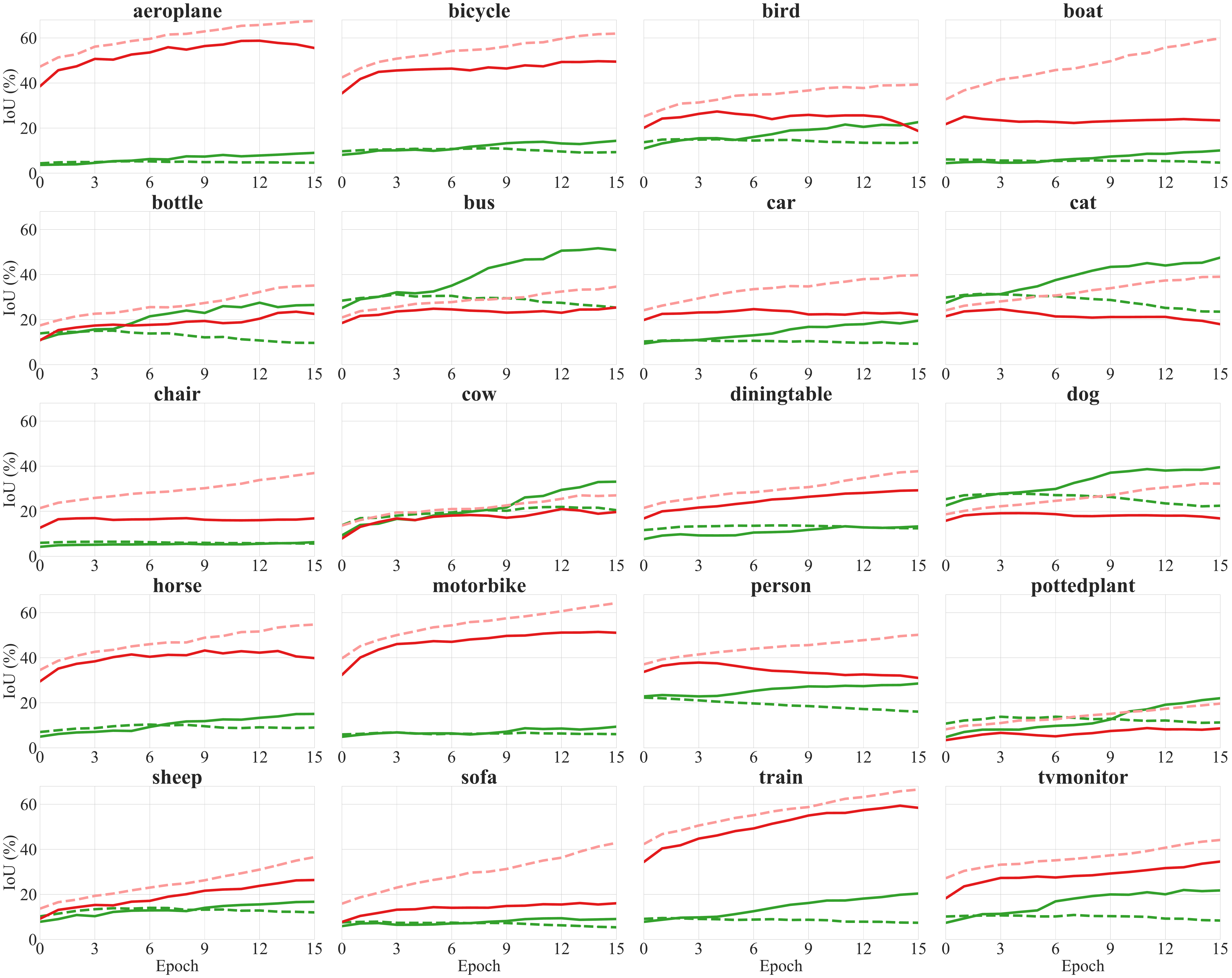}
\includegraphics[width=1\textwidth]{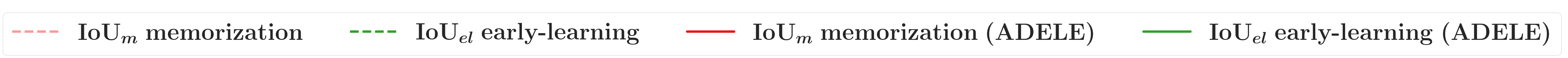}
 \caption{We visualize the effect of early learning ($\text{IoU}_{el}$, green curves) and memorization ($\text{IoU}_{m}$, red curves) on segmentation models trained with (solid lines) and without (dashed lines) ADELE for each category of the WSSS dataset VOC 2012~\cite{everingham2015pascal} 
The WSSS model is a standard DeepLab-v1 network trained with annotations obtained from SEAM~\cite{wang2020self}. 
$\text{IoU}_{el}$ is the IOU between the model output and the ground truth computed over the incorrectly-labeled pixels. $\text{IoU}_{m}$ is the IOU between the model output and the incorrect annotations. For all classes, $\text{IoU}_m$ increases substantially as training proceeds  because the model gradually memorizes the incorrect annotations. This again occurs at different speeds for different categories. In contrast, $\text{IoU}_{el}$ first increases during an early-learning stage where the model learns to correctly segment the incorrectly-labeled pixels, but eventually decreases as memorization occurs (the phenomenon is more evident when we zoom in, as shown in Figure~\ref{fig:ELzoomin} in the Appendix). Like memorization, early-learning also happens at varying speeds for the different semantic categories.}
    \label{fig:early_learning_voc}
\end{figure}

\begin{figure*}[htbp]
\centering
\includegraphics[width=\linewidth]{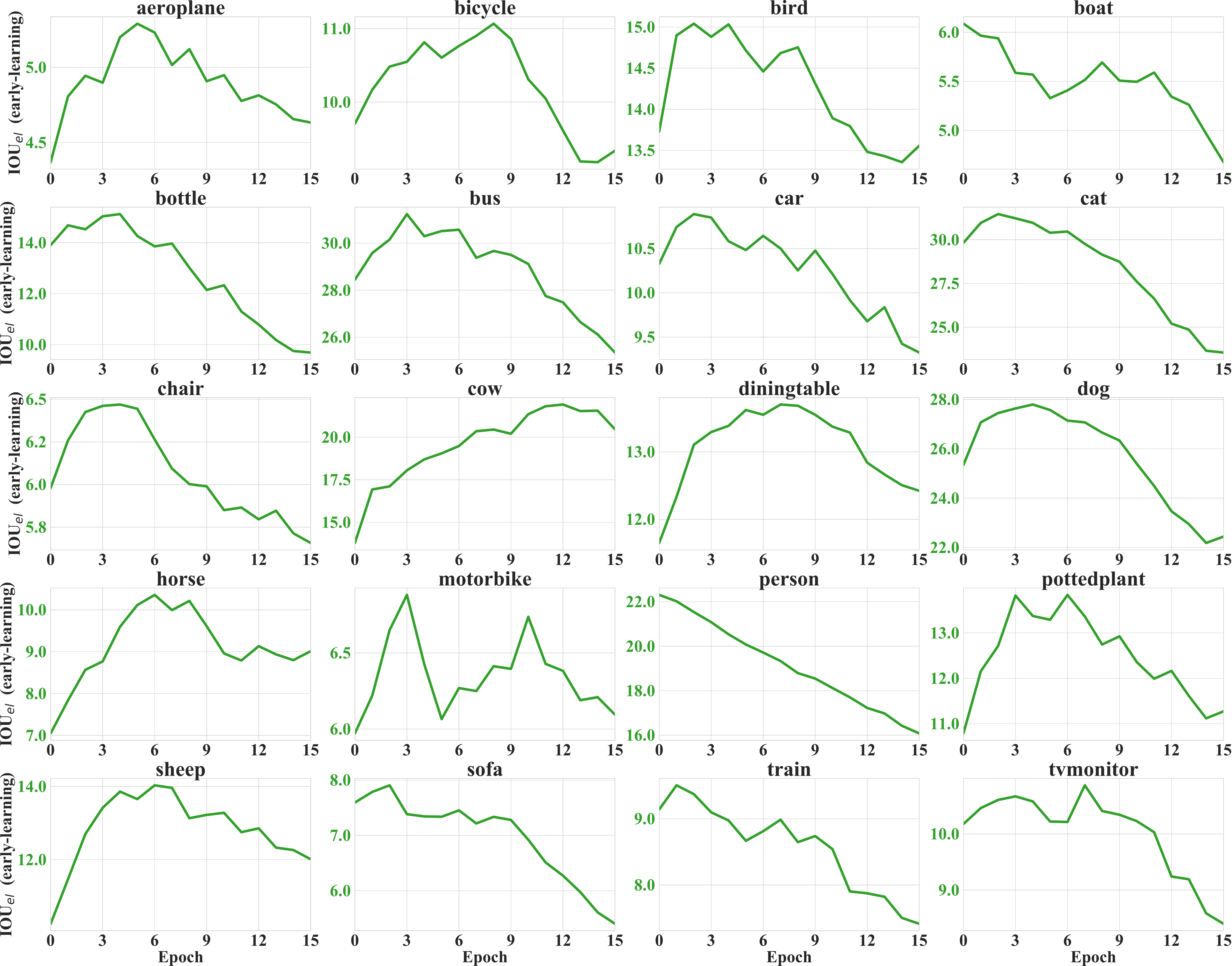}
\caption{Zoomed-in illustration of early learning for the different semantic categories on PASCAL VOC 2012. $\text{IoU}_{el}$ first  increases during an early-learning stage where the model learns to correctly segment the incorrectly-labeled pixels, but eventually decreases as memorization occurs. Early learning happens at varying speeds for the different semantic categories. The experimental setting is the same as Figure~\ref{fig:EL}.}
\label{fig:ELzoomin}
\end{figure*}
\begin{figure}[h]
    \centering
    \includegraphics[width=0.6\linewidth]{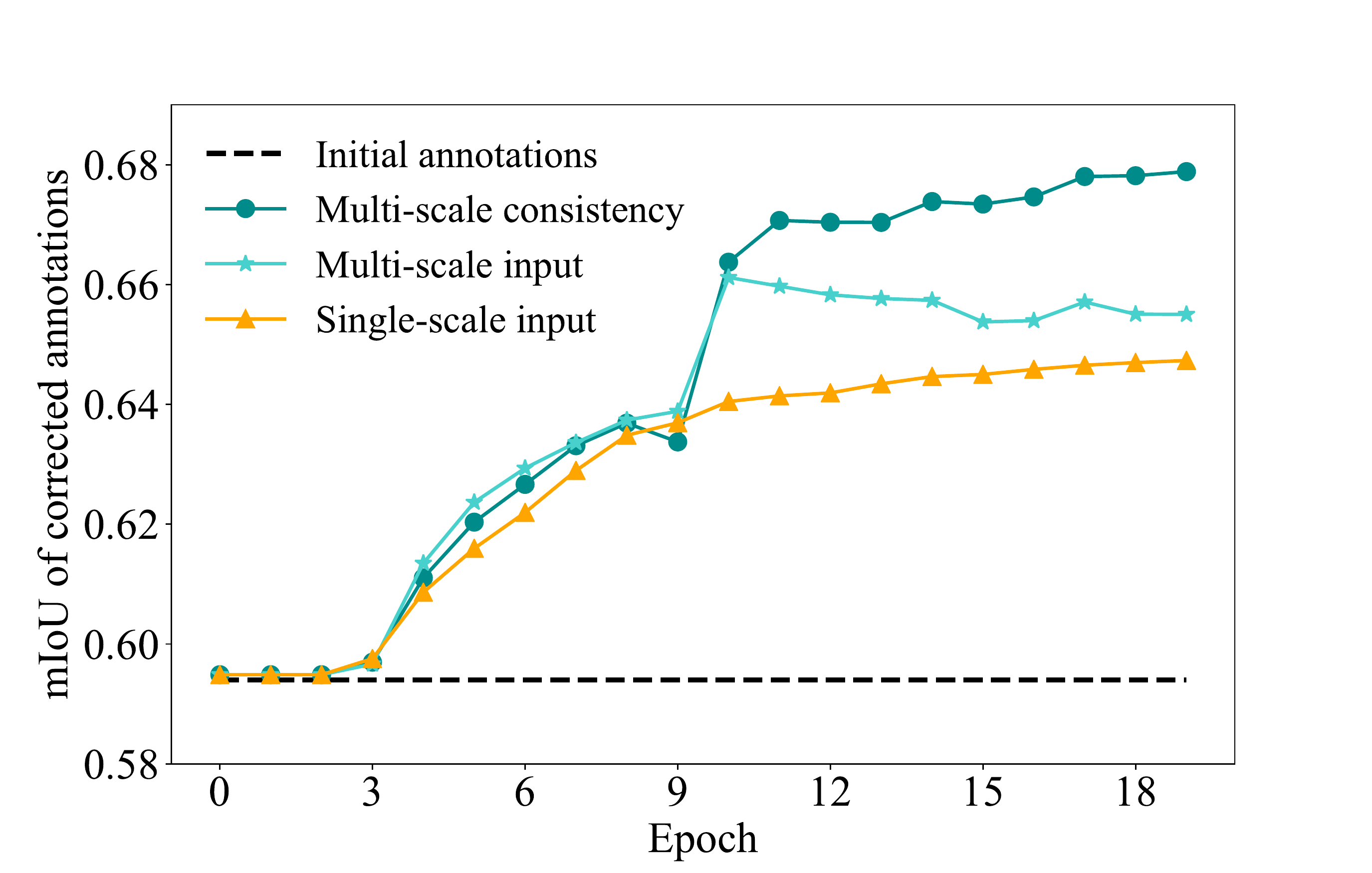}
    \caption{Multi-scaleconsistency regularization leads to more accurate corrected annotations (result on Pascal VOC).}
    \label{fig:correction_quality_VOC}
\end{figure}

\begin{figure*}[t]
\resizebox{0.95\linewidth}{!}{
\begin{tabular}{>{\centering\arraybackslash}m{0.33\linewidth} >{\centering\arraybackslash}m{0.33\linewidth} >{\centering\arraybackslash}m{0.33\linewidth}}
\includegraphics[width=0.95\linewidth]{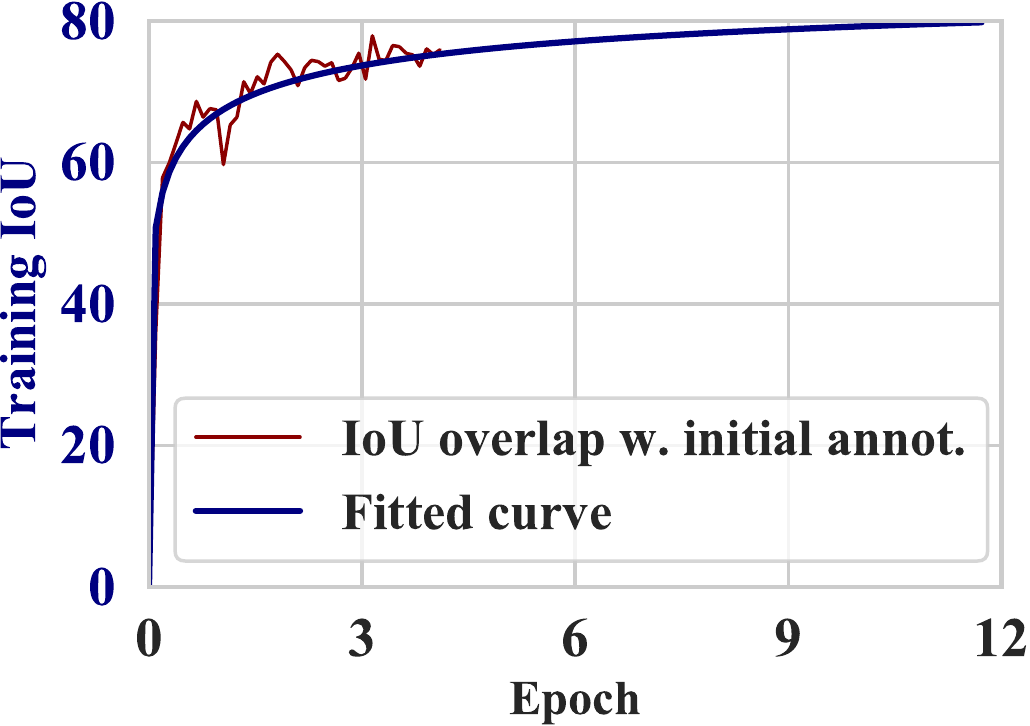} &
\includegraphics[width=0.95\linewidth]{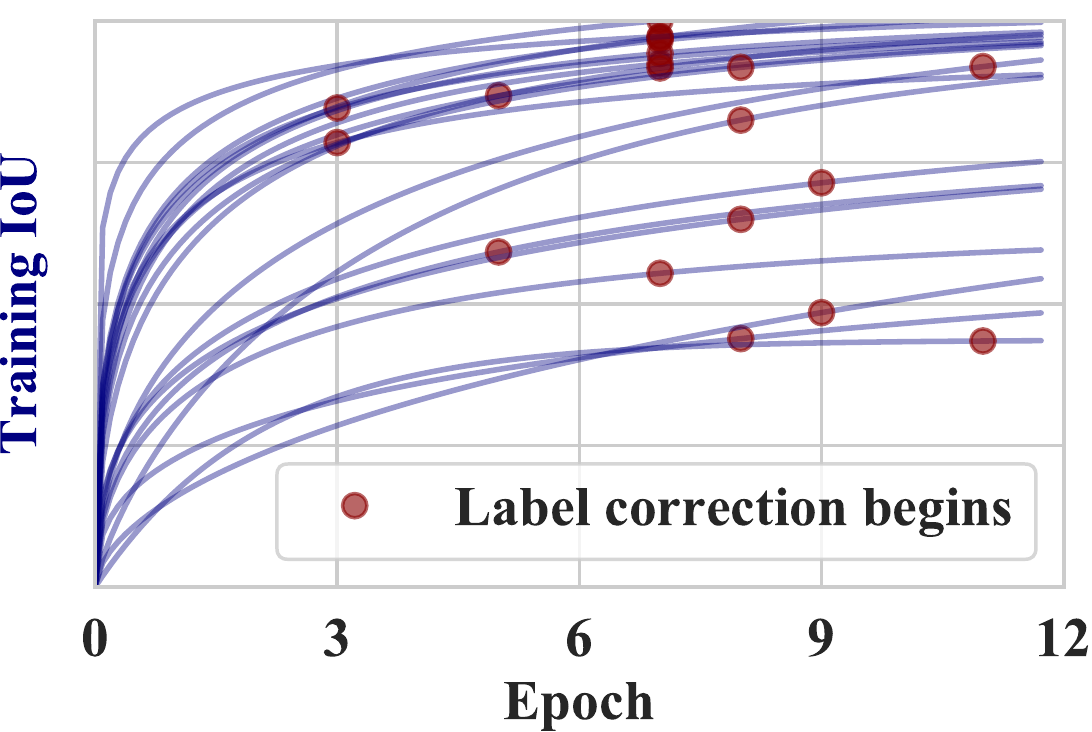}  &
\includegraphics[width=1.1\linewidth]{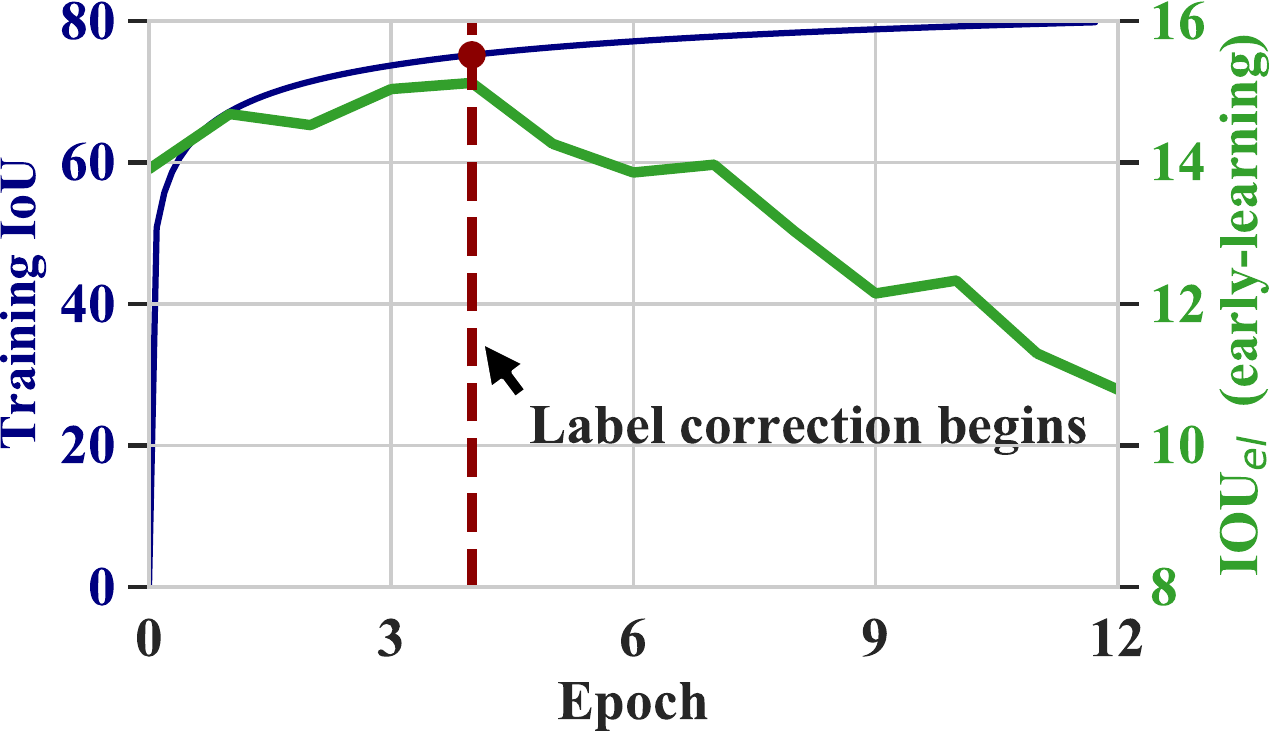}
\end{tabular} 
}
\caption{Illustration of the proposed curve fitting method to decide when to begin label correction in ADELE (Results on Pascal VOC). On the left, we plot the IoU between the model predictions and the initial noisy annotations for the same model used in Figures~\ref{fig:ELzoomin} and~\ref{fig:compare_fig2} and the corresponding fit with the parametric model in Equation~\ref{eq: fitting_function}. The label correction beginning iteration is based on the relative slope change of the fitted curve. The center image shows the label correction times for different semantic categories, showing that they are quite different. On the right graph, the green line shows the $\text{IoU}_{el}$ for a given category. The $\text{IoU}_{el}$ equals the IoU between the model output and the ground truth computed over the incorrectly-labeled pixels, and therefore quantifies early-learning. The label correction begins close to the end of the early-learning phase, as desired. 
}
\label{fig:curvefittin_voc}
\end{figure*}

\begin{figure*}[htbp]
\centering
\includegraphics[width=\linewidth]{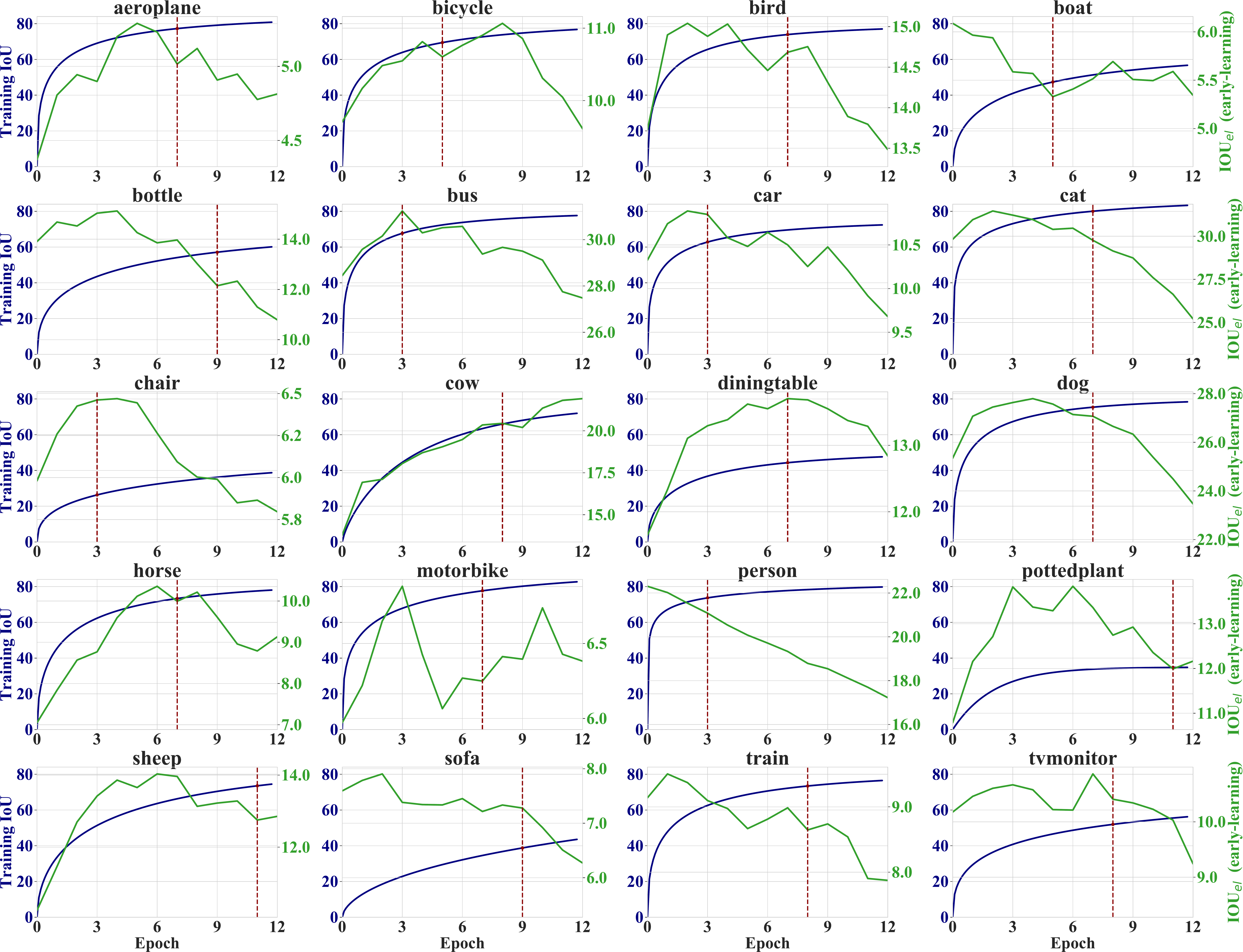}
\caption{Illustration of the proposed curve fitting method to decide when to begin label correction in ADELE. The blue line shows the parametric model in Equation~\ref{eq: fitting_function} fit to the IoU between the model predictions and the initial noisy annotations for the same WSSS model used in Figures~\ref{fig:EL} and~\ref{fig:compare_fig2}. The green line shows the $\text{IoU}_{el}$ for a given category. The $\text{IoU}_{el}$ equals the IOU between the model output and the ground truth computed over the incorrectly-labeled pixels, and therefore quantifies early-learning. The label correction begins close to the end of the early-learning phase for most of the categories with a few exceptions (boat and motorbike).}
\label{fig:figure3_moreexp}
\end{figure*}

\begin{figure}[h]
    \centering
         \resizebox{0.6\linewidth}{!}{
        \begin{tabular}{c c c c c }
    \centering
       \Huge  Input &    \Huge Ground Truth &   \Huge SEAM &  \Huge  SEAM+ADELE \\
        \centering
        \includegraphics[ width=0.35\linewidth]{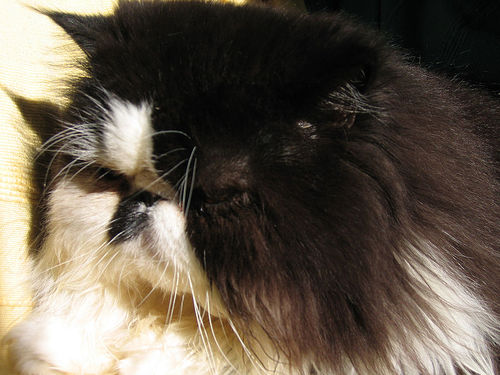}
        &   \includegraphics[ width=0.35\linewidth]{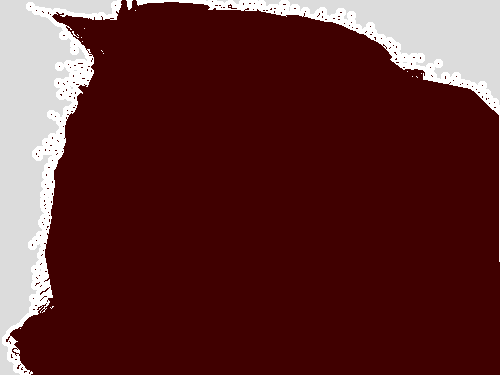}
         &   
         \includegraphics[ width=0.35\linewidth]{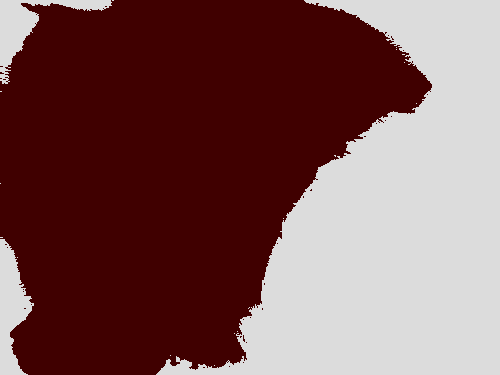} &                  \includegraphics[ width=0.35\linewidth]{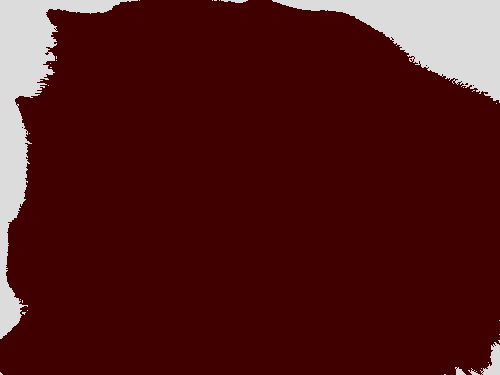}
\\
        \includegraphics[ width=0.35\linewidth,trim={0 0px 0 70px},clip]{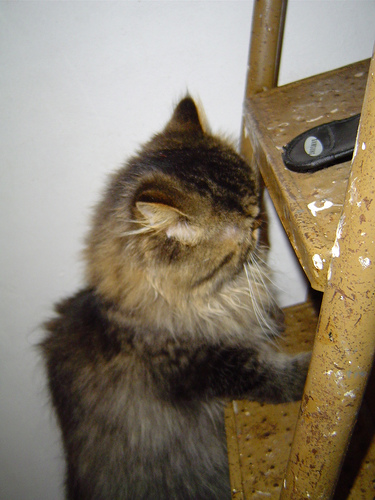}
        &   \includegraphics[ width=0.35\linewidth,trim={0 0px 0 70px},clip]{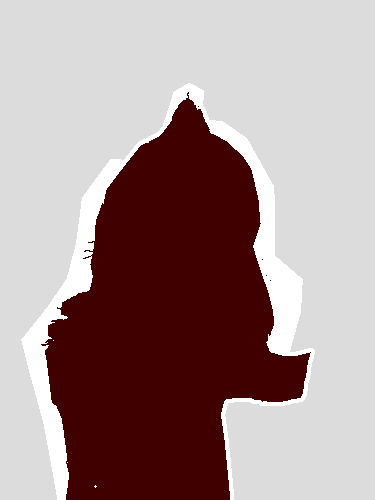}
         &   
         \includegraphics[ width=0.35\linewidth,trim={0 0px 0 70px},clip]{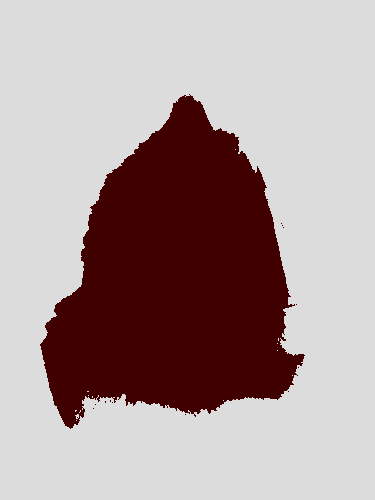} &                  \includegraphics[ width=0.35\linewidth,trim={0 0px 0 70px},clip]{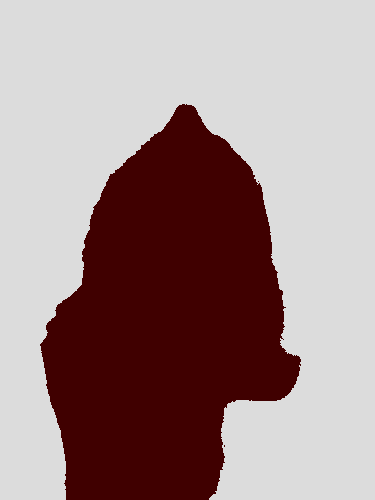}\\
        \includegraphics[ width=0.35\linewidth]{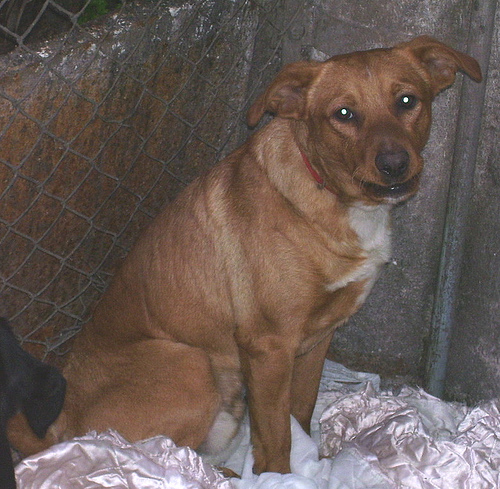}
        &   \includegraphics[ width=0.35\linewidth]{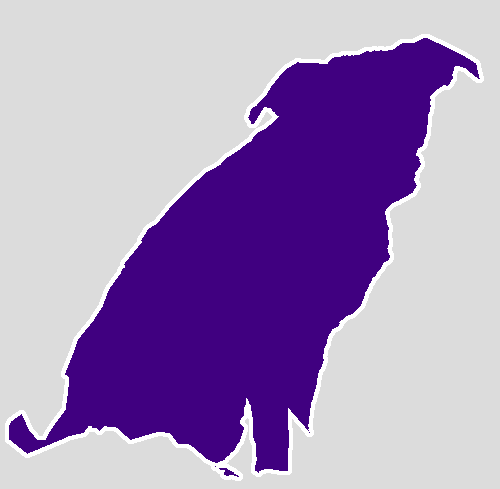}
         &   
         \includegraphics[ width=0.35\linewidth]{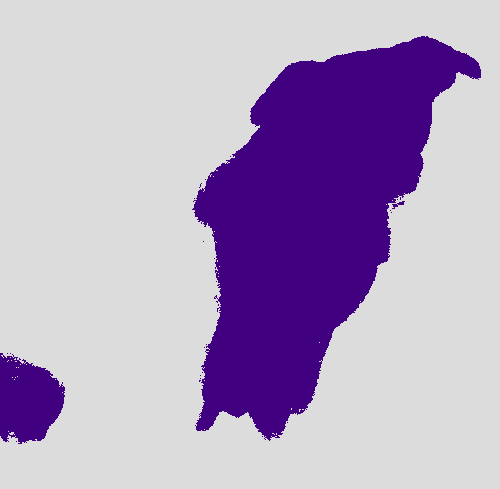} &                  \includegraphics[ width=0.35\linewidth]{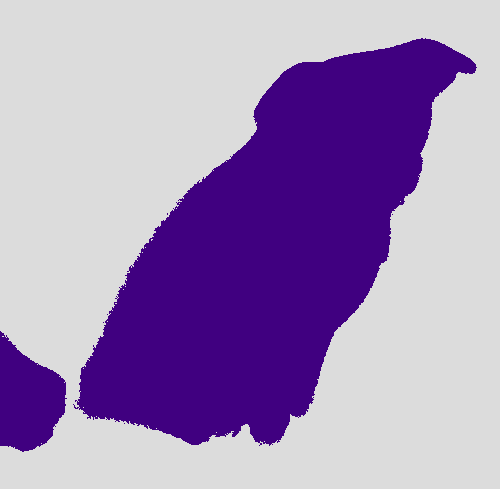}\\
                 \includegraphics[ width=0.35\linewidth]{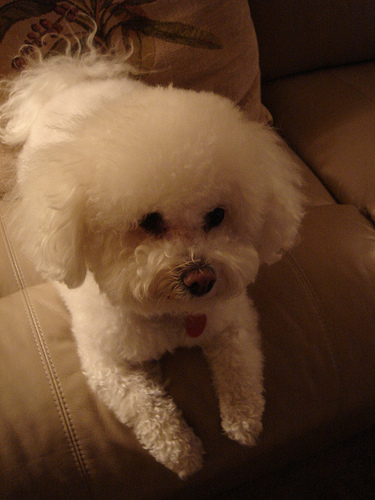}
        &   \includegraphics[ width=0.35\linewidth]{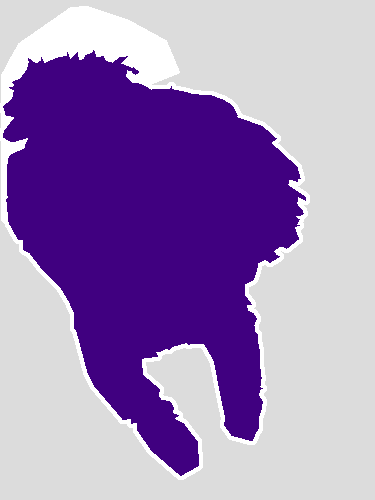}
         &   
         \includegraphics[ width=0.35\linewidth]{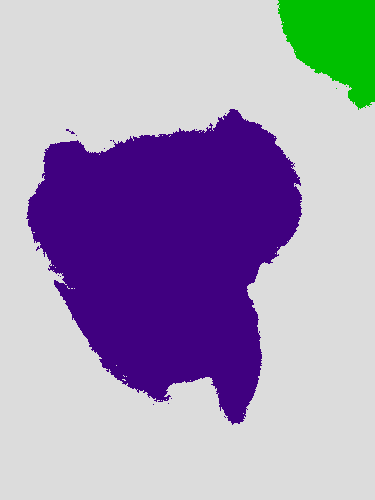} &                  \includegraphics[ width=0.35\linewidth]{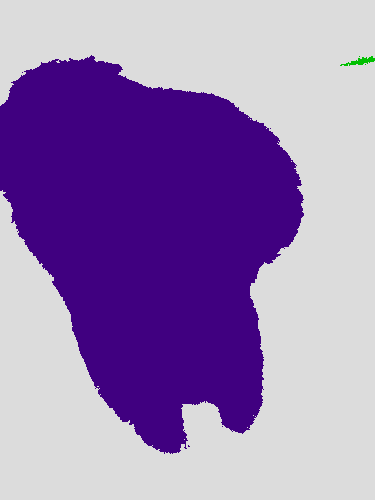}\\
     \includegraphics[ width=0.35\linewidth]{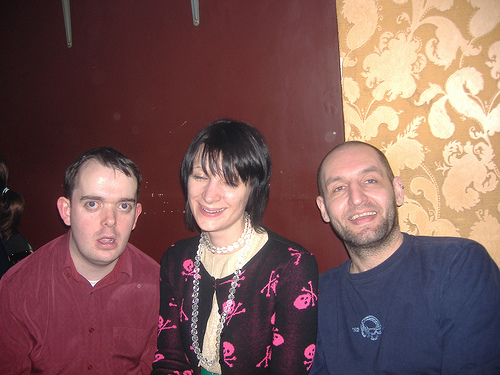}
        &   \includegraphics[ width=0.35\linewidth]{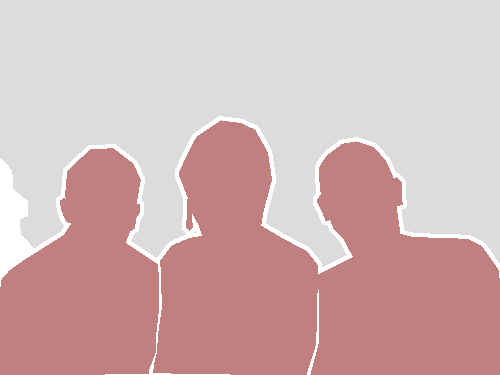}
         &   
         \includegraphics[ width=0.35\linewidth]{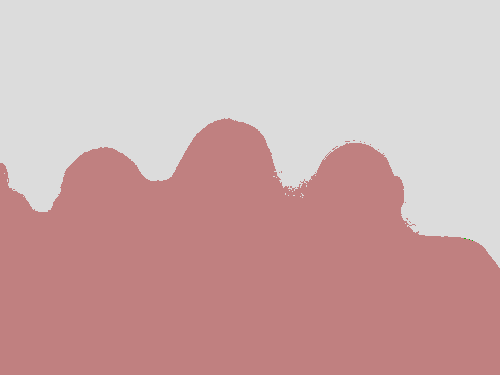} &                  \includegraphics[ width=0.35\linewidth]{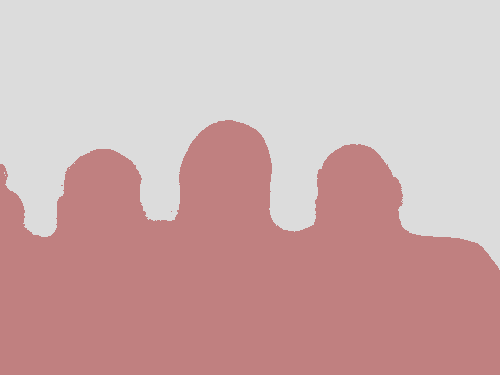}\\
              \includegraphics[ width=0.35\linewidth]{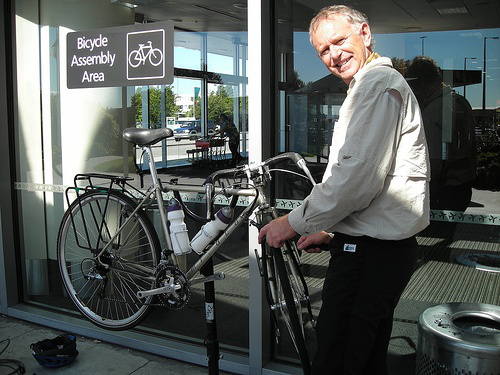}
        &   \includegraphics[ width=0.35\linewidth]{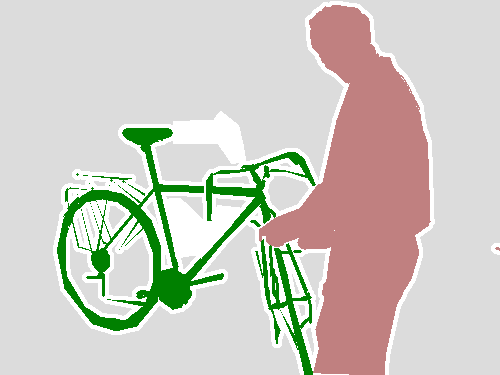}
         &   
         \includegraphics[ width=0.35\linewidth]{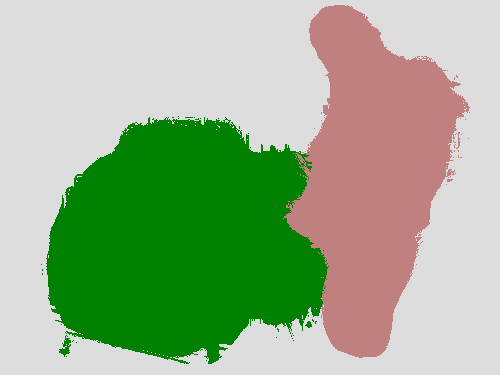} &                  \includegraphics[ width=0.35\linewidth]{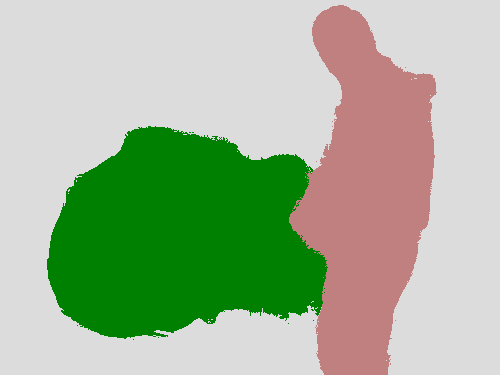}\\
                       \includegraphics[ width=0.35\linewidth]{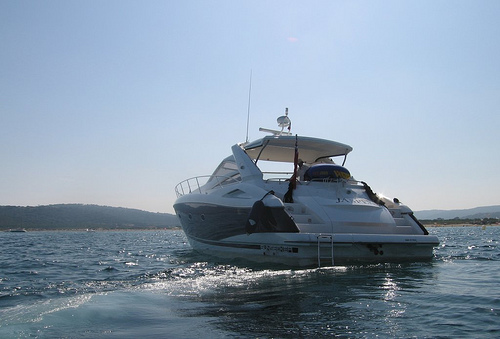}
        &   \includegraphics[ width=0.35\linewidth]{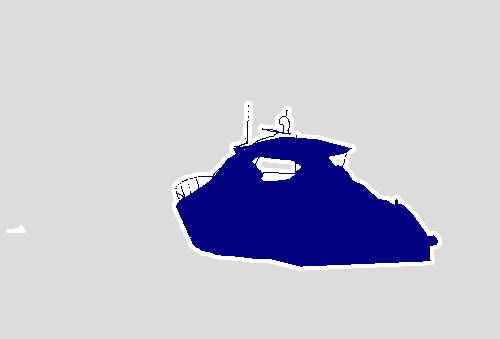}
         &   
         \includegraphics[ width=0.35\linewidth]{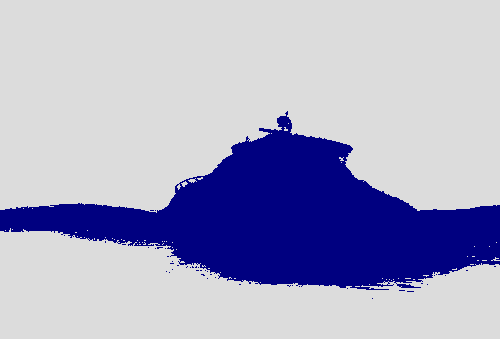} &                  \includegraphics[ width=0.35\linewidth]{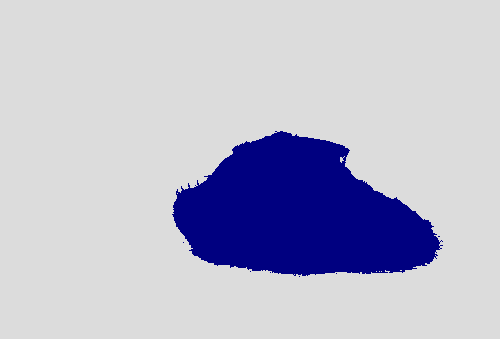}\\
    \includegraphics[ width=0.35\linewidth]{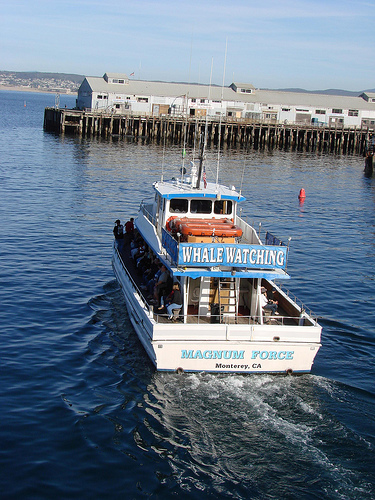}
        &   \includegraphics[ width=0.35\linewidth]{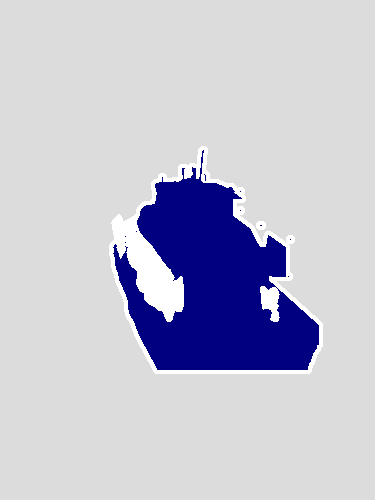}
         &   
         \includegraphics[ width=0.35\linewidth]{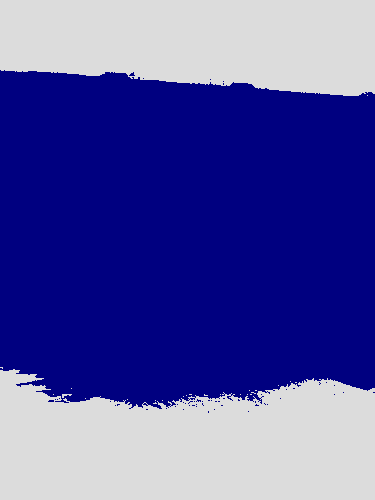} &                  \includegraphics[ width=0.35\linewidth]{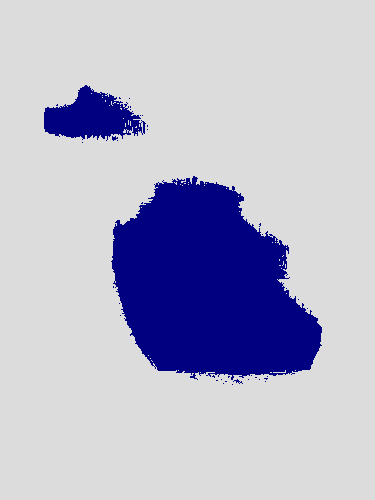}\\
     \end{tabular}  
     }
     \caption{Additional segmentation results of SEAM and SEAM+ADELE for several examples on the validation set of PASCAL VOC 2012. We set the background color to gray for ease of visualization. Supplementary for Figure~\ref{fig:compare_fig_val}.}
    \label{fig:addfigures}
\end{figure}

\begin{figure}[h]
    \centering
         \resizebox{0.6\linewidth}{!}{
        \begin{tabular}{c c c c c }
    \centering
       \Huge  Input &    \Huge Ground Truth &   \Huge SEAM &  \Huge  SEAM+ADELE \\
        \centering
        \includegraphics[ width=0.35\linewidth]{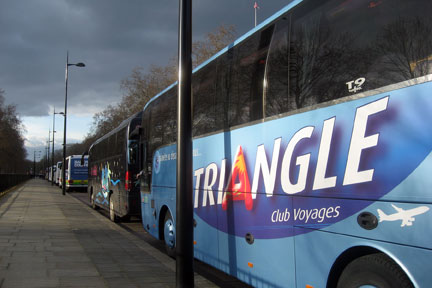}
        &   \includegraphics[ width=0.35\linewidth]{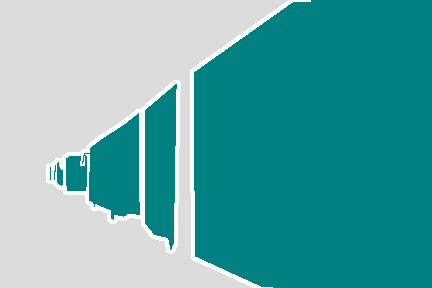}
         &   
         \includegraphics[ width=0.35\linewidth]{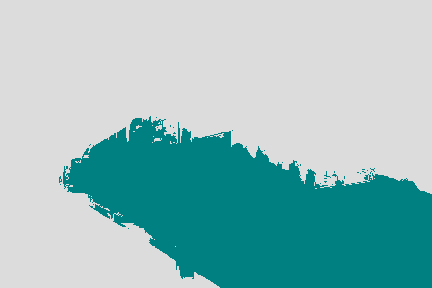} &                  \includegraphics[ width=0.35\linewidth]{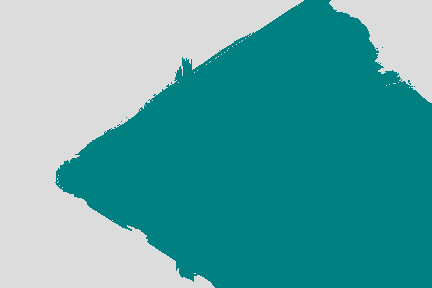}
\\
        \includegraphics[ width=0.35\linewidth]{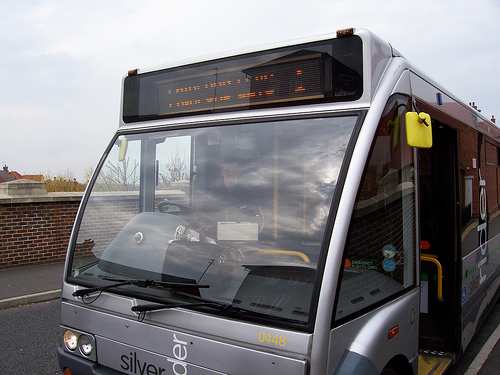}
        &   \includegraphics[ width=0.35\linewidth]{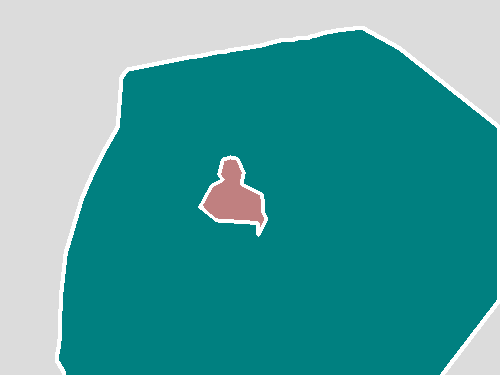}
         &   
         \includegraphics[ width=0.35\linewidth]{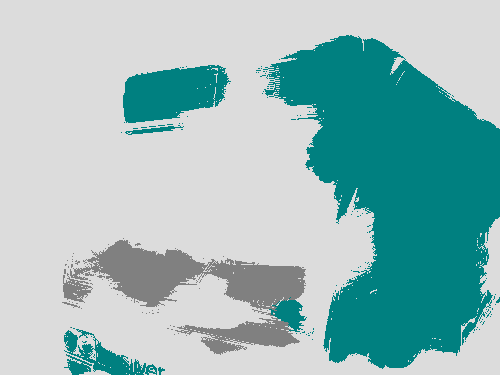} &                  \includegraphics[ width=0.35\linewidth]{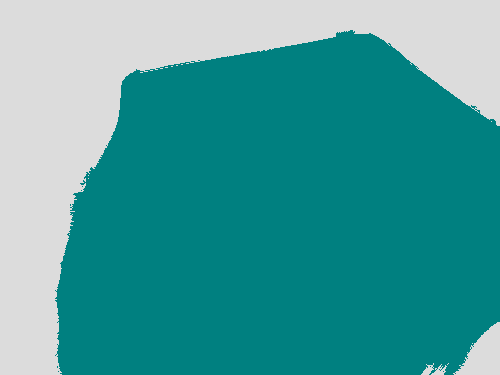}\\
        \includegraphics[ width=0.35\linewidth]{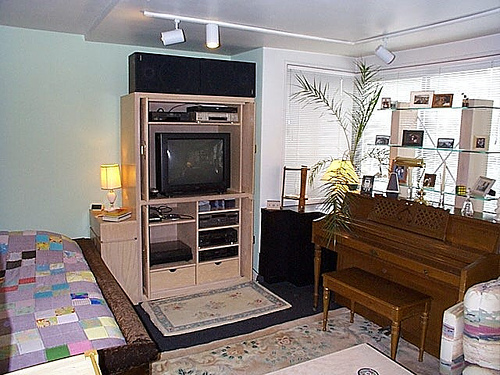}
        &   \includegraphics[ width=0.35\linewidth]{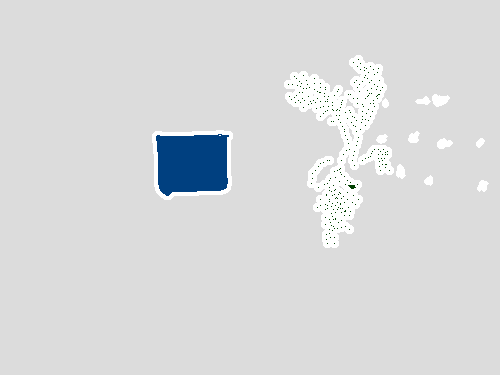}
         &   
         \includegraphics[ width=0.35\linewidth]{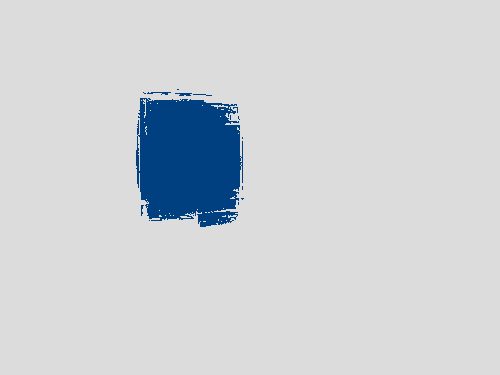} &                  \includegraphics[ width=0.35\linewidth]{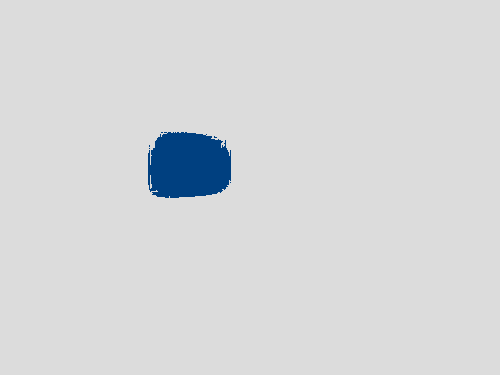} \\
                 \includegraphics[ width=0.35\linewidth]{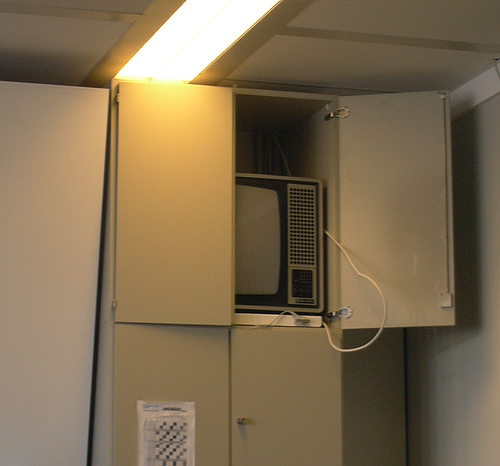}
        &   \includegraphics[ width=0.35\linewidth]{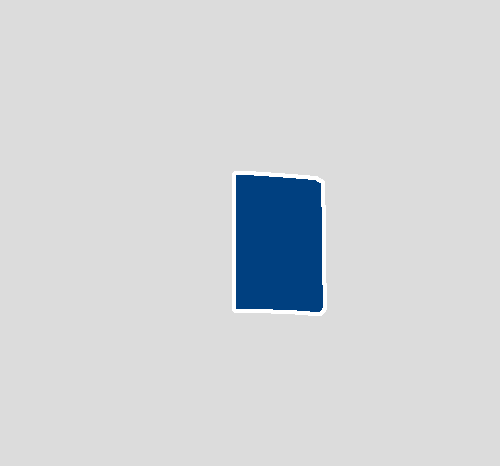}
         &   
         \includegraphics[ width=0.35\linewidth]{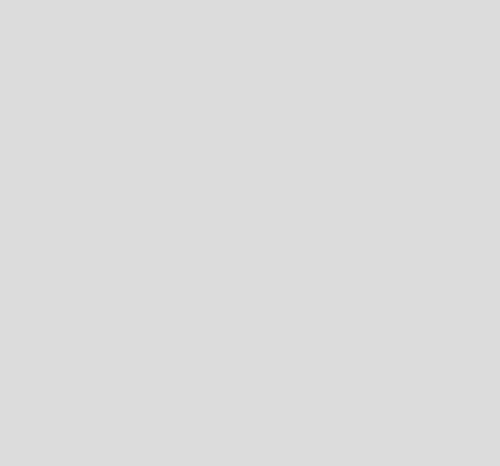} &                  \includegraphics[ width=0.35\linewidth]{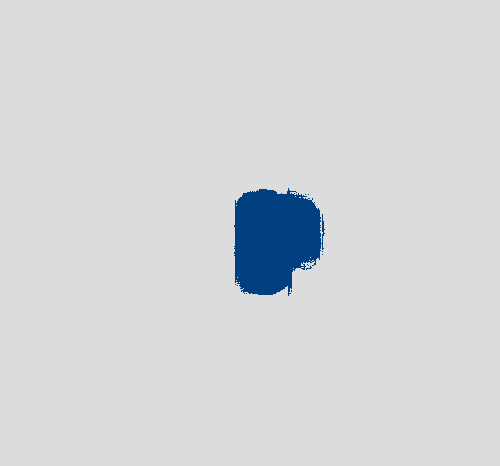} \\
                          \includegraphics[ width=0.35\linewidth]{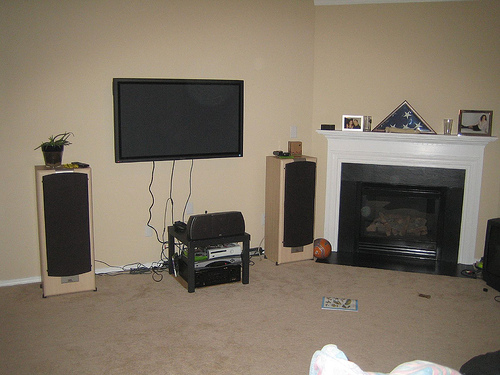}
        &   \includegraphics[ width=0.35\linewidth]{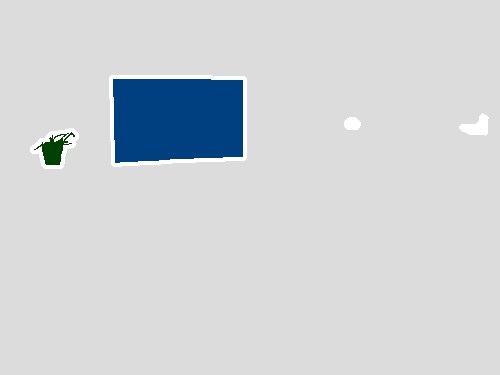}
         &   
         \includegraphics[ width=0.35\linewidth]{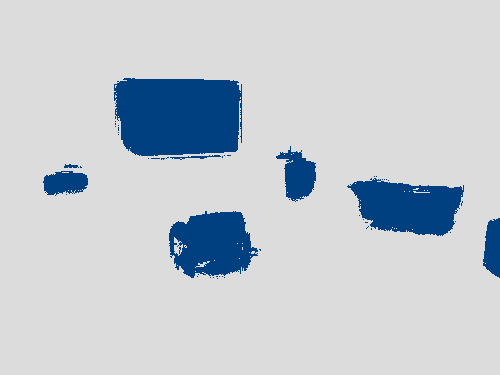} &                  \includegraphics[ width=0.35\linewidth]{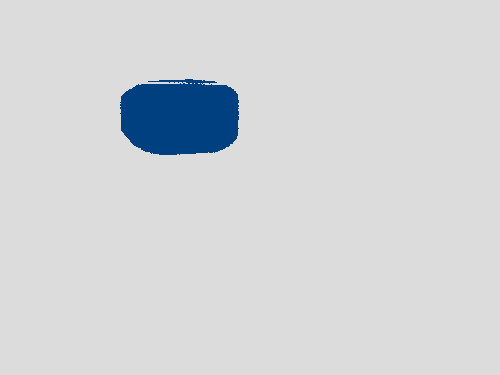} \\
        \includegraphics[ width=0.35\linewidth]{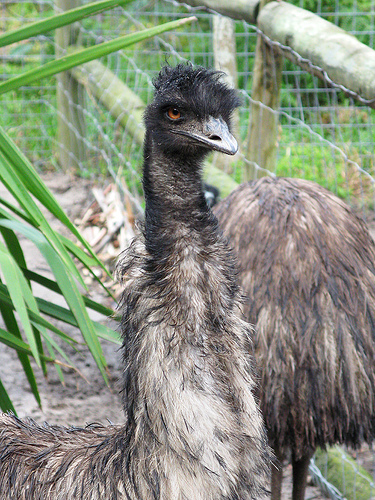}
        &   \includegraphics[ width=0.35\linewidth]{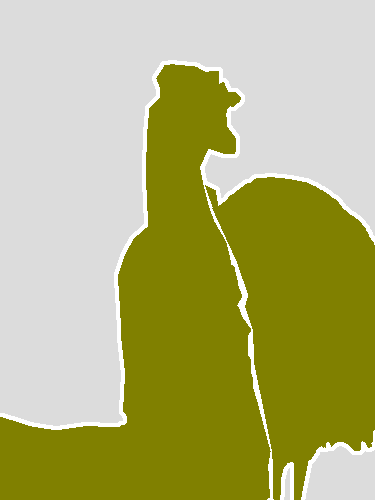}
         &   
         \includegraphics[ width=0.35\linewidth]{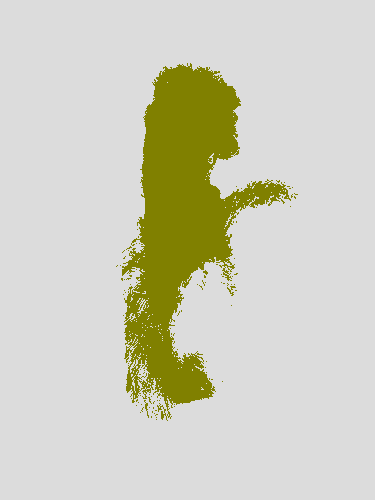} &                  \includegraphics[ width=0.35\linewidth]{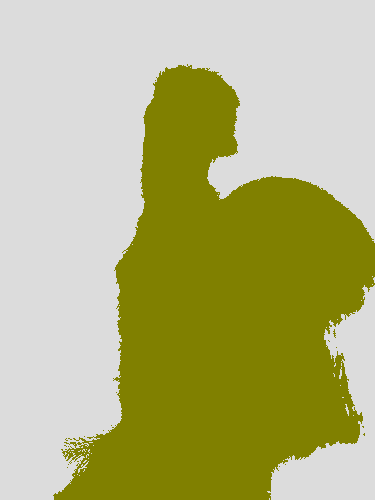} \\
        \includegraphics[ width=0.35\linewidth]{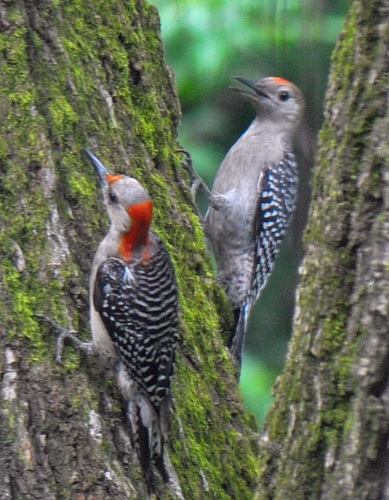}
        &   \includegraphics[ width=0.35\linewidth]{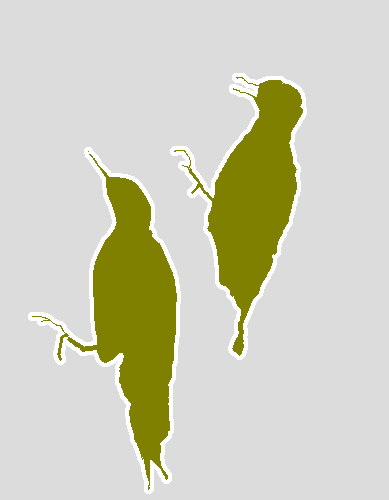}
         &   
         \includegraphics[ width=0.35\linewidth]{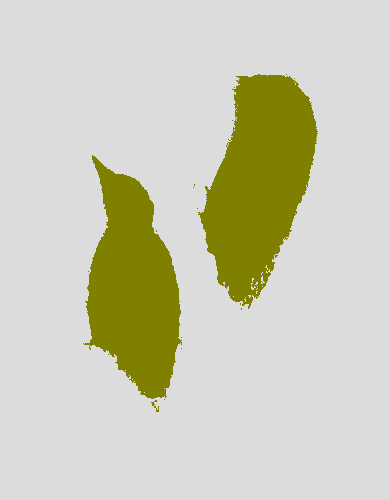} &                  \includegraphics[ width=0.35\linewidth]{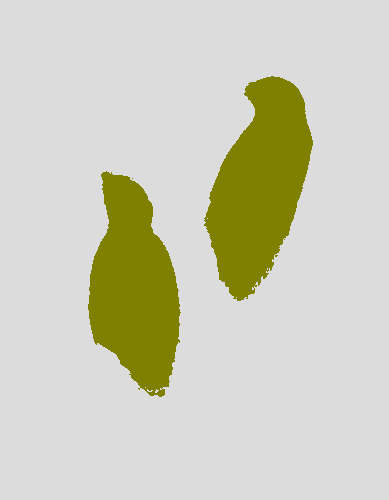} \\
                \includegraphics[ width=0.35\linewidth]{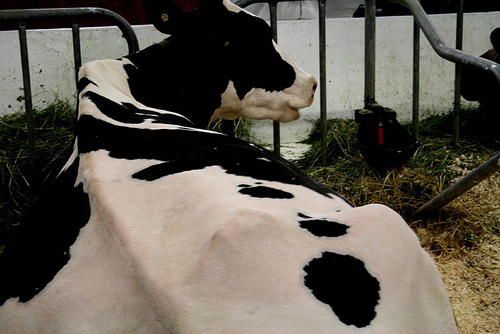}
        &   \includegraphics[ width=0.35\linewidth]{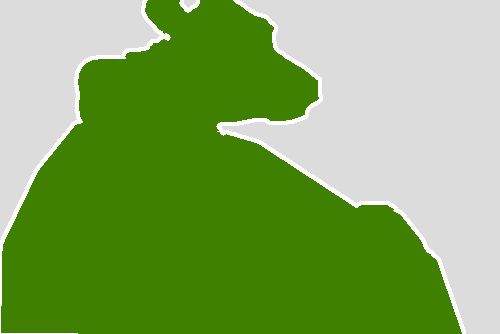}
         &   
         \includegraphics[ width=0.35\linewidth]{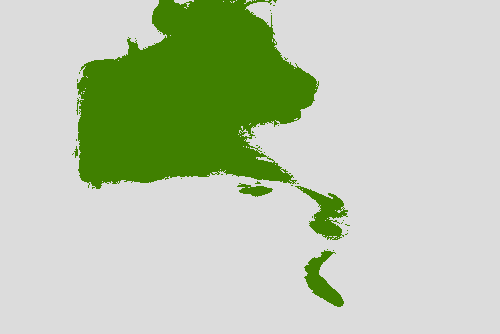} &                  \includegraphics[ width=0.35\linewidth]{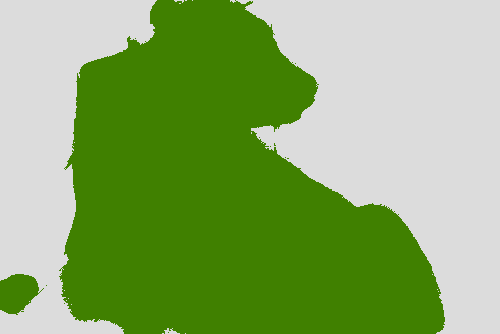}
\\
        \includegraphics[ width=0.35\linewidth]{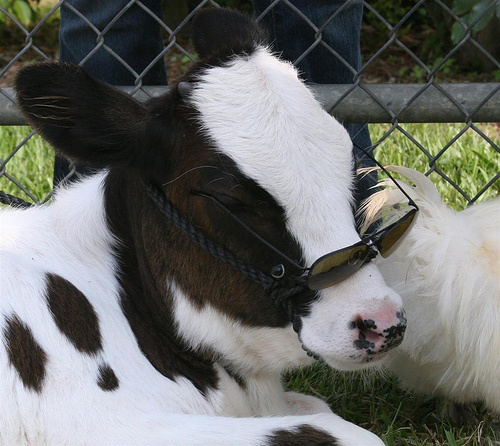}
        &   \includegraphics[ width=0.35\linewidth]{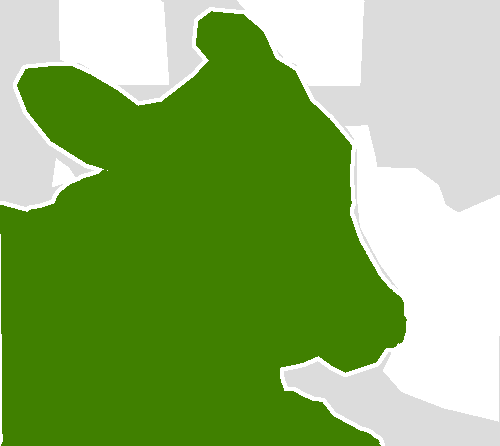}
         &   
         \includegraphics[ width=0.35\linewidth]{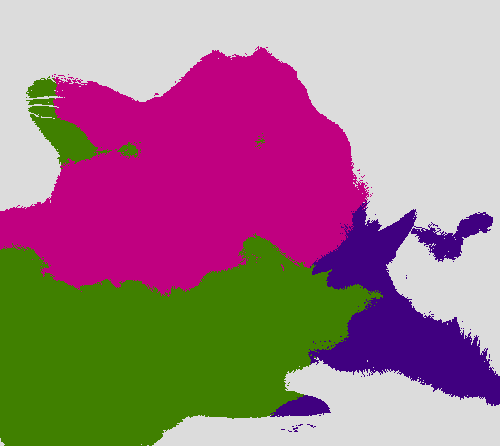} &                  \includegraphics[ width=0.35\linewidth]{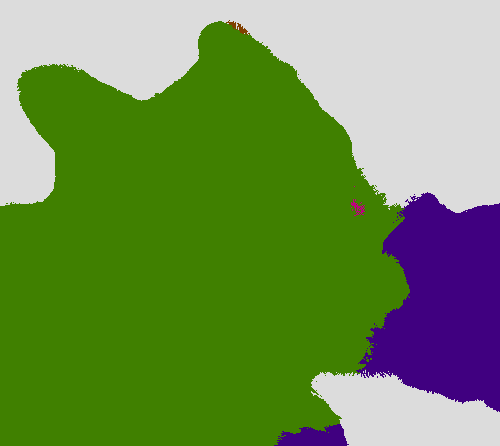}
\\
     \end{tabular}  
     }
     \caption{Additional segmentation results of SEAM and SEAM+ADELE for several examples on the validation set of PASCAL VOC 2012. We set the background color to gray for ease of visualization. Supplementary for Figure~\ref{fig:compare_fig_val}.}
    \label{fig:addfigures2}
\end{figure}

\begin{figure}[h]
    \centering
         \resizebox{0.6\linewidth}{!}{
        \begin{tabular}{c c c c c }
    \centering
       \Huge  Input &    \Huge Ground Truth &   \Huge SEAM &  \Huge  SEAM+ADELE \\
        \centering
                \includegraphics[ width=0.35\linewidth]{appendix/Val_for_VOC/bottle/2010_004551/input.png}
        &   \includegraphics[ width=0.35\linewidth]{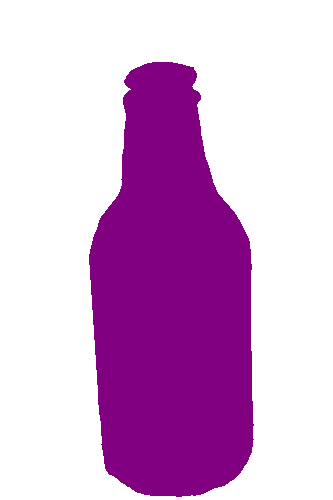}
         &   
         \includegraphics[ width=0.35\linewidth]{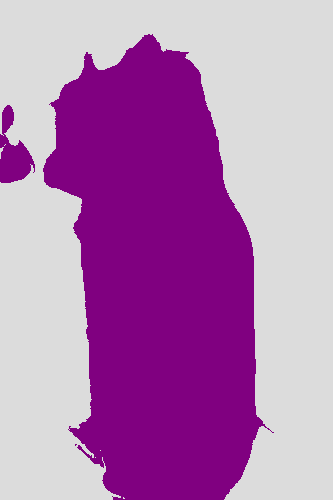} &                  \includegraphics[ width=0.35\linewidth]{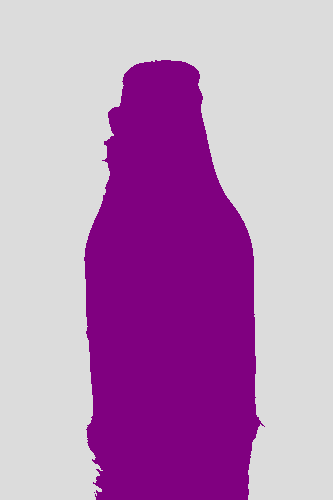} \\
        \includegraphics[ width=0.35\linewidth]{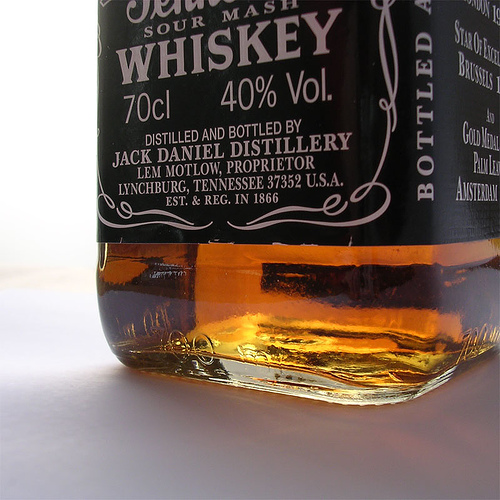}
        &   \includegraphics[ width=0.35\linewidth]{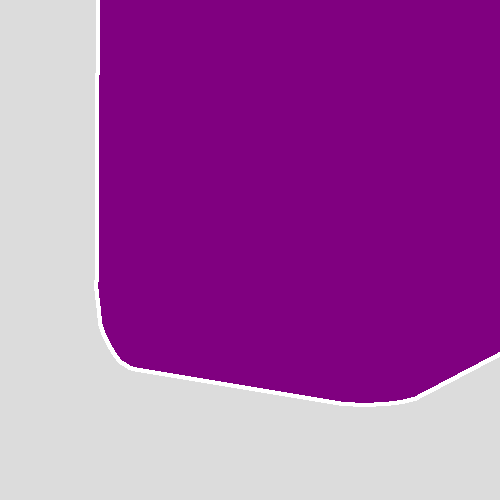}
         &   
         \includegraphics[ width=0.35\linewidth]{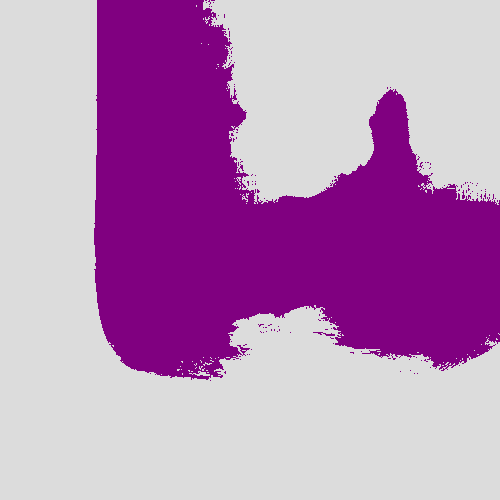} &                  \includegraphics[ width=0.35\linewidth]{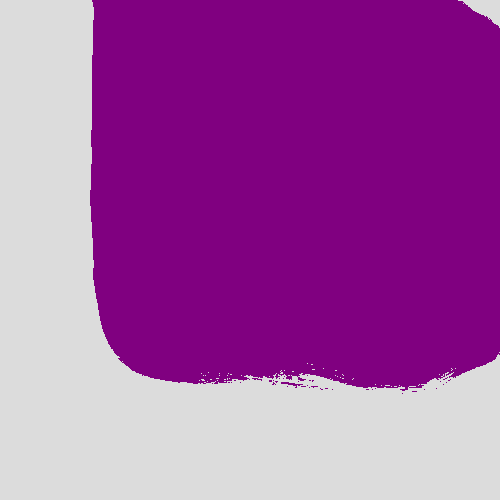}\\
        \includegraphics[ width=0.35\linewidth]{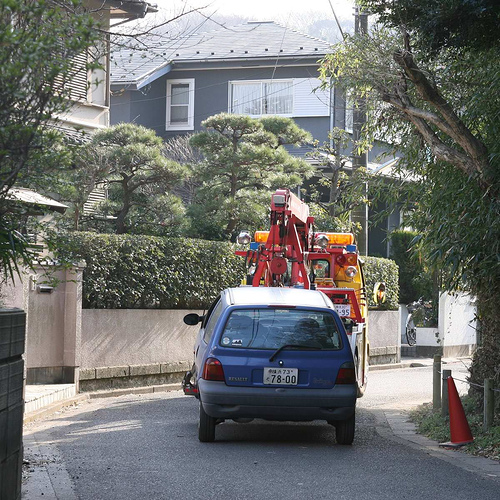}
        &   \includegraphics[ width=0.35\linewidth]{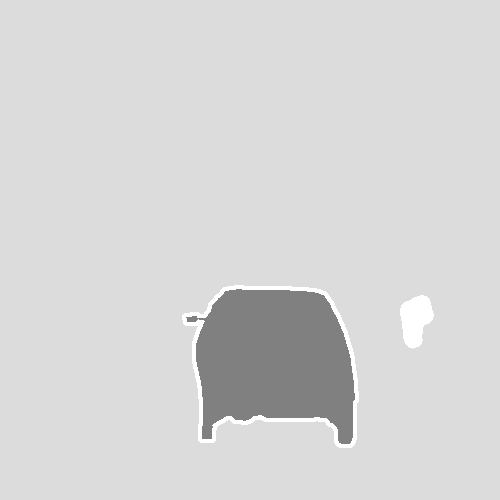}
         &   
         \includegraphics[ width=0.35\linewidth]{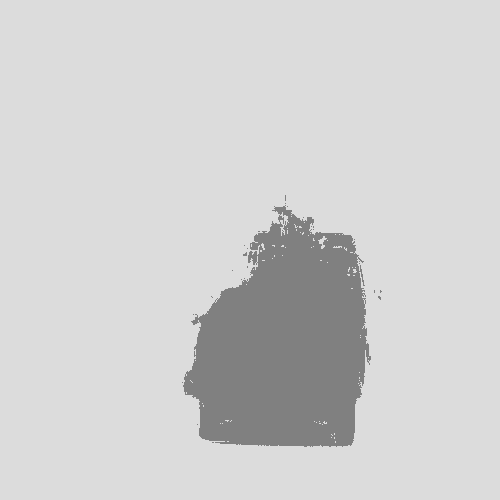} &                  \includegraphics[ width=0.35\linewidth]{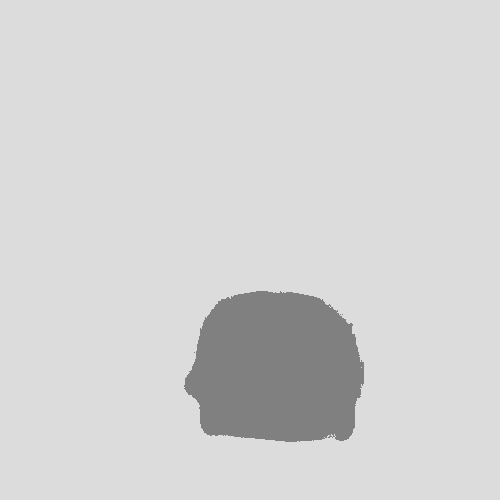}
\\
        \includegraphics[ width=0.35\linewidth]{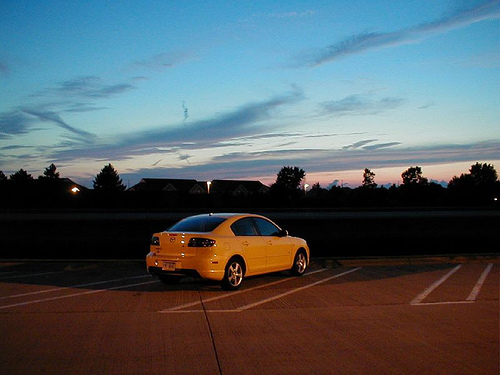}
        &   \includegraphics[ width=0.35\linewidth]{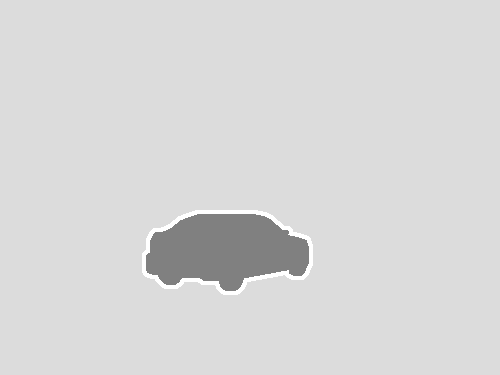}
         &   
         \includegraphics[ width=0.35\linewidth]{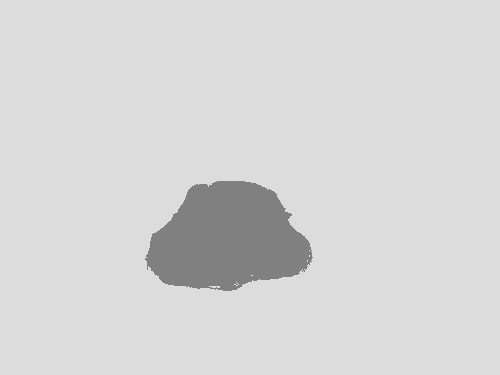} &                  \includegraphics[ width=0.35\linewidth]{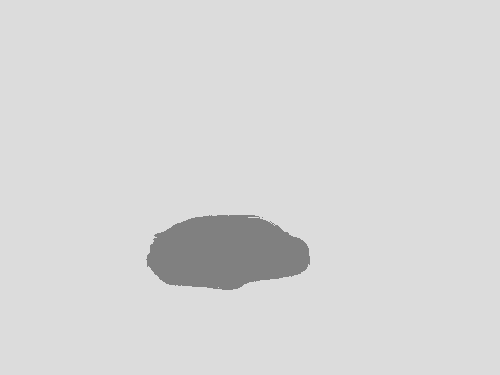}
\\
        \includegraphics[ width=0.35\linewidth]{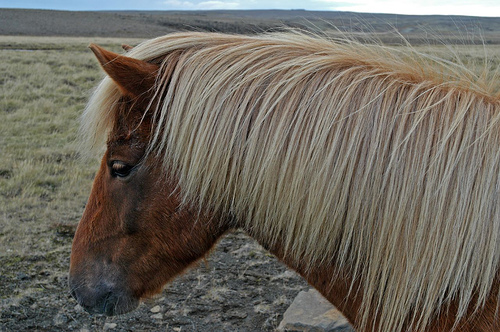}
        &   \includegraphics[ width=0.35\linewidth]{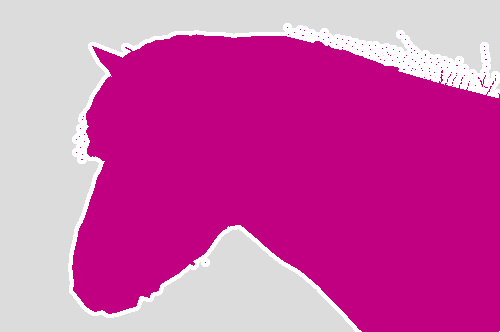}
         &   
         \includegraphics[ width=0.35\linewidth]{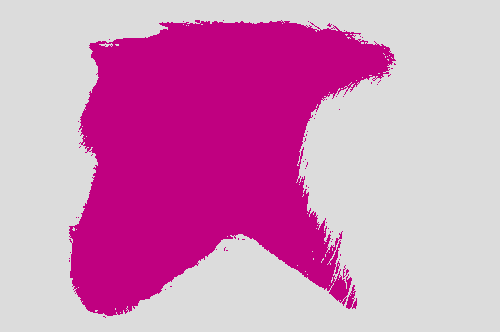} &                  \includegraphics[ width=0.35\linewidth]{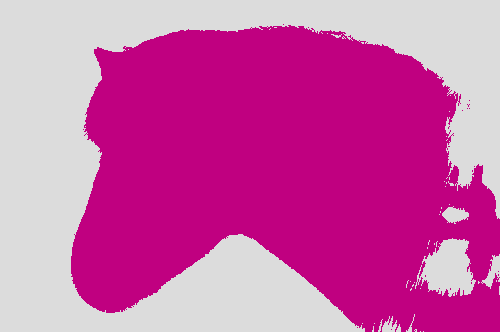}
\\
        \includegraphics[ width=0.35\linewidth]{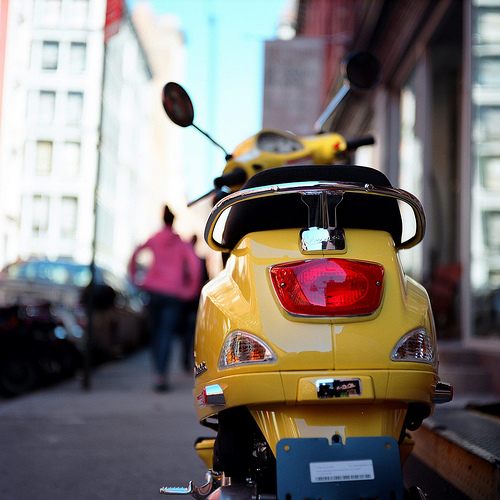}
        &   \includegraphics[ width=0.35\linewidth]{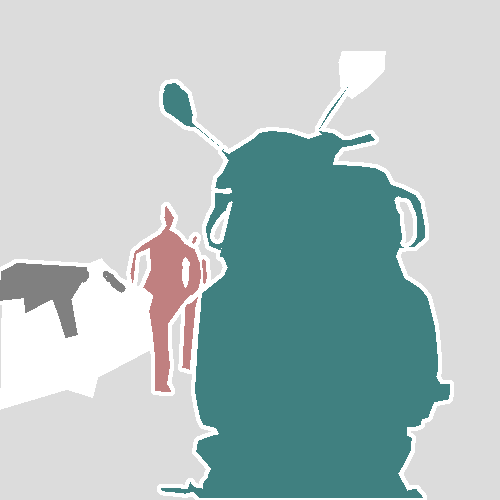}
         &   
         \includegraphics[ width=0.35\linewidth]{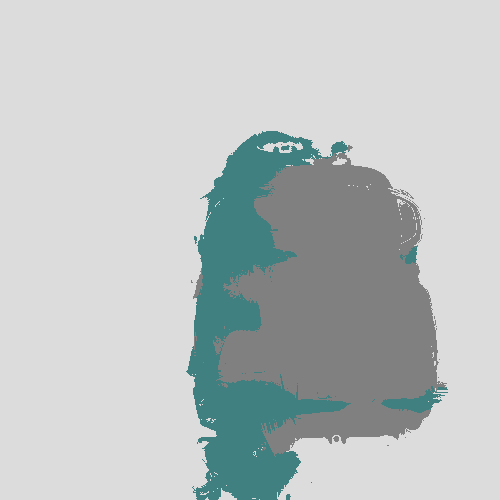} &                  \includegraphics[ width=0.35\linewidth]{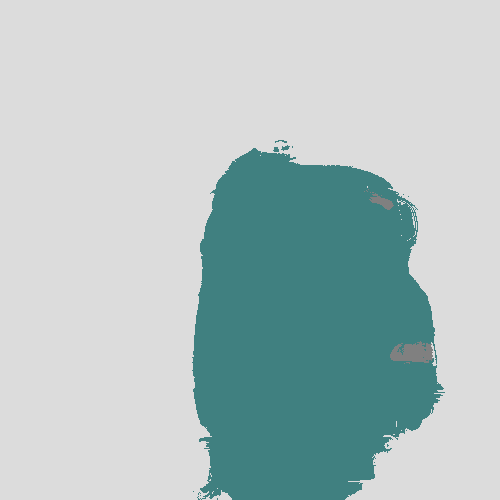}
\\
        \includegraphics[ width=0.35\linewidth]{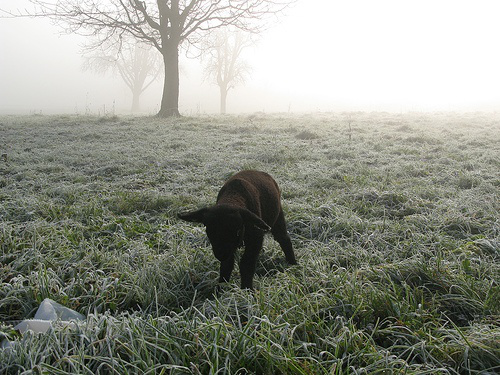}
        &   \includegraphics[ width=0.35\linewidth]{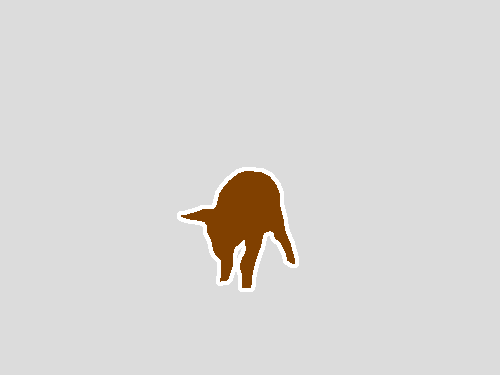}
         &   
         \includegraphics[ width=0.35\linewidth]{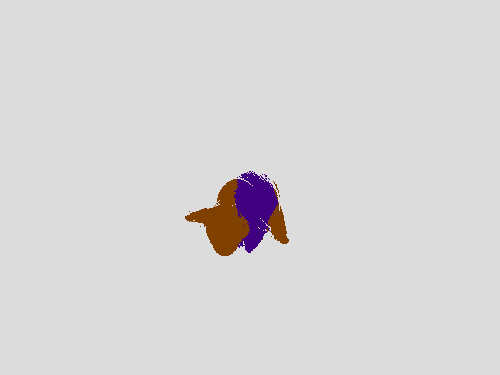} &                  \includegraphics[ width=0.35\linewidth]{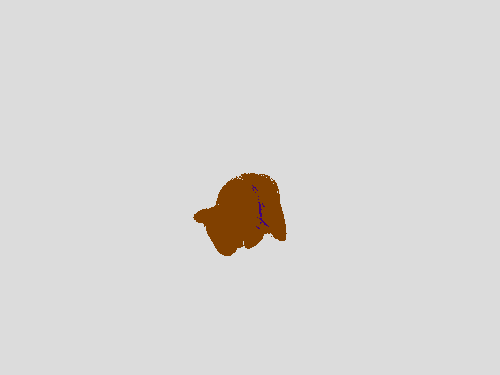} \\
        \includegraphics[ width=0.35\linewidth]{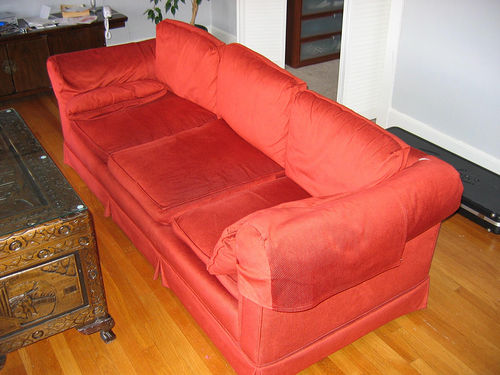}
        &   \includegraphics[ width=0.35\linewidth]{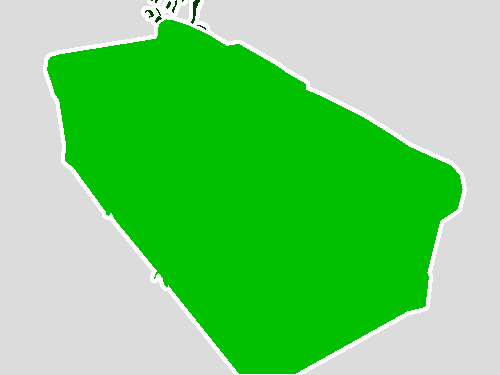}
         &   
         \includegraphics[ width=0.35\linewidth]{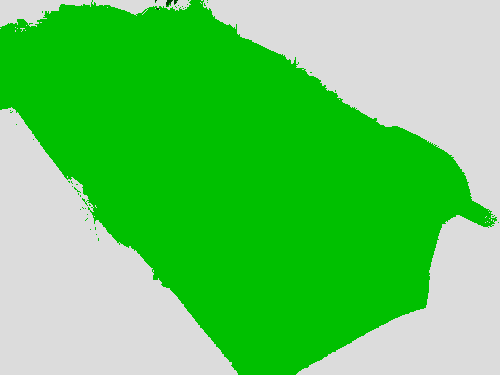} &                  \includegraphics[ width=0.35\linewidth]{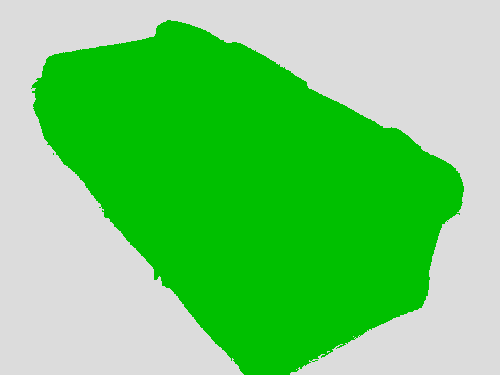}
     \end{tabular}  
     }
     \caption{Additional segmentation results of SEAM and SEAM+ADELE for several examples on the validation set of PASCAL VOC 2012. We set the background color to gray for ease of visualization. Supplementary materials for Figure~\ref{fig:compare_fig_val}.}
    \label{fig:addfigures3}
\end{figure}

\begin{figure*}[ht]
\centering
     \resizebox{0.9\linewidth}{!}{
    \setlength{\fboxrule}{4pt} 
     \setlength{\fboxsep}{0cm}
     \centering
\begin{tabular}{>{\centering\arraybackslash}m{0.16\linewidth} >{\centering\arraybackslash}m{0.16\linewidth} >{\centering\arraybackslash}m{0.16\linewidth} >{\centering\arraybackslash}m{0.16\linewidth} >{\centering\arraybackslash}m{0.16\linewidth} >{\centering\arraybackslash}m{0.16\linewidth}}
         Input  &   Ground truth &  Initial annotations &  Model output after early learning &  Model output after memorization  &  Corrected annotations in ADELE \\
                 \centering
         \includegraphics[ height=1\linewidth,trim={75px 0 50px 0},clip]{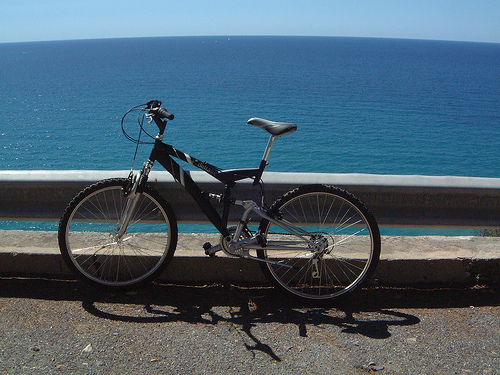}
         &         \includegraphics[ height=1\linewidth,trim={75px 0 50px 0},clip]{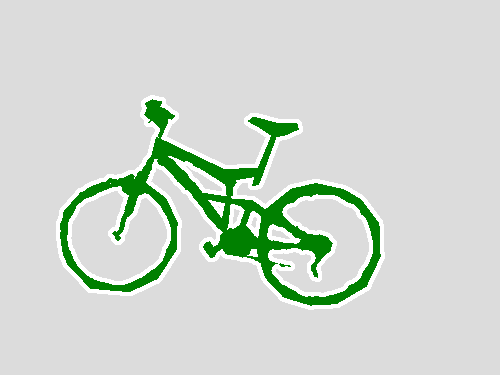} &    
         \includegraphics[ height=1\linewidth,trim={75px 0 50px 0},clip]{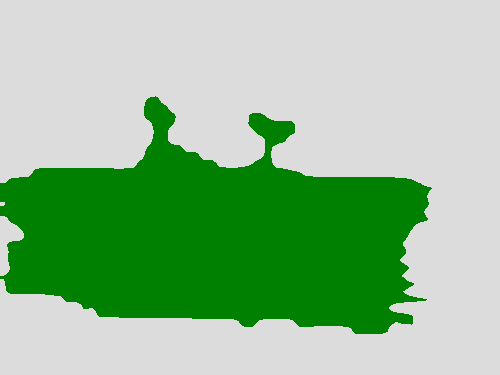} &         {{\includegraphics[ height=1\linewidth,trim={75px 0 50px 0},clip]{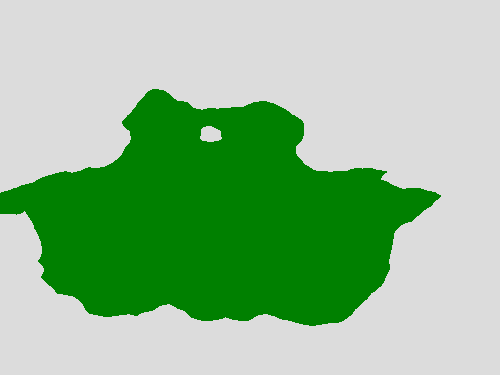}}}
        &   {{\includegraphics[ height=1\linewidth,trim={75px 0 50px 0},clip]{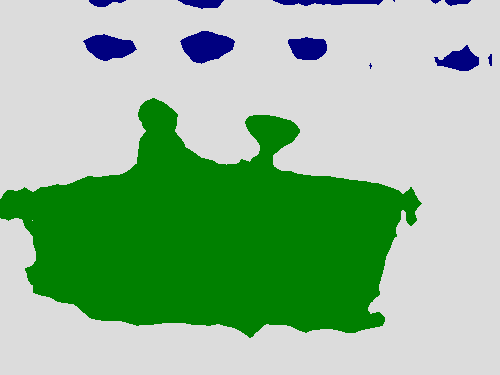}}
        } 
        &  
        \includegraphics[ height=1\linewidth,trim={75px 0 50px 0},clip]{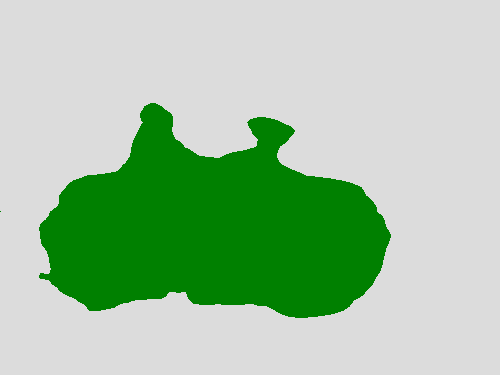}
         \\
                          \centering
         \includegraphics[ height=1\linewidth,trim={67px 0 100px 0},clip]{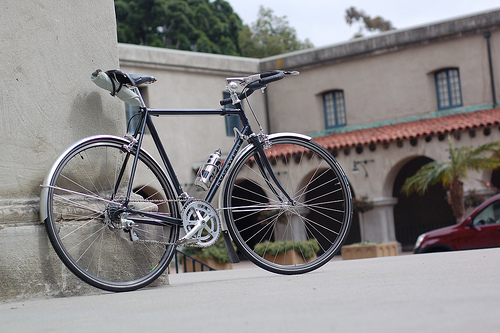}
         &         \includegraphics[ height=1\linewidth,trim={67px 0 100px 0},clip]{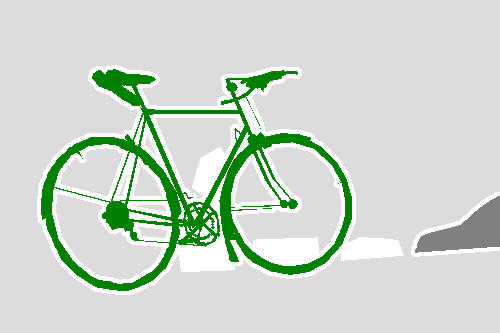} &    
         \includegraphics[ height=1\linewidth,trim={67px 0 100px 0},clip]{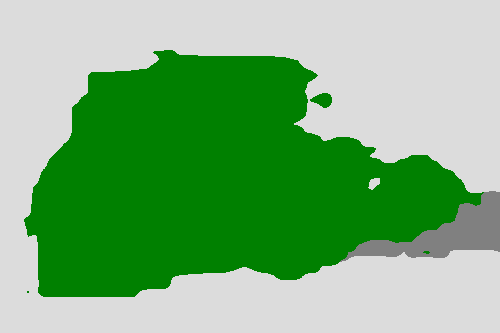} &         {{\includegraphics[ height=1\linewidth,trim={67px 0 100px 0},clip]{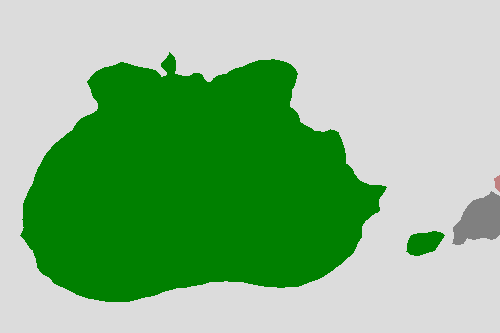}}}
        &   {{\includegraphics[ height=1\linewidth,trim={67px 0 100px 0},clip]{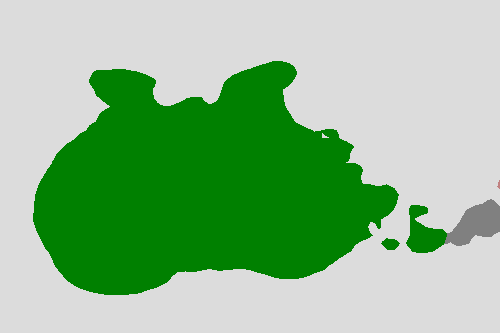}}
        } 
        &  
        \includegraphics[ height=1\linewidth,trim={67px 0 100px 0},clip]{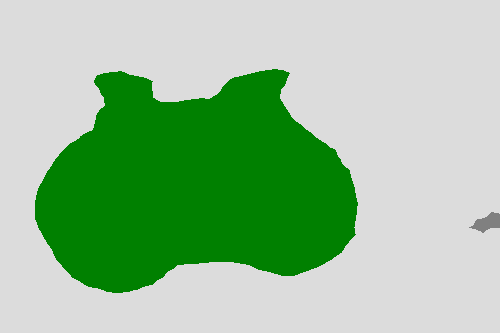}
         \\
         \bigskip
         \includegraphics[ height=1\linewidth,trim={75px 0 50px 0},clip]{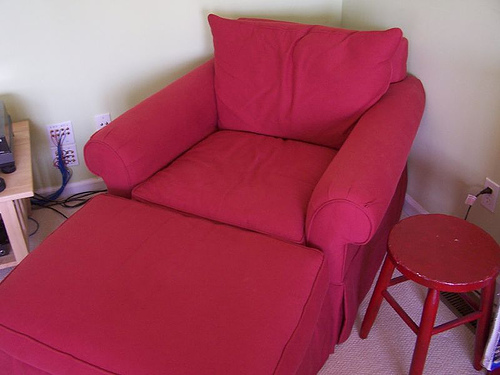}
         &         \includegraphics[ height=1\linewidth,trim={67px 0 100px 0},clip]{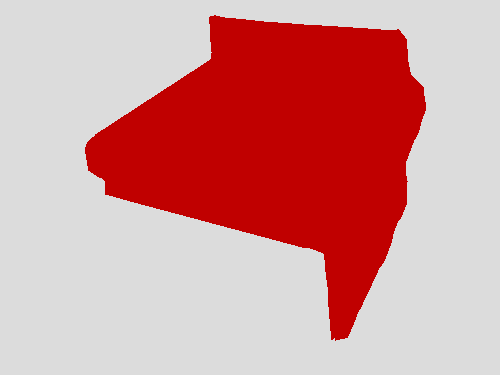} &    
         \includegraphics[ height=1\linewidth,trim={75px 0 50px 0},clip]{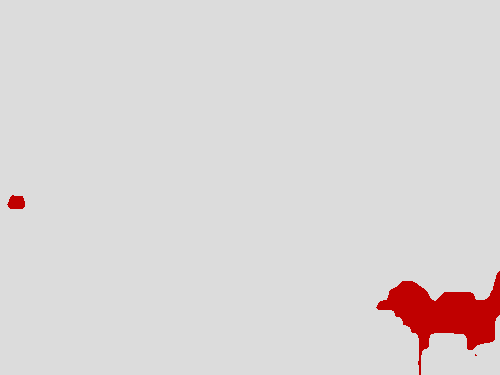} &         {{\includegraphics[ height=1\linewidth,trim={75px 0 50px 0},clip]{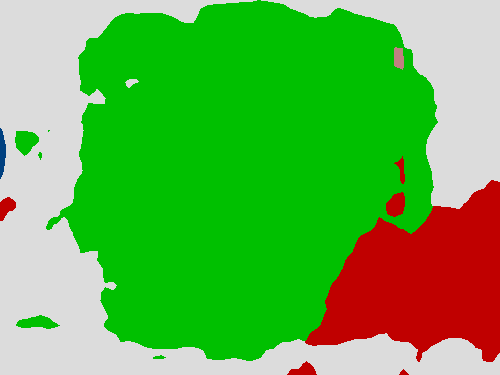}}}
        &   {{\includegraphics[ height=1\linewidth,trim={75px 0 50px 0},clip]{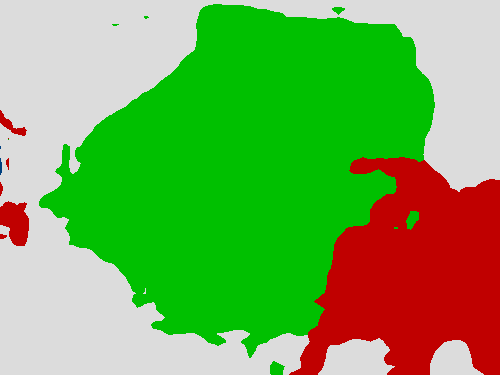}}
        } 
        &  
        \includegraphics[ height=1\linewidth,trim={75px 0 50px 0},clip]{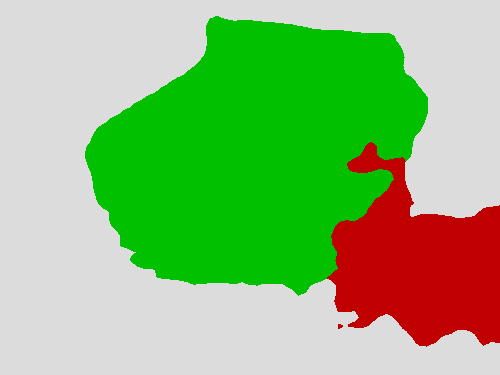} \\
                 \includegraphics[ height=1\linewidth,trim={125px 0 0px 0},clip]{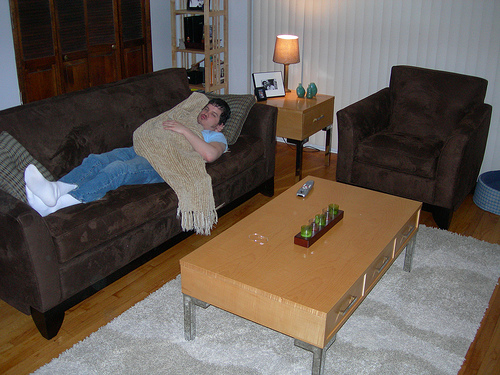}
         &         \includegraphics[ height=1\linewidth,trim={125px 0 0px 0},clip]{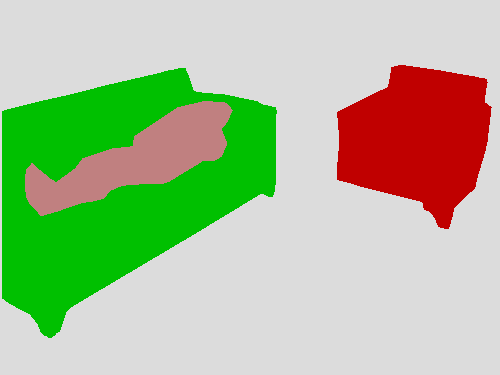} &    
         \includegraphics[ height=1\linewidth,trim={125px 0 0px 0},clip]{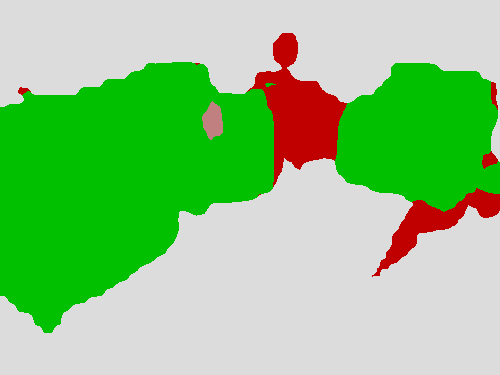} & 
         {{\includegraphics[ height=1\linewidth,trim={125px 0 0px 0},clip]{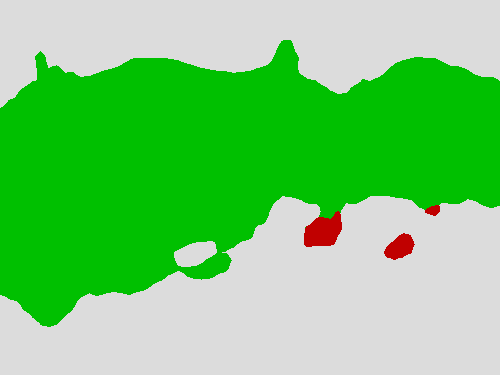}}}
        &   {{\includegraphics[ height=1\linewidth,trim={125px 0 0px 0},clip]{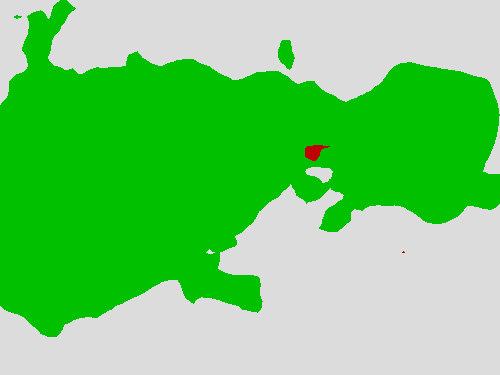}}
        } 
        &  
        \includegraphics[ height=1\linewidth,trim={125px 0 0px 0},clip]{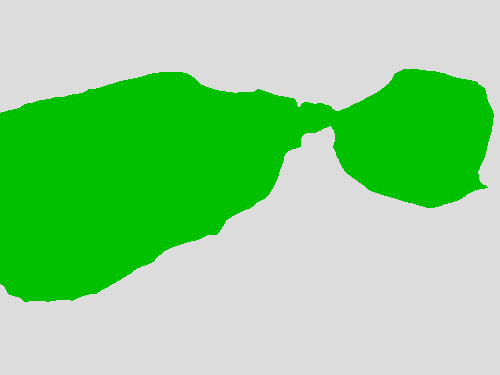}
     \end{tabular}
     }
\centering
\caption{Illustration of a possible limitation of the proposed label-correction approach on PASCAL VOC 2012. The initial annotations coarsely segment the bike and misclassify the chair as a sofa consistently in several training examples.  This highly structured annotation noise could potentially prevent early learning from happening, and therefore from being exploited for label correction. We set the background color to light gray for ease of visualization.
}
\label{fig:limitation}
\end{figure*}

\begin{figure*}[ht]
    \centering
 \resizebox{0.8\linewidth}{!}{
    \begin{tabular}{>{\centering\arraybackslash}p{0.215\textwidth} 
    >{\centering\arraybackslash}p{0.225\textwidth} >{\centering\arraybackslash}p{0.225\textwidth} >{\centering\arraybackslash}p{0.225\textwidth} >{\centering\arraybackslash}p{0.225\textwidth}}	
    \centering
         Input & Ground Truth & Single-scale without label correction & ADELE \\
    \end{tabular}
    }
    \includegraphics[width=0.7\textwidth]{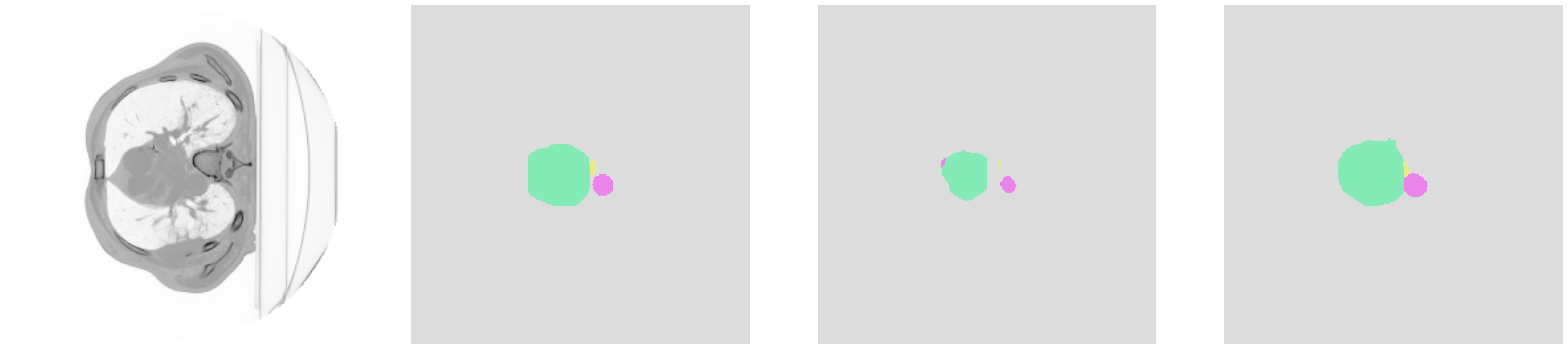} 
    \includegraphics[width=0.7\textwidth]{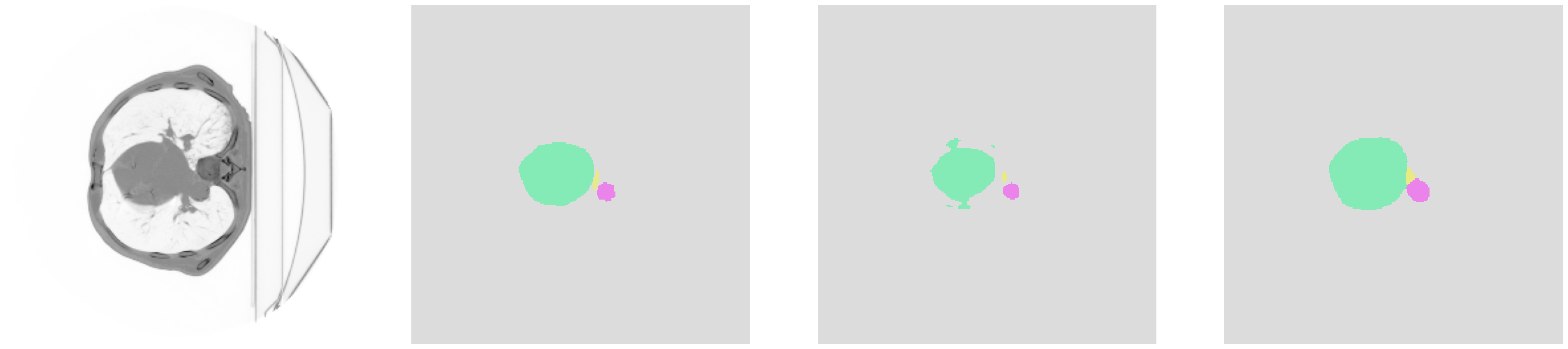} 
    \includegraphics[width=0.7\textwidth]{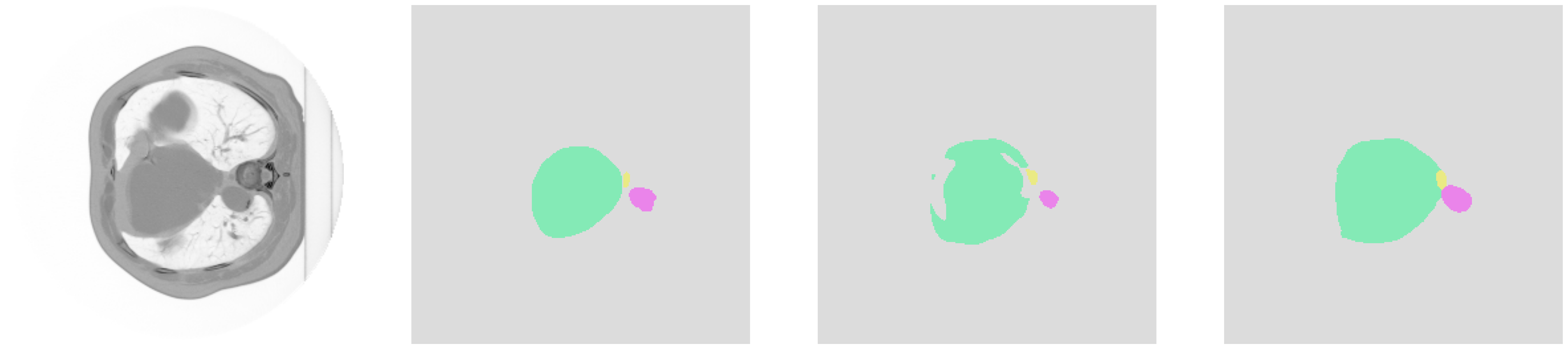} 
    \includegraphics[width=0.7\textwidth]{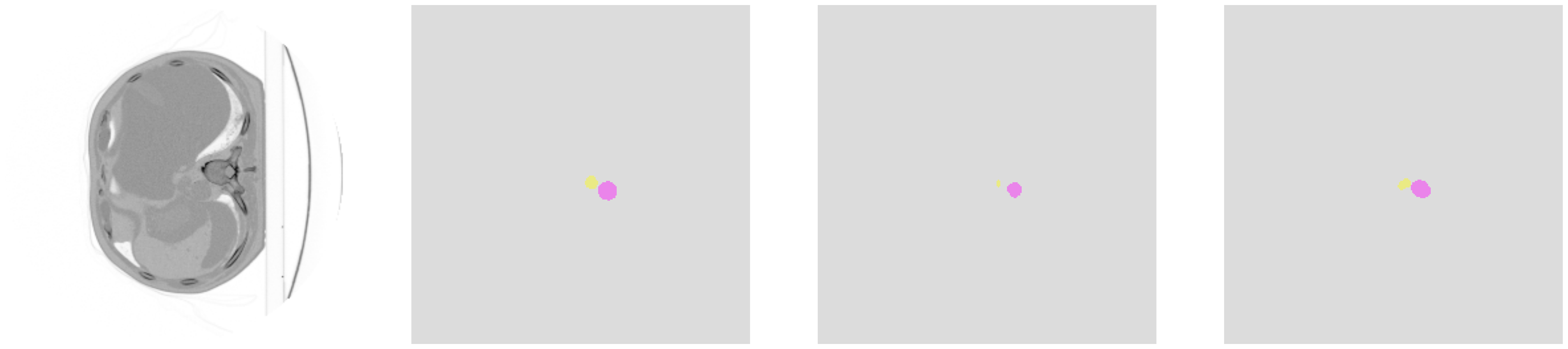}
    \includegraphics[width=0.7\textwidth]{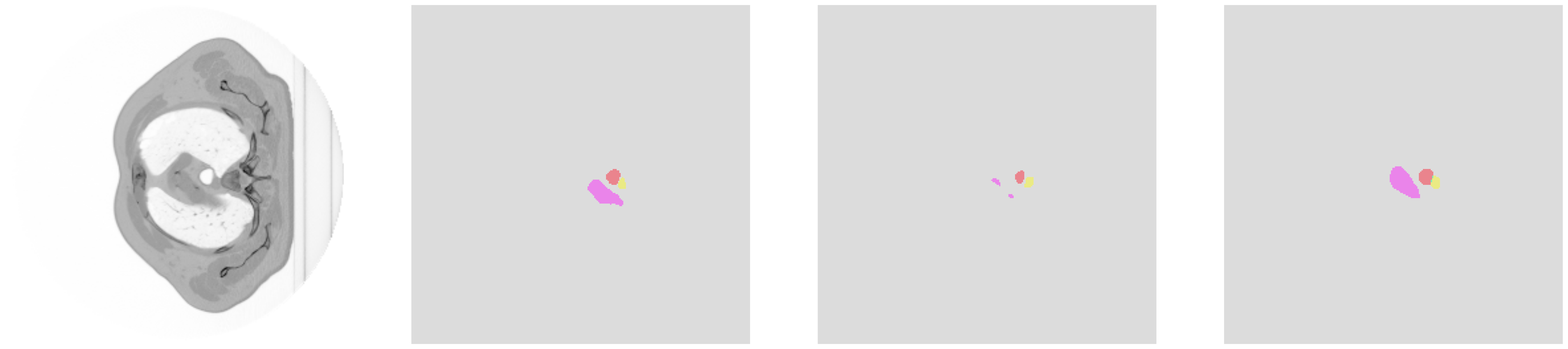} 
    \includegraphics[width=0.7\textwidth]{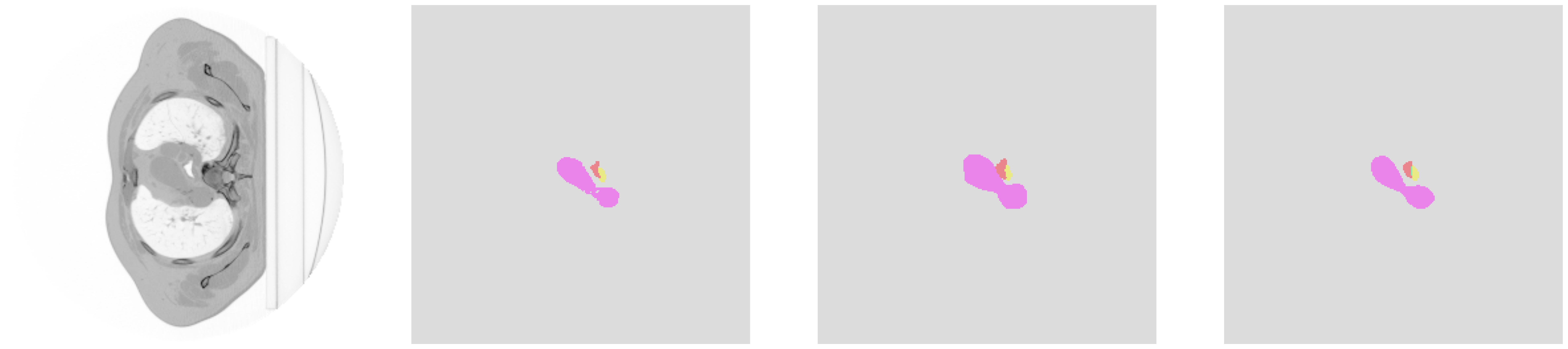}
    \includegraphics[width=0.7\textwidth]{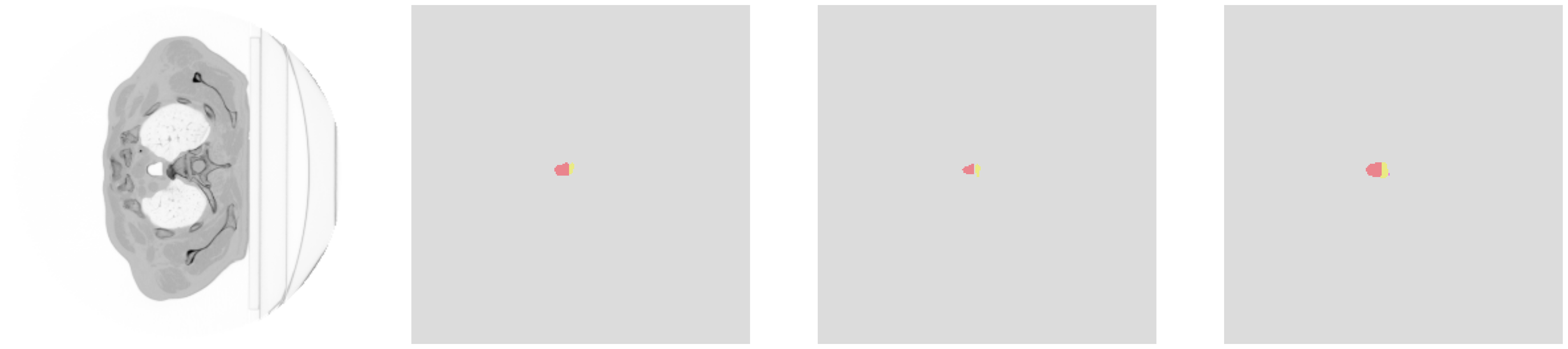}
    \caption{Visualization of the segmentation results of baseline and baseline+ADELE for several examples on the \textit{validation} set of SegTHOR.}
    \label{fig:valsegthor}
\end{figure*}

\clearpage
\section{Additional Experimental Results}
\label{append:addresults}

In this section, we include additional examples (\ref{append:addfigures}) and results (\ref{Append:addtable}). 

\subsection{Additional Examples}
\label{append:addfigures}

\paragraph{WSSS dataset (PASCAL VOC 2012)} 
\begin{itemize}[leftmargin=*]

    \item \textit{Early-learning.} (Supplementary material for Figure~\ref{fig:EL} and~\ref{fig:curvefitting} in Section~\ref{sec:method}) Figure~\ref{fig:early_learning_voc} demonstrates that early-learning and memorization on a segmentation CNN trained with noisy annotations generated by a classification model for the standard WSSS dataset (PASCAL VOC2012).  In Figure~\ref{fig:ELzoomin} we show the zoomed-in version of the early-learning $\text{IoU}_{el}$ curves. Early-learning happens for most of the classes at different speeds.
    Figure~\ref{fig:curvefittin_voc} show the curve fitting illustration for PASCAL VOC. Figure~\ref{fig:figure3_moreexp} shows that the label correction begins close to the end of the early-learning phase for most of the semantic categories for PASCAL VOC.  
    \item \textit{Multiscale consistency.} Figure~\ref{fig:correction_quality_VOC} shows that multi-scaleconsistency regularization leads to more accurate corrected annotations for PASCAL VOC.
    \item  \textit{Superior performance.} (Supplementary material for Figure~\ref{fig:compare_fig_val} in Section~\ref{Sec:ExpWSSS}) We provide more visualization examples on the validation set of PASCAL VOC in Figure~\ref{fig:addfigures}, ~\ref{fig:addfigures2} and ~\ref{fig:addfigures3} comparing ADELE to the baseline method SEAM. ADELE reduces false positives for some examples (\eg boat, person, sofa, bottle, tv \etc) and produces a more complete segmentation of other examples  (\eg cat, dog, bird, bus, bottle, tv, horse \etc).
    \item \textit{Highly structured annotation noise.} (Supplementary material for Figure~\ref{fig:compare_fig_val})  Figure~\ref{fig:limitation} shows training examples with highly structured noise, which may prevent early learning from happening, and therefore from being used for label correction.
\end{itemize}

\paragraph{More visual examples for SegTHOR}
  
\begin{itemize}[leftmargin=*]
    \item \textit{Superior performance}. (Supplementary material for Section~\ref{sec:segthor}). We show some visual examples that indicate ADELE improves segmentation performance on the validation set of SegTHOR dataset in Figure~\ref{fig:valsegthor}. 
\end{itemize}

\subsection{Additional Results}
\label{Append:addtable}
\paragraph{ADELE outperforms methods using external information.}(Supplementary material for Section~\ref{Sec:ExpWSSS}) ADELE using initial labels generated from SEAM~\cite{wang2020self} and ICD~\cite{fan2020learning} achieves state-of-the-art performance without using external saliency models. This performance is on par with or even slightly better than methods that rely on external saliency models ~\cite{jiang2019integral,zhang2020splitting,sun2020mining,yao2021non}.  To show that our method is complementary with other more advanced WSSS methods,  we have conducted an experiment with a recent WSSS method NSROM~\cite{yao2021non}. 
ADELE+NSROM achieves mIoU of 71.6 and 72.0 on the validation and test set respectively, which is the SoTA for WSSS with ResNet segmentation backbone.

\begin{table*}[h]
\centering
\begin{tabular}{r|cccl} 
\toprule
\centering
Method &Supervision & Backbone & val & test \\
\hline
DSRG~\cite{huang2018weakly} & I.+ S. & ResNet-101 & 61.4 & 63.2 \\
SPLIT MERGE~\cite{zhang2020splitting} & I.+S. &   ResNet-50 & 66.6 & 66.7 \\
MCIS~\cite{sun2020mining} & I.+S. & ResNet-101 & {66.2} & {66.9} \\
NSROM~\cite{yao2021non} & I.+S. & ResNet-101 & {68.2} & {68.5} \\
NSROM$^{*}$~\cite{yao2021non} & I.+S. & ResNet-101 & \textbf{70.4} & \textbf{70.2} \\
\hline 
SEAM + Ours & I. & ResNet-38 & {69.3} & {{68.8}} \\
ICD + Ours & I. & ResNet-38 & {68.6} & {68.9} \\
NSROM + Ours & I.+S. & ResNet-38 & \textbf{71.6} & \textbf{72.0} \\
  \bottomrule
\end{tabular}
\caption{Comparison with previous works that use external saliency models~\cite{jiang2013salient} on the PASCAL VOC 2012 dataset~\cite{everingham2015pascal}. I. stands for image-level labels, S. stands for external saliency models. * denotes model is pre-trained on MS-COCO}
\label{Table:comparewSoTA_2}
\end{table*}

\paragraph{ADELE improves performance for most categories.}(Supplementary material for Figure~\ref{fig:perclasscomparision} in Section~\ref{Sec:ExpWSSS})
Table~\ref{Table:perclasscomparision} shows that ADELE performs substantially better than the baseline method SEAM on most of the semantic categories, as well as on average.

\paragraph{ADELE improves performance across different noise levels.} We provided the full result of Figure~\ref{fig:noise_level_baselines} in Table~\ref{Table:noiselevels} for SegTHOR dataset, which shows that ADELE can improve the model performance across different noise level.
 \begin{table}[H]
 \resizebox{1.0\linewidth}{!}{
 \setlength{\tabcolsep}{1pt}
\begin{tabular}{l|ccccccccccccccccccccc|c} 
\toprule
Method & bkg& aero& bike& bird& boat& bottle& bus& car& cat& chair& cow& table& dog& horse& mbk& person& plant& sheep& sofa& train& tv& mIoU \\
\hline
SEAM& 88.8& 68.5&\textbf{33.3}& 85.7& 40.4& 67.3& 78.9& 76.3& 81.9& \textbf{29.1}& 75.5& \textbf{48.1}& 79.9& 73.8& 71.4& 75.2& \textbf{48.9}& 79.8& 40.9& \textbf{58.2}& 53.0& 64.5 \\
SEAM + Ours& \textbf{91.1} & \textbf{77.6}&	33.0 &	\textbf{88.9}&	\textbf{67.1}&	\textbf{71.7}&	\textbf{88.8}&	\textbf{82.5}&	\textbf{89.0}&	26.6&	\textbf{83.8}&	44.6&	\textbf{84.4}&	\textbf{77.8}&	\textbf{74.8}&	\textbf{78.5}&	43.8&	\textbf{84.8}&	\textbf{44.6}&	{56.1}	& \textbf{65.3}&\textbf{69.3}\\
  \bottomrule
\end{tabular}
}
\smallskip
\caption{Category-wise comparison of the IoU ($\%$) of SEAM~\cite{wang2020self} and SEAM combined with the proposed method ADELE on the validation set of PASCAL VOC 2012.
}
\label{Table:perclasscomparision}
\end{table}

\begin{table}[H]
\centering
\centering
\begin{tabular}{>{\centering\arraybackslash}m{0.1\linewidth} >{\centering\arraybackslash}m{0.1\linewidth}| >{\centering\arraybackslash}m{0.20\linewidth} >{\centering\arraybackslash}m{0.16\linewidth} >{\centering\arraybackslash}m{0.16\linewidth}>{\centering\arraybackslash}m{0.16\linewidth}}
\toprule
Iteration of erosion/dilation & Annotation mIoU & Multiscale input augmentation (baseline) & Multiscale label correction & Multiscale consistency regularization & ADELE \\
\hline
0 & 1.00 & 0.745 & 0.743 & 0.762& \textbf{0.766} \\
1 & 0.91 & 0.743 & 0.726 &\textbf{0.759} & {0.757} \\
2 & 0.73 & 0.702 &0.710 &0.714 & \textbf{0.734} \\
3 & 0.61 & 0.646 & 0.658 &0.647 &	\textbf{0.711} \\
4 & 0.52 & 0.556 &0.651&0.606&	\textbf{0.666} \\
6 & 0.39 & 0.446 & 0.556&0.514 &	\textbf{0.564} \\
8 & 0.28 & 0.416 &0.407&0.423&	\textbf{0.481} \\
 \bottomrule
\end{tabular}
\caption{ The mIoU ($\%$) comparison of the baseline and ADELE on the test set of SegTHOR~\cite{lambert2020segthor}. We report the test mIoU of the model that achieves best mIoU on the validation set.  It can be seen that ADELE stably improve the model performance across different noise levels. 
}
\label{Table:noiselevels}
\end{table}

\section{Implementation details}
\label{append:implementationdetails}

 \begin{algorithm}[t]
\caption{\label{alg:algo_ELR}\ \ 
Pseudocode for Proposed Annotation Correction. 
}
\begin{tabbing}

\Req $\vy_i\; 1 \leq i \leq N$  // training noisy annotations\\ 
\Req $\vx_i ,\; 1 \leq i \leq N$  // training images\\ 
\Req $\text{NN}(\vx)$ // segmentation network  \\
\Req $C$ \= = number of categories \\
\Req $t$ \= the training epoch \\
\Req $f_c(t),\; 1 \leq c \leq C$ // curve function fitted on training IoU for each category $c$  \\
\Req $\tau$ \= = threshold, $0 < \tau < 1$ \\
\Req $r$  \=  = threshold for when to correct annotation \\
\Req $S$ \= = set of rescaling transforms \\

{\bf for} each minibatch $B$ {\bf do}\\
\X {\bf for} $b$ in $B$ {\bf do}\\
\XX {\bf for} $s$ in $S$ {\bf do}\\\
\XXX ${p_{bs}} = \text{NN}(S(\vx_i))$    // evaluate the network to obtain model's outputs corresponding to each scaled version of inputs\hspace*{-16mm} \= \\
\XX {\bf end for}\\
\XX $q_b$ \= $\gets \frac{1}{S}\sum_{s=1}^S p_{bs}$  \hspace{6mm}// {obtain the averaged outputs across different scales}\\
\X {\bf end for}\\

\X {\bf for} each class $c$ in $[1, C]$ {\bf do} \\
\XX {\bf If} ${|f_c'(1) - f_c'(t)|}/{|f_c'(1)|} > r$ {\bf do}  // when growth of IoU slowdown \\
\XXX {\bf for} image $x_b$ containing object of category $c$ {\bf do} \\
\XXXX $\vy_b[q_{bc} \geq \tau]  = \argmax q_b[q_{bc} \geq \tau]$;
// convert outputs to hard label and conduct  pixel-wise annotations correction\\
\XXX {\bf end for}\\

\X {\bf end for}\\
{\bf return} $\vy_1,\vy_2,\dots, \vy_N$ // corrected annotations\\ 
\end{tabbing}
\vspace*{-1.5mm}
\end{algorithm}
\subsection{Hyperparameters}
 Table~\ref{Table:hyperparameter} provides a complete list of hyperparameter values for our experiments. We use almost identical hyperparameters of ADELE on PASCAL VOC 2012 and SegTHOR. $\lambda$ (\emph{consistency strength}) controls the strength of the consistency regularization added to the cross entropy loss. $\rho$ (\emph{consistency confidence threshold}) is the threshold that determines when multiscale consistency regularization is applied (when the maximum prediction probability for any category in any pixel of the average $q$ is above $\rho$). $r$ (\emph{curve fitting threshold}) is the threshold that controls label correction for each semantic category (see Equation~\eqref{eq:threshold_r}). $\tau$ (\emph{label correction confidence threshold}) is the threshold that determines which pixels are corrected by the model prediction. Only the pixels with confidence (maximum prediction probability for any category) above $\tau$ are corrected. Note that we report ablation studies for most of these hyperparameters in Section~\ref{append:additionalablation}.

\begin{table}[H]
\centering
\begin{tabular}{r|cccl} 
\toprule
\centering
Dataset & $\lambda$ & $\rho$ & $r$ & $\tau$ \\
\hline
PASCAL VOC 2012 & 1 & 0.8 & 0.9 & 0.8 \\
SegThor & 1 & 0.8 & 0.9 & 0.7 \\
  \bottomrule
\end{tabular}
\smallskip
\caption{Complete list of ADELE hyperparameters for PASCAL VOC 2012~\cite{everingham2015pascal} and SegTHOR~\cite{lambert2020segthor}. }
\label{Table:hyperparameter}
\end{table}

\subsection{Medical segmentation with simulated noise: SegTHOR}
Here we provide implementation details for our experiments on SegTHOR (Supplementary material for Section~\ref{sec:segthor} in the main paper).

\paragraph{Details of Noise Synthesis.}
We apply dilation and erosion on the ground-truth segmentation masks to simulate ``over-annotation'' and ``under-annotation'' labels respectively. In particular, ``over-annotation'' labels tend to assign background pixels surrounding a target organ to the class of this organ. On the contrary, ``under-annotation'' labels tend to assign a foreground pixel on the edge of a target organ to the background class. These two noise patterns have been previously utilized to simulate common errors that human annotators make during manual segmentation~\cite{zhang2020disentangling}.
In our experiments, we randomly choose the type and degree of synthetic noise that will be applied to each example.

\paragraph{Training.} For SegTHOR~\cite{lambert2020segthor}, we use one NVIDIA V100 GPU to train the model. We train the UNet~\cite{ronneberger2015u} using SGD optimizer. To optimize the hyper-parameters, we search for
the learning rate in $\{0.1, 0.01, 0.001, 0.0001, 0.00001\}$ and set to 0.01. $\lambda$,  $\rho$,  $r$ and $\tau$ are set according to Table~\ref{Table:hyperparameter}.
We trained our model for 100 epochs with a batch size of 5. 

\paragraph{Inference during testing.} We conduct the single-scale evaluation using the input without augmentations. 

\paragraph{Generating model prediction to conduct label correction.} For the medical dataset, since we do not apply any augmentation to the input images, we directly use the outputs during training at every iteration to correct the labels (the model output are processed with $\argmax$ to produce hard labels for label correction). This is in contrast to PASCAL VOC where we compute the outputs at the end of each epoch, as explained in Section~\ref{subsec:addablationSegTHOR}.

\subsection{Weakly Supervised Semantic Segmentation: PASCAL VOC 2012}

 We provide the implementation details of the model for PASCAL VOC 2012. (Supplementary material for Section~\ref{Sec:ExpWSSS}).

\paragraph{Training.} For PASCAL VOC 2012~\cite{everingham2015pascal}, we use two NVIDIA Quadro RTX 8000 GPUs to train the model. We use the official code of  AffinityNet~\cite{ahn2018learning}, SEAM~\cite{wang2020self} and ICD~\cite{fan2020learning} to generate initial pixel-level annotations that are used for training the segmentation network. The experimental settings to train the segmentation network follow~\cite{ahn2018learning,zhang2020reliability,shimoda2019self,wang2020self} in which DeepLabv1~\cite{chen2014semantic} is adopted. The model uses ResNet38~\cite{wu2019wider}  as a backbone, with the initial weight load from  ImageNet~\cite{deng2009imagenet} pretrained classfication model. Following the same settings as SEAM~\cite{wang2020self}, we use a SGD optimizer with momentum 0.9 and weight decay $5e^{-4}$. The initial learning rate $\operatorname{lr}_{i n i t}$ is set to 0.001, and 
reduced following a polynomial function of the iteration number $itr$: $\operatorname{lr}_{i t r}=\operatorname{lr}_{i n i t}\left(1-\frac{i t r}{max\_itr}\right)^{\gamma}$ with $\gamma=0.9$. We train our segmentation network for 20000 iterations ($max\_{itr} = 20000$) with a batch size of 10. 
The input images are randomly scaled and then randomly cropped to $448\times448$.

\paragraph{Inference during testing.} During testing, we use the same inference pipeline as SEAM~\cite{wang2020self}, which includes multi-scale inference~\cite{ahn2018learning,wang2020self,fan2020learning, zhang2020splitting} and CRF~\cite{krahenbuhl2011efficient}. 

\paragraph{Generating model predictions to conduct label correction.}

Updating the model using the model predictions after processing each batch 
would be difficult for the PASCAL VOC 2012 dataset. The reason is that random data augmentations (\eg rescaling, cropping, random changing contrast in the image, \etc) are often applied~\cite{ahn2018learning,wang2020self,fan2020learning, zhang2020splitting} and these augmentations would need to be inverted in order to use the output for label correction  (otherwise the predictions would be inconsistent across training epochs). In fact, some augmentations are not invertible at all, \eg cropping cuts the object out thus is not invertible, rescaling might result in loss of information thus is not invertible as well. In order to avoid this issue on the PASCAL VOC dataset, we evaluate the model at the end of each training epoch on inputs without any random augmentation, and then use the model outputs to perform label correction (the model output are processed with $\argmax$ to produce hard labels for label correction).

\section{Additional ablation studies}
\label{append:additionalablation}

\subsection{Medical segmentation with simulated noise: SegTHOR}
\label{subsec:addablationSegTHOR}

Here we report additional ablation studies for different components in the proposed label correction method on SegTHOR dataset (Supplementary material for Section~\ref{sec:segthor}).

\paragraph{Different options for label correction} As described in Section~\ref{subsec:ADELEmethod}, ADELE uses the model outputs  to correct labels. In this section, we compare two options for computing these outputs. 
\begin{itemize}[leftmargin=*]
    \item \textit{Iteration}: the outputs are computed after each training iteration.
    \item \textit{Epoch}: the outputs are computed at the end of each training epoch.
\end{itemize}
Table~\ref{Table:segTHORtestwerrorbarablation} shows that on SegTHOR \textit{Iteration} performs slightly better on the best test mIoU than \textit{Epoch}, and outperforms it substantially on the mIoU at the last epoch, suggesting that \textit{Iteration} is more effective in preventing memorization. Both two options are improving results with respect to the baseline. 

\begin{figure}[t]
\begin{minipage}[b]{0.55\linewidth}
    \centering
    \includegraphics[width=1\linewidth]{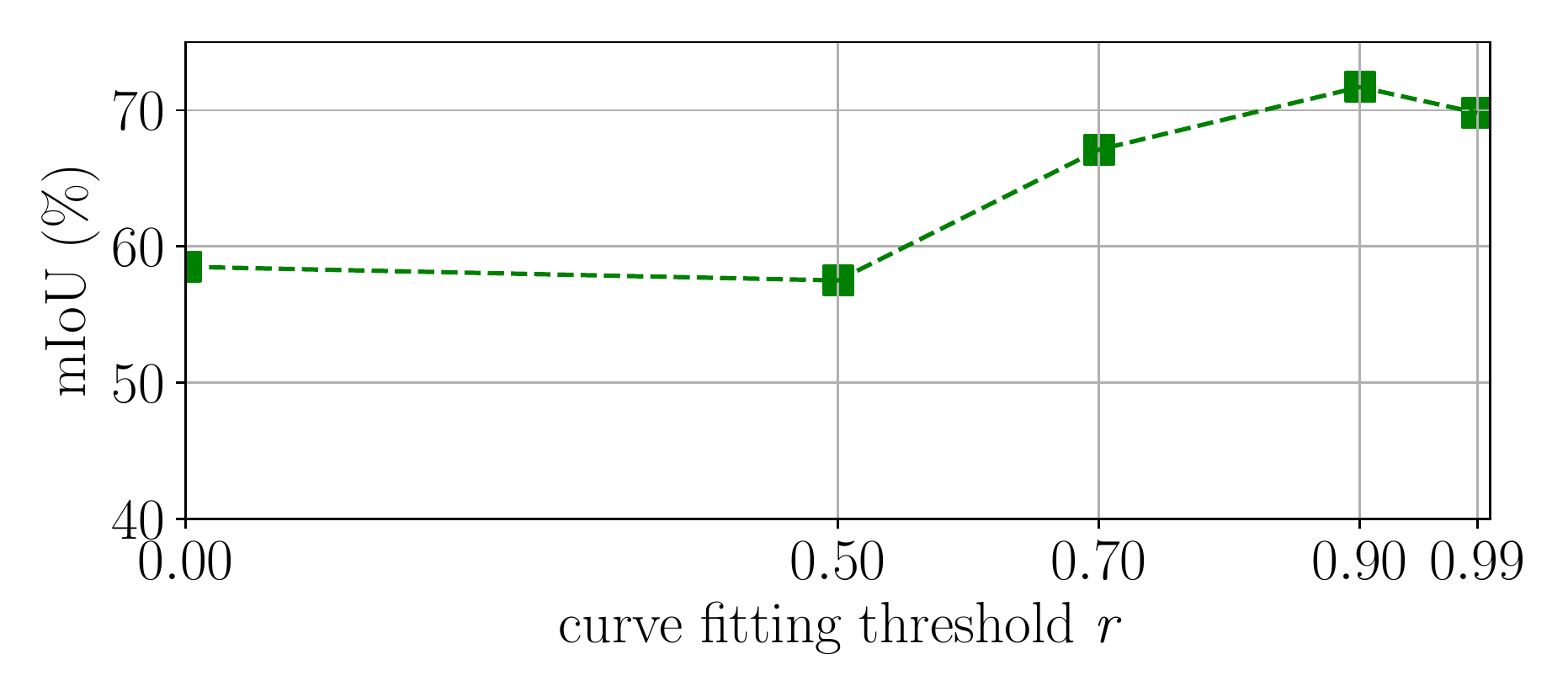}
\end{minipage}
\begin{minipage}[b]{0.44\linewidth}
\centering
 \resizebox{1.0\linewidth}{!}{
\begin{tabular}{r|cccl} 
\toprule
\centering
$r$ & Best val & Last epoch & Max test \\
\hline
0.0 & 58.5 &	55.5 &	58.5  \\
0.5 & 57.5&	55.4	&57.9 \\
0.7 & 67.1&	66.1&	67.1  \\
0.9 &\textbf{71.7}&	\textbf{71.0}&	\textbf{71.7}  \\
0.99 & 69.8	&64.7&	69.8 \\
  \bottomrule
\end{tabular}
}
  \bigskip
  \smallskip
\end{minipage}
\caption{Ablation study for $r$ on SegTHOR. We fixed the other hyperparameters to the default settings in Table~\ref{Table:hyperparameter}. We report the test mIoU ($\%$) at the best validation epoch on the \textbf{Left}, the full result is shown on the \textbf{Right}. }
\label{fig:ablationsegTHORr}
\end{figure}

\begin{figure}[t]
\begin{minipage}[b]{0.55\linewidth}
    \centering
    \includegraphics[width=1\linewidth]{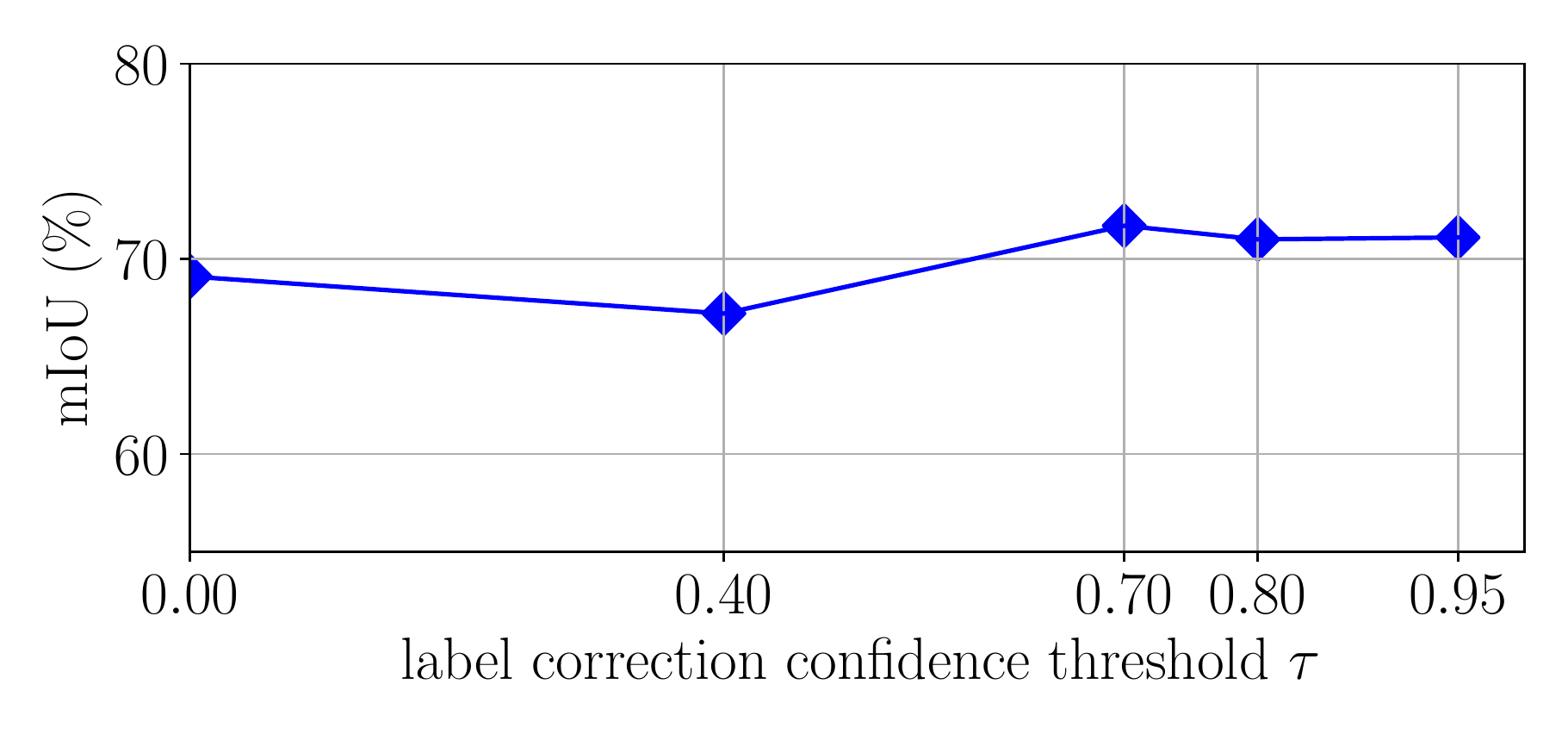}
\end{minipage}
\begin{minipage}[b]{0.44\linewidth}
\centering
 \resizebox{1.0\linewidth}{!}{
\begin{tabular}{r|cccl} 
\toprule
\centering
$\tau$ & Best val & Last epoch & Max test \\
\hline
0&	69.1&	68.8&	69.1 \\
0.4& 67.2 & 67.1& 68.0 \\
0.7	&\textbf{71.7}&	\textbf{71.0}&	\textbf{71.7} \\
0.8	&71.0&	70.3&	71.0 \\
0.95& 71.1 & 71.2 & 71.3\\
  \bottomrule
\end{tabular}
}
\bigskip
  \smallskip
\end{minipage}
\caption{Ablation study for $\tau$ on SegTHOR. We fixed the other hyperparameters to the default settings in Table~\ref{Table:hyperparameter}. We report the test mIoU ($\%$) at the best validation epoch on the \textbf{Left}, the full result is shown on the \textbf{Right}. The result shows that ADELE is not very sensitive to the value of $\tau$.  }
    \label{fig:ablationsegTHORtau}
\end{figure}

\begin{table}[H]
\centering
\begin{tabular}{r|cccl} 
\toprule
\centering
Method & Best val & Last epoch & Max test \\
\hline
ADELE (\textit{Iteration }) & \textbf{71.1 $\pm$ 0.7}&	\textbf{70.8 $\pm$ 0.7}&	\textbf{71.2 $\pm$ 0.6}  \\
ADELE (\textit{Epoch}) & {69.6 $\pm$ 0.8}&	{64.1 $\pm$ 1.3}&	{70.2 $\pm$ 0.5}  \\
  \bottomrule
\end{tabular}
\caption{mIoU($\%$) of ADELE on SegTHOR when label correction is based on model output at the end of each iteration or each epoch. The former achieves better results. }
\label{Table:segTHORtestwerrorbarablation}
\end{table}

\paragraph{{$r$ and $\tau$ for label correction}.} We report the ablation for $r$ and  $\tau$ for ADELE on SegTHOR dataset in Figure~\ref{fig:ablationsegTHORr} and \ref{fig:ablationsegTHORtau} respectively.  For the $r$ value, we observe similar results as in the PASCAL VOC dataset. If the $r$ value is too small (\eg 0, 0.3) and label correction is conducted too early, the network has not been trained well. This degrades the label correction quality, which hurts the generalization of the model. If the $r$ value is too large (\eg 0.99) then the network barely conduct label correction for any class before stopping training. Results are again very robust to the choice of $\tau$ value.

\subsection{Weakly Supervised Semantic Segmentation: PASCAL VOC}

\begin{figure}[t]
    \centering
    \includegraphics[width=1\linewidth,trim={0px 0 0px 0},clip]{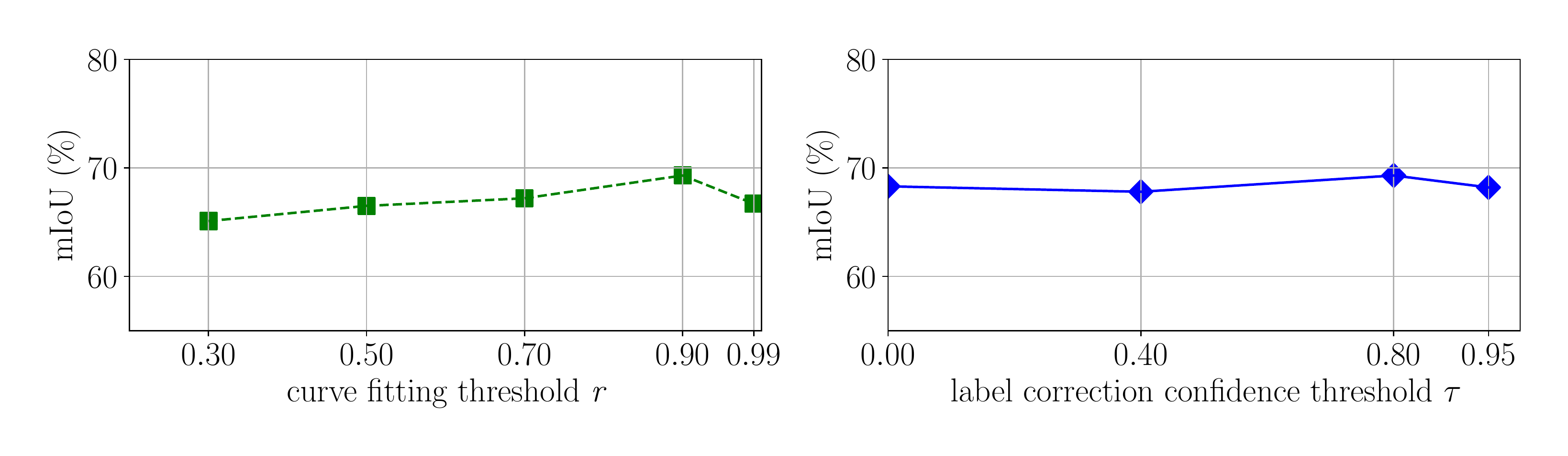}
    \caption{Ablation study for $r$ and $\tau$ on PASCAL VOC. We fixed the other hyperparameters to the default settings in Table~\ref{Table:hyperparameter}. We report the validation mIoU ($\%$) at the last training epoch.}
    \label{fig:ablationVOCrtau}
\end{figure}

\paragraph{$r$ and $\tau$ for label correction.} We report the ablation result for $r$ and  $\tau$ of ADELE on PASCAL VOC dataset in Figure~\ref{fig:ablationVOCrtau}. Small $r$ values  encourage the model to conduct label correction earlier, larger $r$ values delay correction. If the $r$ value is too small (\eg 0.3) and label correction is conducted too early, the network has not been trained well. This degrades the label correction quality, which hurts the generalization of the model. If the $r$ value is too large (\eg 0.99), then the method barely conducts label correction for any class before the end of training, which results in a performance similar to the case without label correction. We observe that our model is very robust to the choice of $\tau$.

\end{document}